\documentclass[journal]{IEEEtran}

\usepackage{times}

\usepackage{multicol}
\usepackage[bookmarks=true]{hyperref}
\hypersetup{
 colorlinks=true,
 linkcolor=red,
 filecolor=blue,
 citecolor=green,      
 urlcolor=blue,
 }

\usepackage{amsmath,amsfonts,amssymb,amsthm,epsfig,epstopdf,url,array,xcolor,soul,subfigure,multirow}
\usepackage{cite}

\usepackage[noend]{algorithmic}
\usepackage[ruled,noend]{algorithm2e}
\SetAlCapHSkip{0pt}

\newtheorem{theorem}{Theorem}

\newtheorem{problem}{Problem}
\newtheorem{remark}{Remark}

\newcommand{\scaleMathLine}[2][1]{\resizebox{#1\linewidth}{!}{$\displaystyle{#2}$}}

\newcommand{\mb}[1]{{\mathbf{#1}}}
\newcommand{\mc}[1]{{\mathcal{#1}}}

\IEEEoverridecommandlockouts

\begin{document}
% paper title
\title{Robust Multi-Robot Active Target Tracking \\ Against Sensing and Communication Attacks} % for ieee conference

% You will get a Paper-ID when submitting a pdf file to the conference system
\author{Lifeng Zhou,~\IEEEmembership{Student Member,~IEEE,} and Vijay Kumar,~\IEEEmembership{Fellow,~IEEE}
\thanks{L. Zhou and V. Kumar are with the GRASP Laboratory, University of Pennsylvania, Philadelphia, PA, USA (email: {\tt\small\{lfzhou, kumar\}@seas.upenn.edu}).}
\thanks{This research was sponsored by the Army Research Lab through ARL DCIST CRA W911NF-17-2-0181.}% <-this % stops a space
}

\maketitle

\begin{abstract}
The problem of multi-robot target tracking asks for actively planning the joint motion of robots to track targets. In this paper, we focus on such target tracking problems in adversarial environments, where attacks or failures may deactivate robots' sensors and communications. In contrast to the previous works that consider no attacks or sensing attacks only, we formalize the first robust multi-robot tracking framework that accounts for any fixed numbers of worst-case sensing \textit{and} communication attacks. To secure against such attacks, we design the first robust planning algorithm, named \textit{Robust Active Target Tracking} (\texttt{RATT}), which approximates the communication attacks to \textit{equivalent} sensing attacks and then optimizes against the approximated and original sensing attacks. We show that \texttt{RATT} provides provable suboptimality bounds on the tracking quality for any non-decreasing objective function. Our analysis utilizes the notations of curvature for set functions introduced in combinatorial optimization. In addition, \texttt{RATT} runs in polynomial time and terminates with the same running time as state-of-the-art algorithms for (non-robust) target tracking. Finally, we evaluate \texttt{RATT} with both qualitative and quantitative simulations across various scenarios. In the evaluations, \texttt{RATT} exhibits a tracking quality that is near-optimal and superior to varying non-robust heuristics. We also demonstrate \texttt{RATT}'s superiority and robustness against varying attack models (\textit{e.g.}. worst-case and bounded rational attacks).      
\end{abstract}

\begin{keywords}
Multi-Robot Systems; Active Target Tracking; Robotics in Adversarial Environments; Algorithm Design \& Analysis; Combinatorial Optimization.
\end{keywords}

\IEEEpeerreviewmaketitle

\section{Introduction}
Detecting, localizing, and tracking targets is central to various robotic applications such as surveillance, monitoring, and exploration. Typical examples include: 
\begin{itemize}
\item \textit{Area monitoring:} Localize and track invasive fish (\textit{e.g.}, carp) in a lake;~\cite{tokekar2013tracking}
\item \textit{Search and rescue:} Search and explore a burning building to localize the trapped people inside;~\cite{grocholsky2006cooperative}
\item \textit{Adversarial agent tracking:} Detect and follow adversarial agents that try to escape in an urban environment.~\cite{zengin2007real}
\end{itemize}
Such application scenarios can greatly gain from deploying multiple robots that act as mobile sensors and collaboratively plan their actions. The problem of planning the motion of a team of robots to track targets is known as \textit{multi-robot active target tracking} in the literature~\cite{robin2016multi}. This problem is challenging since the targets can be mobile and their motion model is generally partially known~\cite{spletzer2003dynamic,atanasov2014information,zhou2011multirobot,tokekar2014multi,zhou2018active,zhou2019sensor}. The targets may move adversarially or deliberately release spoofing signals to avoid the detection and tracking of robots~\cite{pierson2016intercepting,mitra2019resilient}. The targets may be indistinguishable and their number may be unknown and varying over time~\cite{dames2017detecting}. Nevertheless, researchers have designed a number of algorithms that achieve near-optimal target tracking while tackling all of the aforementioned challenges~\cite{spletzer2003dynamic,atanasov2014information,zhou2011multirobot,tokekar2014multi,zhou2018active,zhou2019sensor,pierson2016intercepting,mitra2019resilient,dames2017detecting}.  

\begin{figure}
\centering
\includegraphics[width=0.9\columnwidth]{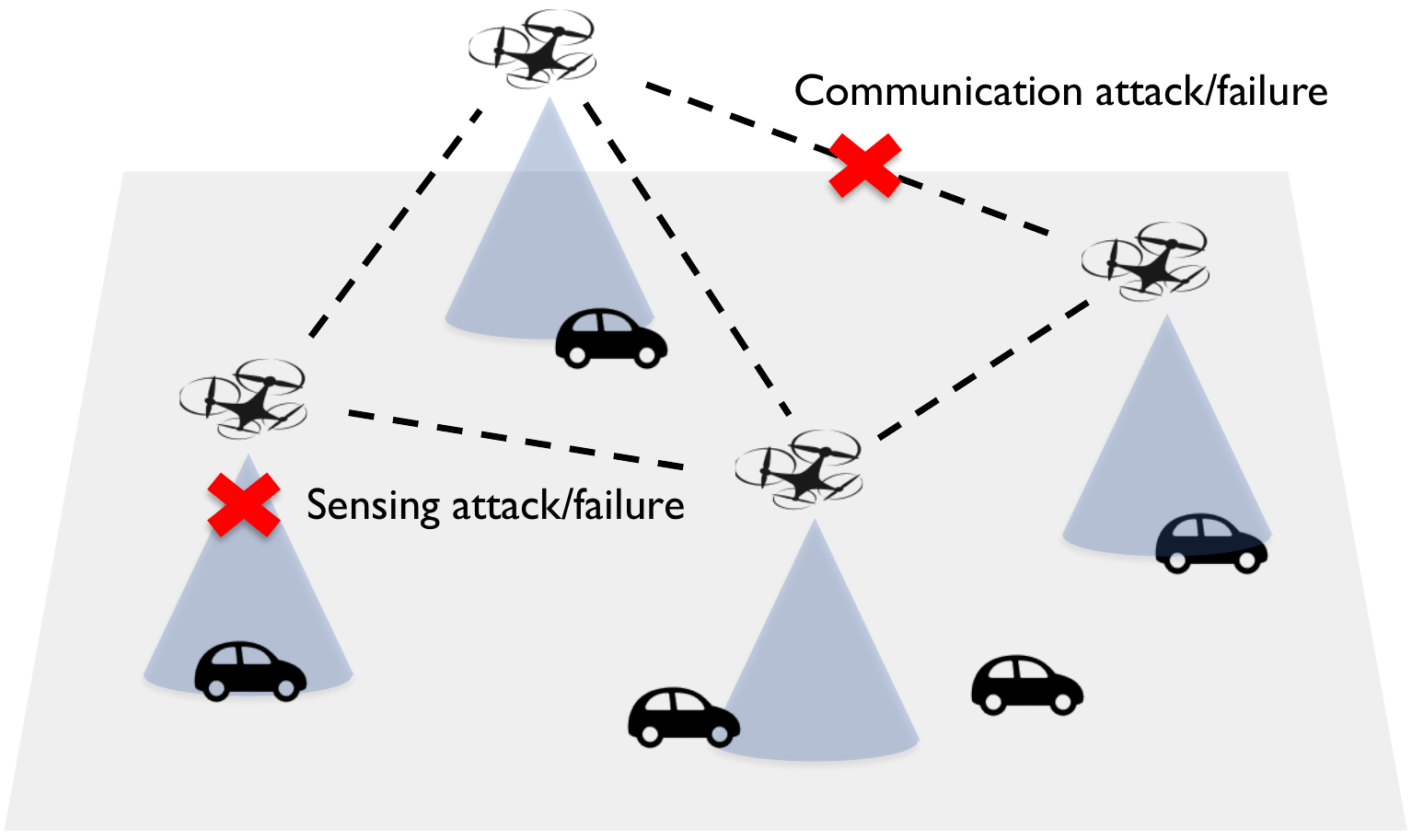}
\caption{Adversarial target tracking: a team of aerial robots is tasked to track multiple ground vehicles in an adversarial environment, where the attacks/failures can disable robots' sensors (\textit{e.g.}, block the camera's field of view) and cut off their communications. 
\label{fig:adv_tracking}}
\end{figure}

However, most robotics studies draw little attention to robustifying target tracking against attacks and failures that may disable robots' sensors (\textit{e.g.}, cameras' field of view may be blocked) and/or communications (\textit{e.g.}, communications may be jammed and disrupted); see also Fig.~\ref{fig:adv_tracking}. For example, 
% \cite{saulnier2017resilient} focuses on multi-robot formation control, instead of target tracking; 
\cite{pierson2016intercepting} focuses on cooperative planning for robots to capture multiple evaders that move adversarially to escape, instead of considering attacks that may disable robots' sensors or communications; \cite{mitra2019resilient} focuses on distributed state estimation against adversaries that broadcast deceptive sensing information to mislead robots, instead of securing against adversaries that may deactivate the sensors of robots; \cite{ramachandran2020resilience} focuses on adaptive resource reconfiguration to recover robots' resources (\textit{e.g.}, sensing capabilities) from failures, instead of robust algorithms that aim to withstand attacks or failures.   

Recently, a number of related studies have investigated robust target tracking against attacks or failures that result in a withdrawal of robots or their sensors. For instance, \cite{zhou2018resilient} proposes a robust coordination algorithm that ensures provably near-optimal tracking even though some robots are attacked and their tracking sensors are disabled. \cite{zhou2020distributed,liu2021distributed} extend the attack-robust algorithm~\cite{zhou2018resilient} to a constrained communication scenario where robots can only communicate locally. In particular, \cite{zhou2020distributed,liu2021distributed} propose distributed robust planning algorithms that utilize respectively the divide-and-conquer and consensus-based mechanisms to achieve robust coordination against robot or sensor attacks. Following this line, \cite{schlotfeldt2021resilient} presents an adaptive attack-robust algorithm against sensing attacks/failures during information acquisition tasks such as target tracking, map exploration, and area surveillance. Notably, all of these studies capitalize on recent algorithmic results on robust combinatorial optimization for set functions against subset attacks~\cite{tzoumas2018resilient}.     

\textbf{Contributions.} In this paper, unlike the aforementioned works, we aim to investigate robust target tracking that accounts for both \textit{sensing and communication} attacks/failures. To this end, we make four key contributions as follows. 

\textbf{1. Problem.} We formalize the problem of robust active target tracking against \textit{worst-case} sensing and communication attacks/failures. The problem asks for jointly optimizing robots' control inputs in the presence of worst-case attacks that could disable robots' sensing and communications. The upper bounds on the numbers of sensing and communication attacks are assumed known and fixed.  To the best of our knowledge, this is the first work to formalize this problem.   

\textbf{2. Solution.} We design the first robust algorithm, named \textit{Robust Active Target Tracking} (\texttt{RATT}), for tackling the problem. In particular, \texttt{RATT} first approximates the number of communication attacks to an \textit{equivalent} number of sensing attacks, and then leverages the algorithmic results in \cite{zhou2018resilient,tzoumas2018resilient} for robust target tracking against total sensing attacks (\textit{i.e.}, the original and approximated sensing attacks).    

\textbf{3. Analysis.} We prove \texttt{RATT} has two main properties. 
\begin{itemize}
    \item \texttt{RATT} provides suboptimality guarantees for any target tracking objective function that is 1) non-decreasing and 2) non-decreasing and submodular. The analysis of \texttt{RATT}'s approximation bounds is based on the notations of curvature (\textit{e.g.}, \textit{curvature}~\cite{conforti1984submodular} and \textit{total curvature}~\cite{sviridenko2017optimal}) from combinatorial optimization. 
    \item \texttt{RATT} ends with the same running time as state-of-the-art algorithms for non-robust target tracking~\cite{fisher1978analysis, tokekar2014multi}) and robust target tracking against sensing attacks only~\cite{zhou2018resilient}. 
\end{itemize}

\textbf{4. Evaluation.} We demonstrate with qualitative and quantitative simulations both the necessity for robust target tracking and \texttt{RATT}'s robustness against sensing and communication attacks. We show \texttt{RATT} exhibits superior performance against worst-case attacks and retains its superiority against non-worst-case attacks, \textit{e.g.}, bounded rational attack, across various scenarios.

Overall, in this paper we go beyond non-robust target tracking~\cite{atanasov2014information,tokekar2014multi,dames2017detecting} by proposing {robust} target tracking; and beyond robust target tracking against sensing attacks only~\cite{zhou2018resilient,zhou2020distributed,schlotfeldt2021resilient} by focusing against both sensing and communication attacks. 

\section{Problem Formulation}\label{sec:problem_formulation}

We formalize the problem of \textit{robust multi-robot active target tracking}. Particularly, the problem asks for choosing control inputs for robots to optimize the team tracking quality despite \textit{sensing and communication} attacks/failures. To this end, we start by defining the notations used in the paper. 

\textbf{Notation.} Calligraphic fonts denote sets (\textit{e.g.}, $\mc{X}$). Given a set $\mc{X}$, $|\mc{X}|$ denotes its cardinality; $2^{\mc{X}}$ denotes its power set. 
Lowercase or uppercase letters (\textit{e.g.}, $x$ or $X$), bold lowercase letters (\textit{e.g.}, $\mb{x}$), and bold uppercase letters (\textit{e.g.}, $\mb{X}$) denote scalars, vectors, and matrices, respectively. $\mathcal{X}\setminus\mathcal{Y}$ denotes the elements in $\mathcal{X}$ not in~$\mathcal{Y}$. 

Next, we first introduce the framework with some basic conventions and then formally define the problem.  

\subsection{Framework} \label{subsec:framework}
\paragraph{Robots} We consider a target tracking scenario where a team of $N$ mobile robots, denoted by $\mathcal{V}=\{1, \cdots, N\}$, is tasked to track multiple targets. Each robot $i\in \mathcal{V}$ has a discrete-time motion model:
\begin{equation}
\mb{x}_{i,t+1}  =  f_i(\mb{x}_{i,t}, \mb{u}_{i,t}), ~\forall i \in \mathcal{V},
    \label{eq:robot_motion}
\end{equation}
where $\mb{x}_i$ denotes the state of robot $i$ and $\mb{u}_i \in \mathcal{U}_{i}$ denotes its control input. $\mathcal{U}_{i}$ is a \textit{finite} set of available control inputs where robot $i$ can choose from.  

\paragraph{Targets} We consider there are $M$ targets, denoted by $\mathcal{T}=\{1,\cdots, M\}$, to be tracked by the robots. Each target $j \in \mathcal{T}$ moves randomly with a stochastic motion model: 
\begin{equation}
\mb{y}_{j,t+1} =  g(\mb{y}_{j,t}) + \mb{w}_{j,t}, ~\forall j \in \mathcal{T},
\label{eq:target_motion}
\end{equation}
where $\mb{y}_j$ denotes the state of target $j$ and $\mb{w}_{j,t}$ denotes the zero-mean white Gaussian process noise with covariance $\mb{Q}_{j,t}$, \textit{i.e.}, $\mb{w}_{j,t} \sim \mathcal{N}(0, \mb{Q}_{j,t})$. We assume $\mb{w}_{j,t}$ to be independent of the process noises of other targets. 

%\mb{y}_{j,t+1} =  \mb{A}_j \mb{y}_{j,t} + \mb{w}_{j,t}, ~\forall j \in \mathcal{T},
%$\mb{A}_j$ is the state transition matrix of target $j$ with suitable dimension,

% Stacking the targets' states into a compact vector $\mb{y}:= [\mb{y}_1; \cdots; \mb{y}_M]$, we have:
% \begin{equation}
% \mb{y}_{t+1} =  \mb{A} \mb{y}_{t} + \mb{w}_{t}, 
% \label{eq:target_motion_compact}
% \end{equation}
% where $\mb{A} := \mb{A}_1 \oplus \mb{A}_2 \oplus \cdots \oplus \mb{A}_M$, and $\mb{w}_t : = [\mb{w}_{1,t}; \mb{w}_{2,t}; \cdots; \mb{w}_{M,t}]$. As discussed, we let $\mb{w}_t \sim \mc{N}(0,\mb{Q}_t)$ with $\mb{Q}_t := \mb{Q}_{1,t} \oplus \mb{Q}_{2,t} \oplus \cdots \oplus \mb{Q}_{M,t}$.

\paragraph{Sensing} We consider a robot $i\in \mathcal{V}$ makes an observation of a target $j \in \mc{T}$ by the following measurement model:
\begin{equation}
 \mb{z}_{i,t}^j = h_i^j(\mb{x}_{i,t}, \mb{y}_{j,t}) + \mb{v}_{i,t}^j(\mb{x}_{i,t}, \mb{y}_{j,t}), ~i\in \mc{V}, j\in \mc{T},
    \label{eq:measure_model}
\end{equation}
where $\mb{z}_{i,t}^j$ denotes the measurement of target $j$ obtained by robot $i$'s on-board sensor and $\mb{v}_{i,t}^j(\mb{x}_{i,t}, \mb{y}_{j,t})$ denotes the zero-mean white Gaussian measurement noise with covariance $\mb{R}(\mb{x}_{i,t}, \mb{y}_{j,t})$, \textit{i.e.}, $\mb{v}_{i,t}^j(\mb{x}_{i,t}, \mb{y}_{j,t}) \sim \mc{N}(0, \mb{R}(\mb{x}_{i,t}, \mb{y}_{j,t}))$. Both the measurement noise $\mb{v}_{i,t}^j$ and sensing function $h_i^j$ depend on the states of the robots and targets, as it naturally holds for, \textit{e.g.}, range and bearing sensors (cf. Section~\ref{sec:simulation}). We assume the measurement noise $\mb{v}_{i,t}^j$ to be independent of the measurement noises of other robots. 

%$\mb{H}_{i,t}^j$ is the corresponding measurement matrix,

% Each robot $i\in \mathcal{V}$ has a limited \textit{sensing range}. The robot $i\in \mathcal{V}$ makes an observation of every target $j$ within its sensing range according to the following observation model:
% \begin{equation}
%  \mb{z}_{i,t}^j = \mb{H}_{i,t}^j \mb{y}_{j,t} + \mb{v}_{i,t}^j(\mb{x}_{i,t}, \mb{y}_{j,t}), ~i\in \mc{V}, j\in \mc{T},
%     \label{eq:measure_model}
% \end{equation}

% Denote the set of robots that observes target $j$ at time step $t$ as $\mc{V}_t^j := \{i_1, i_2, \cdots, i_{n}\} \subseteq \mc{V}$. Then the measurements of each target $j$ collected by $\mc{V}_t^j$ can be written as:
% \begin{equation}
%  \mb{z}_{j,t} = \mb{H}_{j,t} \mb{y}_{j,t} + \mb{v}_{j,t}, ~j \in \mathcal{T},
%     \label{eq:measures_each_tar}
% \end{equation}
% where $\mb{z}_{j,t}:= [\mb{z}_{i_1,t}^j; \mb{z}_{i_2,t}^j; \cdots; \mb{z}_{i_n,t}^j]$ is the collected measurements,  and $\mb{H}_{j,t} := [\mb{H}_{i_1,t}^j; \mb{H}_{i_2,t}^j; \cdots; \mb{H}_{i_n,t}^j]$ is the compact measurement matrix and $\mb{v}_{j,t}:=[\mb{v}_{i_1,t}^j (\mb{x}_{i_1,t}, \mb{y}_{j,t}); \mb{v}_{i_2,t}^j (\mb{x}_{i_2,t}; \cdots; \mb{y}_{j,t}); \mb{v}_{i_n,t}^j (\mb{x}_{i_n,t}, \mb{y}_{j,t})]$ is the compact noise vector. As discussed, we consider $\mb{v}_{j,t} \sim \mc{N}(0, \mb{R}_{j,t})$ with $\mb{R}_{j,t}:= \mb{R}(\mb{x}_{i_1}, \mb{y}_{j,t}) \oplus \mb{R}(\mb{x}_{i_2}, \mb{y}_{j,t}) \oplus \cdots \oplus \mb{R}(\mb{x}_{i_n}, \mb{y}_{j,t})$. 

\paragraph{Communication} We consider centralized communication among robots. That way, the communications among robots can be modeled by an (undirected) fully connected graph $\mc{G}=\{\mathcal{V},\mathcal{E}\}$ with nodes the robots $\mathcal{V}$, and edges the communication links $\mathcal{E}$. Notably, $|\mc{V}|=N$ and $|\mc{E}|=N(N-1)/2$. By centralized communication, all robots can communicate with each other (without communication attacks) to share their sensor measurements. 

\paragraph{Objective} The objective for the robots is to maximize the tracking quality of the targets at the \textit{next} time step by choosing control inputs at the \textit{current} step. In other words, we focus on optimizing the target tracking quality at one
time step ahead. That is because the motions of the targets can be
reliably predicted for the next time step only~\cite{zhou2008optimal,zhou2011multirobot}. In addition, we consider the robots use the Extended Kalman filter (EKF) to estimate the states of the targets. Then, the tracking quality of the robots, denoted henceforth by $\Phi$, depends on the KF's posterior covariance matrix~\cite[Section 4.1]{jawaid2015submodularity}.
% The error covariance $\mb{\Sigma}_{j}$ of each target $j$ can be computed by following the KF's prediction and update steps:
% \begin{align}
% & \color{gray}\text{Prediction:}\color{black} \hspace{2mm} \mb{\Sigma}_{j,t+1|t} =  \mb{A}_j \mb{\Sigma}_{j, t} {\mb{A}}_j^T + \mb{Q}_{j,t}, \label{eq:cov_predict}\\
% & \color{gray}\text{Update:}\color{black} \hspace{6.2mm} \mb{\Sigma}_{j,t+1} = (\mb{\Sigma}_{j,t+1|t}^{-1} + \mb{H}_{j,t}^T \mb{R}_{j,t} \mb{H}_{j,t})^{-1}. \label{eq:cov_update}
% \end{align}
% If a target $j$ is not observed by any robot in $\mathcal{V}$, its error covariance at the next time step is set as the predicted error covariance~(eq.~\eqref{eq:cov_predict}), \textit{i.e.},  $\mb{\Sigma}_{j,t+1} = \mb{\Sigma}_{j,t+1|t}$. The error covariance of all the targets at the next time step can be computed as $\Sigma_{t+1}: = \Sigma_{1, t+1} \oplus \Sigma_{2, t+1} \oplus \cdots \oplus \Sigma_{M, t+1}$. 
% The targets' error covariance is typically used to define the team's tracking quality $\Phi$~\cite[Section 4.1]{jawaid2015submodularity}. 
Examples of $\Phi$ can be appropriate variants of the \textit{mean square error} (\textit{i.e.}, trace of the covariance matrix), the \textit{confidence ellipsoid volume} (\textit{i.e.}, log determinant of  covariance matrix), and the \textit{worst-case error covariance} (\textit{i.e.}, maximum eigenvalue of  covariance matrix)~\cite{jawaid2015submodularity}. Then, the goal of robots is to choose their control inputs $\mb{u}_{\mc{V},t}:=[\mb{u}_{1,t}; \mb{u}_{2,t}; \cdots; \mb{u}_{N,t}]$ at the current time step $t$ (that determine the robots' states at the next time step by~eq.~\eqref{eq:robot_motion}) to maximize the team's tracking quality $\Phi$ at one step ahead. 
 
\paragraph{Attacks} We consider the robots encounter \textit{worst-case} sensing and communication attacks/failures at each time step. The sensing attack on a robot results in the removal of all its sensor measurements to the targets. Then these removed measurements cannot be used to update the estimate of the targets' states by EKF. 

\begin{figure}[t]
\centering{
\subfigure[]
{\includegraphics[width=0.4\columnwidth]{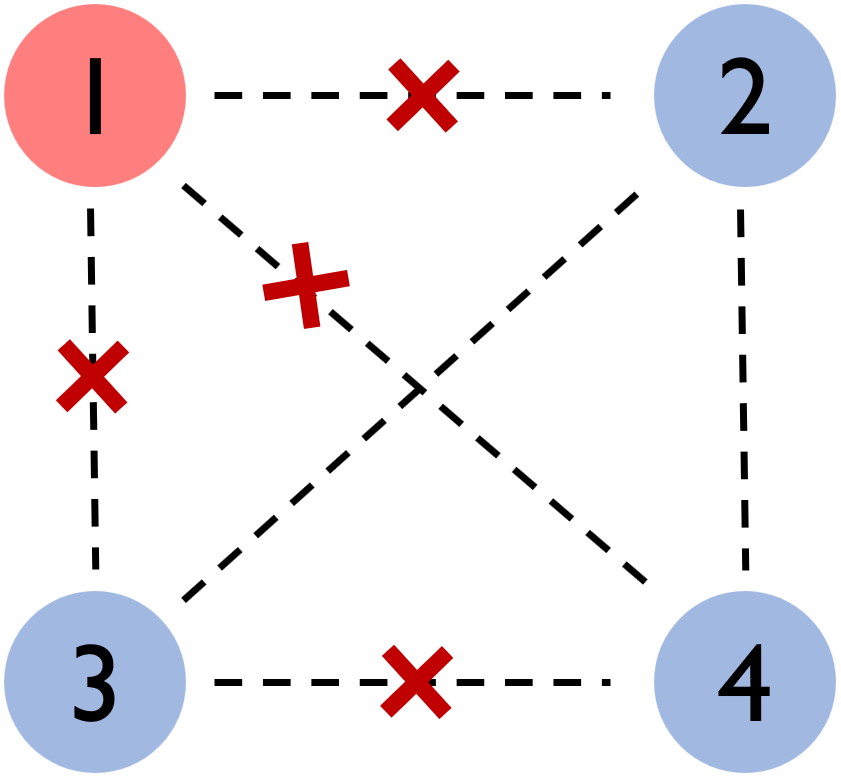}}~~~~~~~
\subfigure[] 
{\includegraphics[width=0.4\columnwidth]{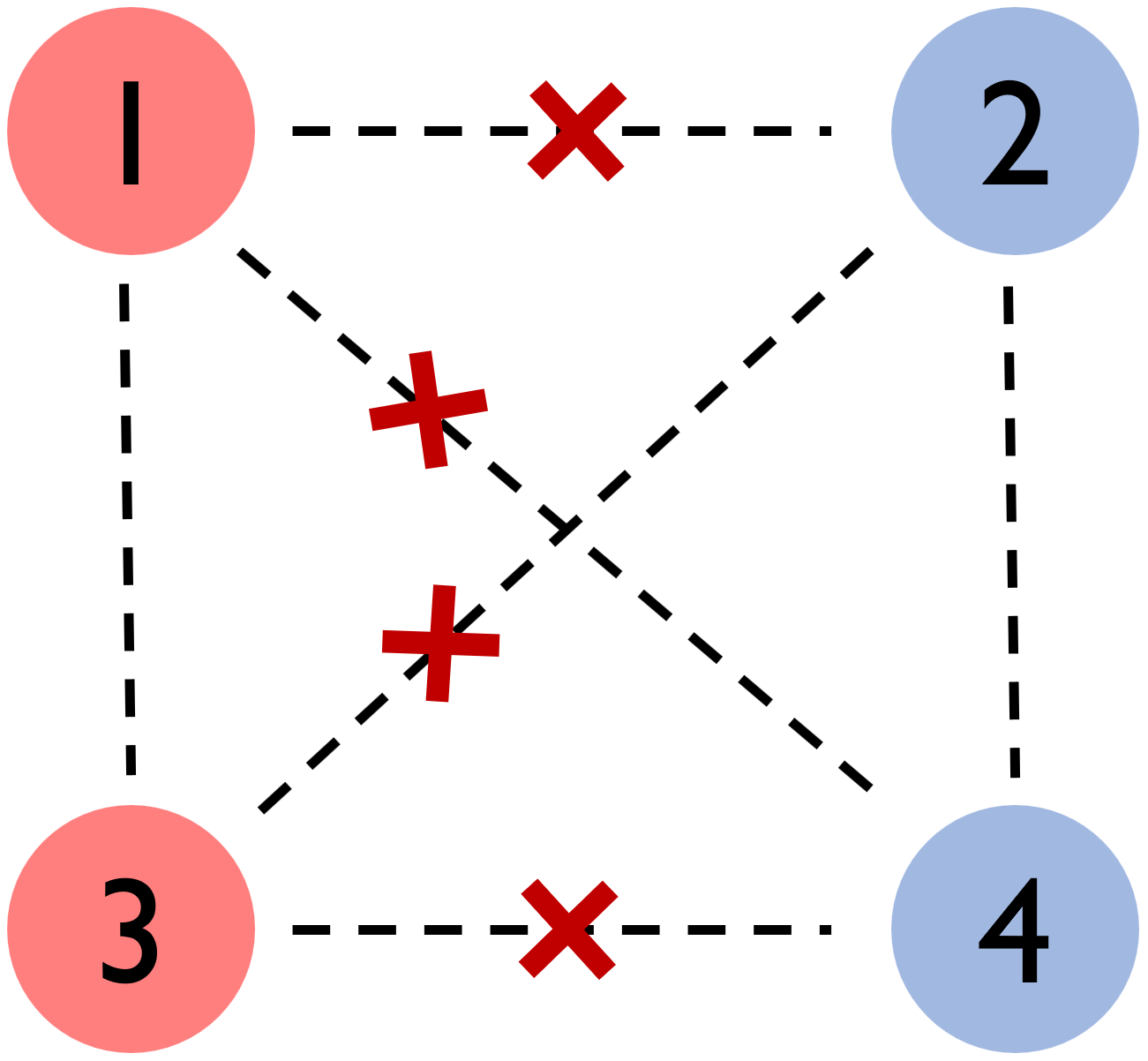}}
\caption{A team of $4$ robots (disks) encounters $4$ communication attacks (red crosses). In subfigure (a), the communication attacks cut off $4$ communication links (dashed lines) and partition the robot team into $2$ isolated subgroups \{1\} and \{2,3,4\}. Thus, the number of robots in the largest subgroup $n_{\max} =3$. The subgroup \{1\} and \{2,3,4\} cannot share sensing measurements. The robots in subgroup \{2,3,4\} can share measurements even though they are not fully connected due to the communication attack between robot 3 and robot 4. In subfigure (b), the $4$ communication attacks partition the robot team into $2$ isolated subgroups (red and blue subgroups, each with two robots), and thus $n_{\max} =2$.  Since at most $2$ robots can communicate (in subfigure (a), at most $3$ robots can communicate), it is more likely the $4$ communication attacks in subfigure (b) are worse for the robot team. 
\label{fig:comm_atk}}
}
\end{figure}

The communication attack cuts off the communication links among robots and thus disables the sharing of sensor measurements. With communication attacks, the robot team may be partitioned into several isolated subgroups (\textit{i.e.}, no communication is available between them, see subgroups \{1\} and \{2,3,4\} in Fig.~\ref{fig:comm_atk}-(a)). Within each subgroup, the robots are connected (not necessarily fully connected, see subgroup \{2,3,4\} in Fig.~\ref{fig:comm_atk}-(a)), and they can share sensor measurements to collectively estimate the states of targets. However, the isolated subgroups cannot utilize the sensor measurements from each other, and thus they update the estimate of the targets' states by EKF individually. In this case, the team's tracking quality is set as the tracking quality of the subgroup that performs the \textit {best}. 

We assume the maximum numbers of sensing and communication attacks/failures at each time step are known, denoted by $\alpha_s$ and $\alpha_c$, respectively. We also assume the attacks at a time step to be effective in this specific time step only. That is, the sensing or communication links attacked at time step $t$ may be active at another time step $t'$. Also, if a robot encounters sensing attack only, it can still act as \textit{information relay node} to communicate with other robots to help them share sensor measurements. 

\subsection{Problem Definition} \label{subsec:prob_def}
We next formally define the main problem in this paper. 
\begin{problem}[\textbf{Robust Multi-Robot Active Target Tracking}] In reference to the framework in Section~\ref{subsec:framework}, consider, a set of mobile robots $\mathcal{V}$, with the motion model~eq.~\eqref{eq:robot_motion}, and with sensing and communication capabilities (Section~\ref{subsec:framework}, c) \& d)); additionally, consider a set of targets $\mathcal{T}$ with motion model~eq.~\eqref{eq:target_motion}; furthermore, consider a multi-target tracking objective function $\Phi$ (Section~\ref{subsec:framework}, e)); finally, consider the maximum numbers of sensing and communication attacks/failures, $\alpha_s$ and $\alpha_c$. For all robots $i\in \mathcal{V}$, find the control inputs $\mb{u_t}$ at the current time step $t$ to maximize the tracking objective $\Phi$ against the worst-case sensing attacks that remove the measurements from (at most) $\alpha_s$ robots and against the worst-case communication attacks that block (at most) $\alpha_c$ communication links among robots at the next time step $t+1$. Formally: 
\begin{align} \label{eqn:resi_target_prob}
\begin{split}
&\max_{\mb{u}_{\mc{V},t}\;\in\;\mc{U}_{\mc{V}}}\;\min_{\mc{A}_{s} \;\subseteq \;\mc{V}, \;\mc{A}_{c} \;\subseteq \;\mc{E}} \Phi(\mc{V}\setminus\mc{A}_{s}, \mc{E}\setminus\mc{A}_{c}) \\
& {\text{s.t.}}\;\; ~ |\mc{A}_{s}| \leq \alpha_s, ~ \alpha_s \leq N, \\
& \hspace{7.5mm} |\mc{A}_{c}| \leq \alpha_c, ~\alpha_c \leq \frac{N(N-1)}{2},
\end{split}
\end{align}
where $\mc{U}_{\mc{V}}\triangleq \bigcup_{i\in\mc{V}} \mc{U}_{i}$ denotes the joint set of available control inputs of all robots; $\mc{A}_{s}$ and $\mc{A}_{c}$ denote the set of robots with sensing removals and the set of communication links (edges) blocked; $\Phi(\mc{V}\setminus\mc{A}_{s}, \mc{E}\setminus\mc{A}_{c})$ is the team's tracking quality with the robots' control input $\mb{u}_{\mc{V},t}$ and the sensing and communication attack sets,  $\mc{A}_{s}$ and $\mc{A}_{c}$. 
\label{prob:rob_tar_tracking}
\end{problem}
Problem~\ref{prob:rob_tar_tracking} captures
% a Stackelberg security game or 
a two-stage sequential game with perfect information between two players~\cite[Chapter 4]{myerson2013game}, namely, the robots (who aim to optimize the team's target tracking quality) and the attacker (who aims to undermine the target tracking quality). The robots 
play first 
% as the leader and start the game 
by selecting the control input $\mb{u}_{\mathcal{V},t}$ at time step $t$ to move to the state $\mb{x}_{t+1}:=[\mb{x}_{1, t+1}; \mb{x}_{2, t+1}; \cdots; \mb{x}_{N, t+1}]$ at the next time step $t+1$ (by eq.~\eqref{eq:robot_motion}). \textit{After observing} $\mb{x}_{t+1}$, the attacker
%, as the follower, 
responds with the worst-case sensing and communication removals $(\mc{A}_{s}, \mc{A}_{c})$.
% Typically, the optimal playing strategies of the robots and the attacker $(\mb{u}_t^\star, (\mathcal{A}_{s}^\star, \mathcal{A}_{c}^\star))$ can be interpreted as a Strong Stackelberg Equilibrium (SSE) where the robots play first and thus can induce the attacker to choose a strategy that is most favorable to them. 

Problem~\ref{prob:rob_tar_tracking} goes beyond the classical objective (\textit{e.g.}, the attacking sets $\mc{A}_{s}, \mc{A}_{c} = \emptyset$) of  multi-robot target tracking, by securing the team performance against worst-case sensing and communication attacks/failures.

\section{Algorithm for Problem~\ref{prob:rob_tar_tracking}}\label{sec:algorithm}
We present the first scalable algorithm (Algorithm~\ref{alg:rob_tar_track}), named \textit{Robust Active Target Tracking} (\texttt{RATT}) for  Problem~\ref{prob:rob_tar_tracking}. \texttt{RATT} sequentially executes two main steps---\textit{communication attack approximation} (\texttt{RATT}'s line~\ref{line:ratt_call_caa}) and \textit{robust maximization against sensing attacks} (\texttt{RATT}'s lines~\ref{line:ratt_compute_totalatk}-\ref{line:ratt_forend_allbaits}). In the first step, \texttt{RATT} utilizes a subroutine (Algorithm~\ref{alg:comm_appro}, named $\texttt{CAA}$) to approximate the number of the communication attacks $\alpha_c$ to a certain number of sensing attacks $\alpha_{c,s}$. In the second step, \texttt{RATT} capitalizes on the algorithm results in~\cite{zhou2018resilient,tzoumas2018resilient} to maximize the team's tracking quality against the \textit{total} number of sensing attacks (\textit{i.e.}, the original number of sensing attacks $\alpha_s$ and the approximated number of sensing attacks $\alpha_{c,s}$). We describe \texttt{RATT}'s two steps in more detail below.

\begin{algorithm}[t]
\caption{Robust Active Target Tracking (\texttt{RATT}).}
\begin{algorithmic}[1]
\REQUIRE  Current time step $t$; set of $N$ robots $\mc{V}$; each robot $i$'s dynamics $f_i$ including current state $\mb{x}_{i,t}$; each robot $i$'s available control inputs $\mathcal{U}_{i}$; set of $M$ targets $\mc{T}$; each target $j$'s dynamics including state transition matrix $\mb{A}_j$, current covariance $\mb{Q}_{j,t}$ of the motion noise, current state estimate $\hat{\mb{y}}_{j,t}$, and current covariance estimate $\mb{\Sigma}_{j, t}$; measurement model of each robot $i$ to each target $j$ including current measurement matrix $\mb{H}_{i,t}^j$ and current covariance $\mb{R}(\mb{x}_{i,t}, \mb{y}_{j,t})$ of the measurement noise; team's target tracking objective function $\Phi$; number of sensing attacks $\alpha_s$; number of communication attacks $\alpha_c$.
\ENSURE Control inputs $\mb{u}_{i,t}$ for all robots $i \in \mc{V}$.  
\medskip

\STATE Compute $\alpha_{c,s}$ by calling $\texttt{CAA}(N, \alpha_c)$; \label{line:ratt_call_caa}
\STATE Set $\alpha := \alpha_s + \alpha_{c,s}$; \label{line:ratt_compute_totalatk}
\IF{$\alpha < N$} \label{line:ratt_ifless_alpha} 
    %  \color{gray}~~~// Generate \textit{bait} set for $\alpha$ robots \nonumber //\color{black} 
    \FOR{$i \in \mc{V}$} \label{line:ratt_forst_idvmax}
    \STATE $\Phi^{\star}_{\{i\}, 0} \triangleq \underset{\mb{u}_{i,t}~\in~\mc{U}_{i}}{\max}\;\Phi(\{i\}\setminus \emptyset)$; \label{line:ratt_idv_robot_max}
    \ENDFOR \label{line:ratt_forend_idxmax}
    \STATE Sort out a subset $\mc{V}_{b}$ of $\alpha$ robots, such that $\Phi^{\star}_{\{i\}, 0} \geq \Phi^{\star}_{\{i'\}, 0}, ~\forall i\in \mc{V}_{b}$ and $\forall i' \in \mc{V} \setminus \mc{V}_{b}$; \label{line:ratt_baits}
    \FOR{$i\in \mc{V}_{b}$} \label{line:ratt_forst_baitassign}
        \STATE $\mb{u}_{i,t} = \text{arg}     \underset{\mb{u}_{i,t}~\in~\mc{U}_{i}}{\max}\Phi(\{i\}\setminus \emptyset)$; \label{line:ratt_baitassign}
        % {$\{$\color{gray} $\alpha$ baits selection\color{black}$\}$}
    \ENDFOR \label{line:ratt_forend_baitassign}
    \STATE $\mc{V}_{g} \leftarrow \emptyset$; \label{line:ratt_gre_emptyset}
    \WHILE{$\mc{V} \setminus (\mc{V}_{b} \cup \mc{V}_{g}) \ne \emptyset$}
        \STATE Find robot $i'\in \mc{V} \setminus (\mc{V}_{b} \cup \mc{V}_{g})$ and its control input $\mb{u}_{i', t}$ such that $\mb{u}_{i', t} \in \text{arg} \underset{\mb{u}_{v,t}~\in~\mc{U}_{\mc{V} \setminus (\mc{V}_{b} \cup \mc{V}_{g})}}{\max}\Phi((\mc{V}_{g} \cup \{v\})\setminus \emptyset)$ \\\hspace{5cm}$-  \Phi(\mc{V}_{g}\setminus \emptyset)$;  \label{line:ratt_gre_margin}
        \STATE $\mc{V}_{g} \leftarrow \mc{V}_{g} \cup \{i'\}$; \label{line:ratt_gre_update}
    \ENDWHILE
\ELSE \label{line:ratt_iflarge_alpha} 
    \FOR{$i\in \mc{V}$} \label{line:ratt_forst_allbaits}
        \STATE $\mb{u}_{i,t} = \text{arg} \underset{\mb{u}_{i,t}~\in~\mc{U}_{i}}{\max}\Phi(\{i\}\setminus \emptyset)$; \label{line:ratt_allbaits_assign}
    \ENDFOR \label{line:ratt_forend_allbaits}
\ENDIF
\end{algorithmic}\label{alg:rob_tar_track}
\end{algorithm}

%%%%%%%%%%%%%%%%%%%%%%%%%%%%%%% a detailed version
\begin{algorithm}[t]
\caption{Communication Attack Approximation (\texttt{CAA}).}
\begin{algorithmic}[1]
\REQUIRE 
Number of robots $N$ in the team; number of worst-case communication attacks $\alpha_c$.
\ENSURE Approximated number of sensing attacks $\alpha_{c,s}$. 
\STATE Initialize $n_{\max} = 0$;  \label{line:caa_ini_nmax}
\STATE Compute $e_r = \frac{N(N-1)}{2} - \alpha_c$; \label{line:caa_remain_edges}
\FOR{each $n \in [1, 2, \cdots, N]$} \label{line:caa_forstart}
    \STATE Compute $q_n = \lfloor{\frac{N}{n}}\rfloor$, $r_n = N -nq_n$; and \\ 
    $\overline{e}_n = q_n \frac{n(n-1)}{2} + \frac{r_n(r_n-1)}{2}$; \label{line:caa_e_max_n}
% \STATE Compute $\underline{e}_{n+1} = \frac{n(n+1)}{2}$; \label{line:caa_e_min_n1}
    \IF{$e_r \leq \overline{e}_n$} \label{line:caa_ifstart}
        \STATE Set $n_{\max} = n$, stop \textbf{for} loop; \label{line:caa_max_subgroup}
    \ENDIF \label{line:endif}
%  {$\{$\color{gray}3rd round communication\color{black}$\}$} 
\ENDFOR \label{line:caa_endfor} 
\STATE \textbf{return} $\alpha_{c,s}= N- n_{\max}$. \label{line:caa_output}
\end{algorithmic}
\label{alg:comm_appro}
\end{algorithm}
%%%%%%%%%%%%%%%%%%%%%%%%%%%%%%

\subsection{Communication attack approximation  (Step-1 of \texttt{RATT})}\label{subsec:subalg_comm_app}
We present the first step of \texttt{RATT}, \textit{communication attack approximation} (\texttt{RATT}'s line~\ref{line:ratt_call_caa}, that calls \texttt{CAA}, with pseudo-code presented in Algorithm~\ref{alg:comm_appro}). \texttt{CAA} approximates the number of communication attacks to a certain number of sensing attacks for the robot team. The intuition behind \texttt{CAA} is from the \textit{equivalent} consequence of communication attacks and sensing attacks. Recall that the sensing attack removes sensor measurements and the communication attack disables measurement sharing among isolated subgroups (Section~\ref{subsec:framework}-(f)). Thus, if one subgroup cannot communicate with the other subgroups to utilize their sensor measurements, from the perspective of this specific subgroup, it is equivalent to the case that the communications are not attacked but the other subgroups encounter sensing attacks and lose their sensor measurements. For example, in Fig.~\ref{fig:comm_atk}-(a), the $4$ communication attacks result in two isolated subgroups \{2,3,4\} and \{1\}, for subgroup \{2,3,4\}, it cannot utilize the sensor measurements from subgroup \{1\}, which can be seen as the scenario that subgroup \{2,3,4\} can communicate with subgroup \{1\} but the measurements of subgroup \{1\} are removed by sensing attacks. Thus, for subgroup \{2,3,4\}, these $4$ communication attacks are equivalent to $1$ sensing attack on robot 1. Likewise, for subgroup \{1\}, these $4$ communication attacks are equivalent to $3$ sensing attacks on robots 2, 3, 4. Therefore, for each subgroup, its approximated number of sensing attacks can be set as the number of robots in the other isolated subgroups.

\begin{figure*}[th!]
\centering{
\subfigure[$\overline{e}_1 = 0$, $n=1$]
{\includegraphics[width=0.35\columnwidth]{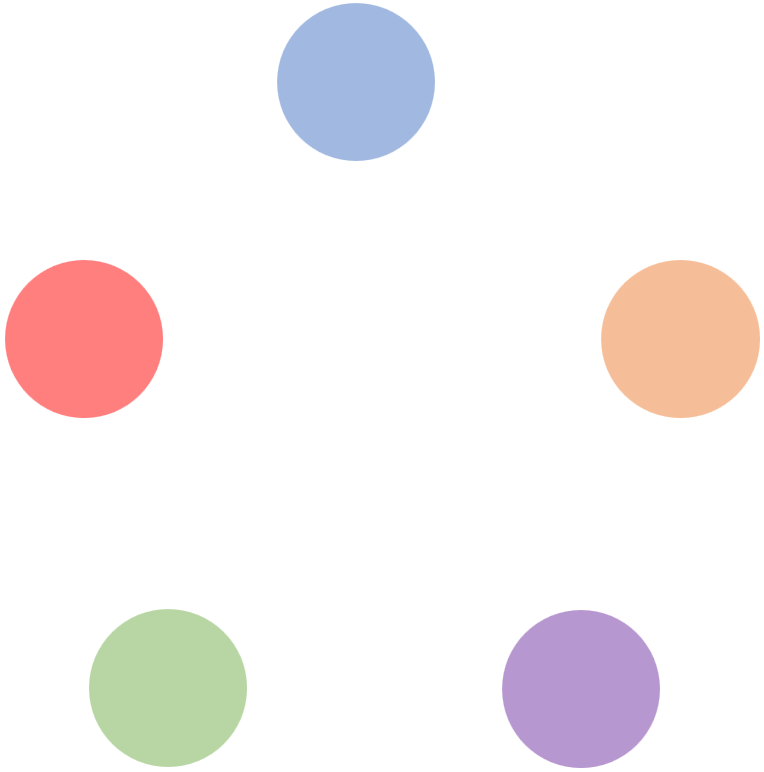}}~~~~
\subfigure[$\overline{e}_2 = 2$, $n=2$] 
{\includegraphics[width=0.35\columnwidth]{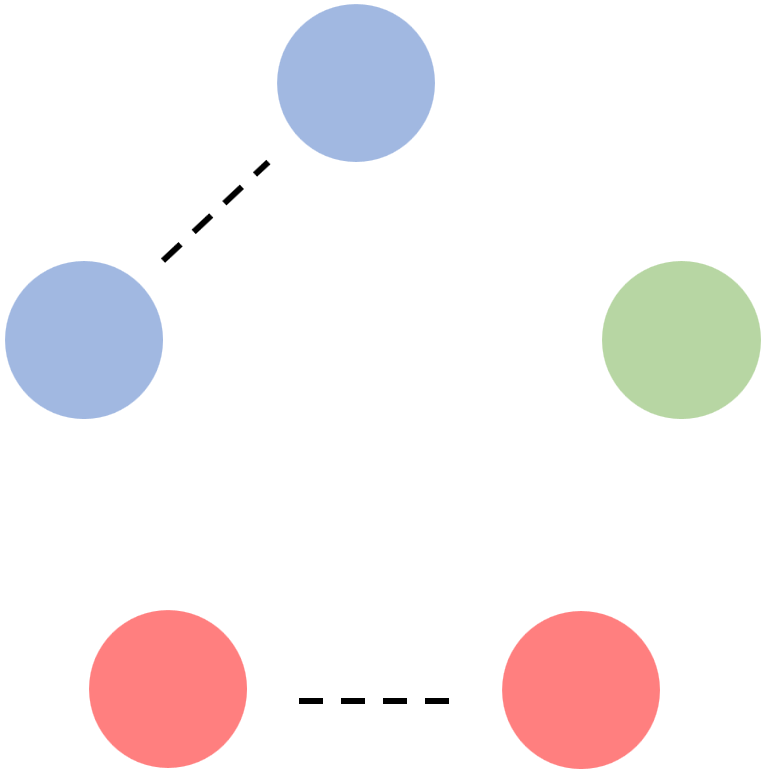}}~~~~
\subfigure[$\overline{e}_3 = 4$, $n =3$] 
{\includegraphics[width=0.35\columnwidth]{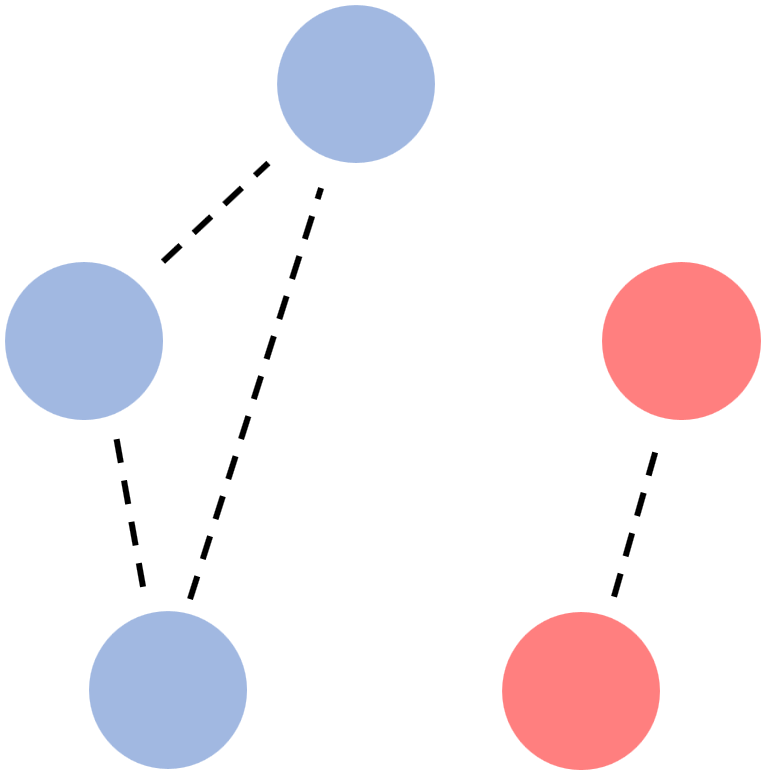}}~~~~
\subfigure[$\overline{e}_4 = 6$, $n=4$] 
{\includegraphics[width=0.35\columnwidth]{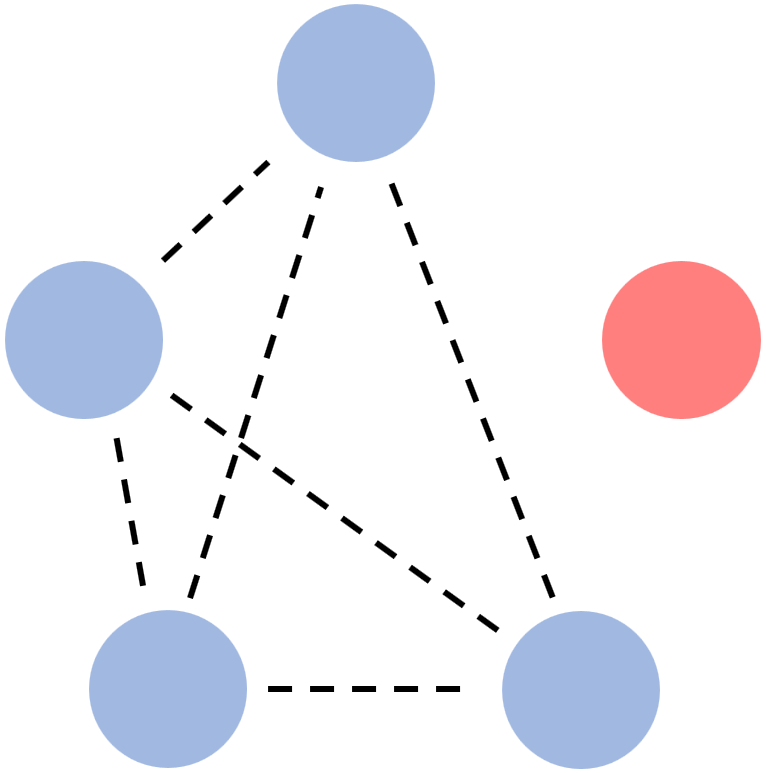}}~~~~
\subfigure[$\overline{e}_5 = 10$, $n=5$] 
{\includegraphics[width=0.35\columnwidth]{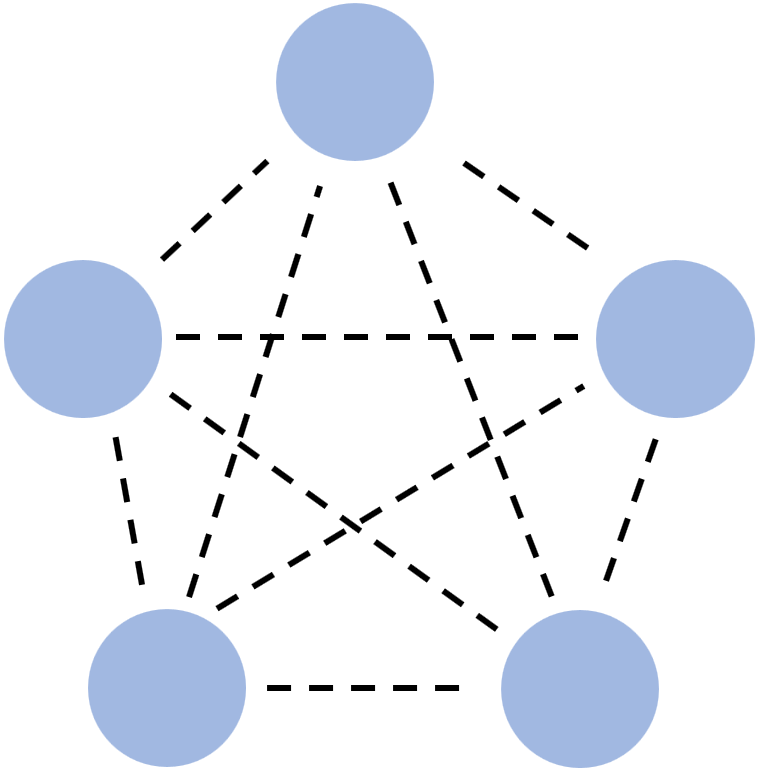}}
\caption{Qualitative description of \texttt{CAA} computing the maximal number of edges $\overline{e}_n, n \in [1, 2, \cdots, N]$ for a team of five robots ($N=5$) by eq.~\eqref{eq:bar_en}. The disks represent the robots. The robots within the same subgroup have the same color. The dashed lines show the available edges between robots.  Based on $\overline{e}_n$, the number of the robots in the largest subgroup $n_{\max}$ and the approximated number of sensing attacks $\alpha_{c,s}$ with any number of worst-case communication attacks $\alpha_c$ can be computed. For example, when $\alpha_c = 7$, one has the remaining number of edges $e_r = N(N-1)/2 - \alpha_c = 10 -7 =3$. Then \texttt{CAA} checks $n \in [1, 2, 3, 4, 5]$ in ascending order, and stops at $n=3$ since $e_r \leq \overline{e}_3$ holds for the first time. Thus, \texttt{CAA} sets $n_{\max} = 3$ and computes $\alpha_{c,s} = N - n_{\max} =2$.
\label{fig:caa_show}}
}
\end{figure*}

Notably, the approximated number of sensing attacks can be varying for different subgroups, which brings the problem of deciding the approximated number of sensing attacks for the entire robot team, denoted henceforth by $\alpha_{c,s}$. To address this issue, we focus on the largest subgroup (\textit{i.e.}, the subgroup with the most number of robots among all subgroups).\footnote{If there exist multiple largest subgroups, we can randomly pick one.} That is because the team's tracking quality equals the tracking quality of the subgroup that performs the best (Section~\ref{subsec:framework}-(f)), and the subgroup with more robots is likely to perform better (since more measurements are available).\footnote{Notably, \texttt{RATT} does not require the assumption that the largest subgroup must perform better than the others for providing hard guarantees against worst-case communication attacks (Section~\ref{subsec:perform_ana}).} In other words, for computing the team's tracking quality, it is more likely that the largest subgroup will be used and other smaller subgroups will be ignored. Therefore, $\alpha_{c,s}$ can be conjectured as the approximated number of sensing attacks in the largest subgroup. Then, as mentioned above, $\alpha_{c,s}$ can be computed as the number of robots that are in other smaller subgroups. We denote the number of robots in the largest subgroup as $n_{\max}$. Clearly, $n_{\max}+\alpha_{c,s} =N$. Then the critical point is to determine $n_{\max}$ after $\alpha_c$ worst-case communication attacks. 

Also, since a smaller subgroup is more likely to perform worse than a larger subgroup, the \textit{worst-case} communication attacks tend to separate the robot team \textit{evenly} into more, smaller subgroups. In this way, the communication attacks are more presumably to compromise the coordination between the robots to the largest extent. More specifically, the worst-case communication attacks aim to make $n_{\max}$ as small as possible.  Considering the intent of the worst-case communication attacks, we present \texttt{CAA} (Algorithm~\ref{alg:comm_appro}) that efficiently computes $n_{\max}$ and $\alpha_{c,s}$ as follows: 
\begin{itemize}
    \item \texttt{CAA} initializes the number of robots in the largest subgroup $n_{\max}$ as zero (Algorithm~\ref{alg:comm_appro}, line~\ref{line:caa_ini_nmax}). 
    \item  \texttt{CAA} computes the number of the remaining (unattacked) edges (communication links)  $e_r$ the robot team has after $\alpha_c$ communication attacks (Algorithm~\ref{alg:comm_appro}, line~\ref{line:caa_remain_edges}). 
% which corresponds to the case without communication attacks ($\alpha_c = 0$) and all robots are fully connected. 
    \item \texttt{CAA} determines $n_{\max}$ by iterating over all possible candidates $n \in [1, 2, \cdots, N]$ in \textit{ascending} order (Algorithm~\ref{alg:comm_appro}, lines~\ref{line:caa_forstart}-\ref{line:caa_endfor}). For each iteration (each $n$),
    \begin{enumerate}
        \item  \texttt{CAA} first computes the \textit{maximal} number of available (unattacked) edges $\overline{e}_n$ the robot team \textit{can} have when the number of robots in the largest subgroup is $n$ (Algorithm~\ref{alg:comm_appro}, line~\ref{line:caa_e_max_n}, cf. Fig.~\ref{fig:caa_show}). Specifically, it computes the number of the largest subgroups (each has $n$ robots) as $$
        q_n = \lfloor{\frac{N}{n}}\rfloor,$$
        the remaining robots in the smaller subgroup as $$r_n = N -nq_n,$$ and $\overline{e}_n$ as the total number of available edges in the largest subgroups and the smaller subgroup, \textit{i.e.}, 
    \begin{equation}
    e_{n} = q_n \frac{n(n-1)}{2} + \frac{r_n(r_n-1)}{2}.\footnote{The way \texttt{CAA} computes $\overline{e}_n$ is because the communication attacks execute in the worst-case and tend to separate the robot team evenly into more, smaller subgroups.}
    \label{eq:bar_en}
    \end{equation}
    \item \texttt{CAA} then compares the number of the remaining edges $e_r$ with $\overline{e}_n$. If $e_r$ is less than or equal to $\overline{e}_n$  (Algorithm~\ref{alg:comm_appro}, line~\ref{line:caa_ifstart}), \texttt{CAA}  sets $n_{\max}$ as $n$ and stops the loop (Algorithm~\ref{alg:comm_appro}, line~\ref{line:caa_max_subgroup}) to make sure that $e_r \leq \overline{e}_n$ holds for the first time.\footnote{Notably, by definition of $\overline{e}_n$, a larger $n$ gives a larger $\overline{e}_n$. Thus, if $e_r \leq \overline{e}_n$ holds, $e_r \leq \overline{e}_{n'}$ also holds with any $n'> n$.} 
    \end{enumerate}
    \item With $n_{\max}$ obtained, \texttt{CAA} computes the approximated number of sensing attacks $\alpha_{c,s}$ as $N-n_{\max}$. 
\end{itemize}
An illustrative example of how \texttt{CAA} computes $\overline{e}_n$, $n_{\max}$, and $\alpha_{c,s}$ is shown in Fig.~\ref{fig:caa_show}. 

\subsection{Robust maximization against sensing attacks (Step-2 of \texttt{RATT}}\label{subsec:robust_alg}
We now present the second step of \texttt{RATT}: \textit{robust maximization against total sensing attacks} (\texttt{RATT}'s lines~\ref{line:ratt_compute_totalatk}-\ref{line:ratt_forend_allbaits}). This step follows the traits of the robust algorithm in~\cite{zhou2018resilient,tzoumas2018resilient}, which applies a two-step procedure to select robust actions for the robots against a fixed number of attacks (\textit{e.g.}, $\alpha'$). We recall here the two steps of the robust algorithm. First, it tries to conjecture the worst-case $\alpha'$ attacks to the robots, and toward this end, builds a \textit{bait} action set to attract these attacks. Particularly, the bait set contains the \textit{top} $\alpha'$ individual actions that contribute the most to the team performance from $\alpha'$ robots. In the second step, the robust algorithm greedily selects actions for the remaining $N-\alpha'$ robots by the standard greedy algorithm~\cite[Section~2]{fisher1978analysis}. On this basis, \texttt{RATT}'s second step is as follows:
\begin{itemize}
    \item \texttt{RATT} computes the total number of sensing attacks $\alpha$ as the summation of the original number of sensing attacks $\alpha_s$ and the approximated number of sensing attacked $\alpha_{c,s}$ (Algorithm~\ref{alg:rob_tar_track}, line~\ref{line:ratt_compute_totalatk}). 
    \item If the total number of sensing attacks $\alpha$ is less than the number of robots $N$ (Algorithm~\ref{alg:rob_tar_track}, line~\ref{line:ratt_ifless_alpha}), \texttt{RATT} follows a two-step process, similar to the robust algorithm~\cite{zhou2018resilient,tzoumas2018resilient}, to choose robots' control inputs against the total $\alpha$ sensing attacks.  
    \begin{enumerate}
        \item First, \texttt{RATT} selects the bait control inputs for $\alpha$ robots to attract the total $\alpha$ sensing attacks (Algorithm~\ref{alg:rob_tar_track}, lines~\ref{line:ratt_forst_idvmax}-\ref{line:ratt_baitassign}). Specifically, it first computes the maximum \textit{individual} tracking quality $\Phi^{\star}_{\{i\}, 0}$\footnote{In Algorithm~\ref{alg:rob_tar_track}, the number of communication attacks is approximated by a certain number of sensing attacks. Thus, for clarify, we omit here the dependence of tracking quality $\Phi$ on the communication links $\mc{E}$.} that each robot $i$ can attain by executing one of its control inputs, assuming no attacks (Algorithm~\ref{alg:rob_tar_track}, lines~\ref{line:ratt_forst_idvmax}-\ref{line:ratt_idv_robot_max}). It then finds the \textit{top} $\alpha$ robots $\mc{V}_b$ whose maximum individual tracking qualities are among the top $\alpha$ ones (Algorithm~\ref{alg:rob_tar_track}, line~\ref{line:ratt_baits}). Finally, for each robot in $\mc{V}_b$, \texttt{RATT} decides its control input as the one that achieves its maximum individual tracking quality (Algorithm~\ref{alg:rob_tar_track}, line~\ref{line:ratt_baitassign}).  
        \item Second, \texttt{RATT} greedily selects control inputs for the rest of the $N-\alpha$ robots $\mc{V}\setminus \mc{V}_b$ by following the standard greedy algorithm~\cite[Section~2]{fisher1978analysis} (Algorithm~\ref{alg:rob_tar_track}, lines~\ref{line:ratt_gre_emptyset}-\ref{line:ratt_gre_update}). Particularly, at each round, it finds a robot $i'\in \mc{V}\setminus \mc{V}_b$ and selects for it a control input which provides the maximum marginal gain in the \textit{joint} tracking quality (Algorithm~\ref{alg:rob_tar_track}, line~\ref{line:ratt_gre_margin}).     
    \end{enumerate}
    \item Instead, if the total number of sensing attacks $\alpha$ is larger than or equal to the number of robots $N$\footnote{This could happen since the approximated number of sensing attacks $\alpha_{c,s}$ can be larger than $N-\alpha_s$ when the number of communication attacks $\alpha_c$ is large enough.} (Algorithm~\ref{alg:rob_tar_track}, lines~\ref{line:ratt_iflarge_alpha}), \texttt{RATT} conjectures all the robots' sensing will be attacked. To this end, it lets each robot do its individual best; \textit{i.e.}, it chooses for each robot the control input that achieves the maximum individual tracking quality (Algorithm~\ref{alg:rob_tar_track}, lines~\ref{line:ratt_forst_allbaits}-\ref{line:ratt_forend_allbaits}).   
\end{itemize}

All in all, \texttt{RATT} decides the control inputs for all robots by sequentially executing $\texttt{CAA}$ and the robust maximization. In addition, \texttt{RATT} is valid for any number of sensing and communication attacks. 

\section{Performance Analysis of Algorithm~\ref{prob:rob_tar_tracking}}~\label{sec:analysis}

In this section, we quantify \texttt{RATT}'s performance, by bounding its approximation performance and running time. To this end, we treat objective function $\Phi$ as a set function (\textit{i.e.}, defined on the set of robots) and leverage the properties of set functions such as monotonicity, submodularity, and curvatures. We start with reviewing these useful properties and then analyze the performance of \texttt{RATT}. 

\subsection{Properties of Set Functions}\label{subsec:curv}
\vspace{3pt}\noindent\textbf{Normalized set function~\cite{nemhauser1978analysis}:} Consider a finite, discrete set $\mc{X}$, a set function $\phi: 2^{\mathcal{X}} \mapsto \mathbb{R}$ is normalized if $\phi(\emptyset) = 0$.

\vspace{3pt}\noindent\textbf{Non-negative set function~\cite{nemhauser1978analysis}:} Consider a finite, discrete set $\mc{X}$, a set function $\phi: 2^{\mathcal{X}} \mapsto \mathbb{R}$ is non-negative if $\phi(\mc{Y}) \geq 0$ for all $\mc{Y} \subseteq \mc{X}$.  

\vspace{3pt}\noindent\textbf{Non-decreasing set function~{\cite{nemhauser1978analysis}}:} Consider a finite, discrete set $\mc{X}$, a set function $\phi: 2^{\mathcal{X}} \mapsto \mathbb{R}$ is non-decreasing if for any sets $\mc{Y}\subseteq \mc{Y}'\subseteq \mathcal{X}$, $\phi(\mc{Y})\leq \phi(\mc{Y}')$. 

\vspace{3pt}\noindent\textbf{Modularity~\cite{nemhauser1978analysis}:} Consider a finite, discrete set $\mc{X}$, a set function $\phi: 2^{\mathcal{X}} \mapsto \mathbb{R}$ is modular if $\phi(\mc{Y}) = \sum_{y\in\mc{Y}}$ for all $\mc{Y} \subseteq \mc{X}$. 

That is, for a modular function $\phi$, the contributions of $\mc{X}$'s elements to function $\phi$ are additive. Hence, modularity also implies $\phi(\{x\} \cup \mc{Y}) - \phi(\mc{Y}) = \phi(x)$ for all $\mc{Y} \subseteq \mc{X}$ and $x\in \mc{X} \setminus \mc{Y}$. 

\vspace{3pt}\noindent\textbf{Submodularity~{\cite[Proposition 2.1]{nemhauser1978analysis}}:} Consider a finite ground set $\mc{X}$, a set function $\phi: 2^{\mathcal{X}} \mapsto \mathbb{R}$ is submodular if $\phi(\mc{Y}\cup \{x\})- \phi(\mc{Y}) \geq \phi(\mc{Y}'\cup \{x\})- \phi(\mc{Y}')$ for all $\mc{Y}\subseteq \mc{Y}'\subseteq \mathcal{X}$ and $x\in \mathcal{X}$. 

Hence, if $\phi$ is submodular, the marginal contribution $\phi(\mc{Y}\cup \{x\})- \phi(\mc{Y})$  diminishes as $\mc{Y}$ grows, for all $x \in \mc{X}$. For a submodular function $\phi$, the contributions of $\mc{X}$'s elements to function $\phi$ is subadditive, contrary to $\phi$ being modular. Specifically, consider, without loss of generality, $\phi$ to be non-negative, then submodularity implies $\phi(\{x\} \cup \mc{Y}) - \phi(\mc{Y}) \leq \phi(x)$. Therefore, $x$'s contribution to $\phi$ diminishes in the presence of $\mc{Y}$. 

\vspace{3pt}\noindent\textbf{Curvature~{\cite{conforti1984submodular}}:} Consider a finite, discrete set $\mc{V}$, and a non-decreasing submodular function $\phi:2^{\mathcal{X}}\mapsto\mathbb{R}$ such that, without loss of generality, $\phi({x})\neq 0$, for all $x \in \mathcal{X}$. Then, $\phi$'s \textit{curvature} is defined as 
\begin{equation}\label{eq:curvature}
k_\phi\triangleq 1-\min_{x\in\mathcal{X}}\frac{\phi(\mathcal{X})-\phi(\mathcal{X}\setminus\{x\})}{\phi({x})}.
\end{equation}

The definition of curvature implies $k_\phi \in [0, 1]$. Particularly, if $k_\phi=0$, then $\phi(\mathcal{X})-\phi(\mathcal{X}\setminus\{x\}) = \phi(x)$, for all $x\in\mc{V}$ (\textit{i.e.}, $\phi$ is modular). On the other hand, if $k_\phi=1$, then there exist $x\in\mathcal{X}$ such that $\phi(\mathcal{X})-\phi(\mathcal{X}\setminus\{x\})= 0$ (\textit{i.e.}, $x$ has no additional contribution in the presence of $\mathcal{X}\setminus\{x\}$). Therefore, the curvature $k_\phi$ measures how far $\phi$ is from modularity: a smaller $k_\phi$ means $\phi$ is closer to being modular and a larger $k_\phi$ means $\phi$ is farther from being modular (\textit{i.e.}, is more submodular). 

\vspace{3pt}\noindent\textbf{Total curvature~{\cite[Section 8]{sviridenko2017optimal}}:} Consider a finite, discrete set $\mc{X}$, and a non-decreasing set function $\phi:2^{\mathcal{X}}\mapsto\mathbb{R}$. Then, $\phi$'s \textit{total curvature} is defined as
\begin{equation}\label{eq:total_curvature}
c_\phi\triangleq 1-\min_{x\in\mathcal{X}}\;\min_{\mc{Y}, \mc{Y'}\subseteq \mc{X}\setminus x}\; \frac{\phi(\mathcal{Y} \cup \{x\})-\phi(\mathcal{Y})}{\phi(\mathcal{Y'} \cup \{x\})-\phi(\mathcal{Y'})}.
\end{equation}

The definition of total curvature implies $c_\phi \in [0, 1]$. Particularly, if $c_\phi=0$, then $\phi$ is modular. Instead, if $c_\phi=1$, then eq.~\eqref{eq:total_curvature} implies $\phi$ is non-decreasing, which is already assumed in the definition of total curvature. If $\phi$ is submodular, then total curvature becomes curvature, \textit{i.e.}, $c_\phi = k_\phi$. 

\subsection{Performance Analysis for \texttt{RATT}} \label{subsec:perform_ana}
We present \texttt{RATT}'s suboptimality bounds on its approximation performance and the upper bound on its running time. To this end, we use the following notations:
\begin{itemize}

    %  \item $\Phi_{\mc{V}, \alpha_s}^\star$ is the optimal value of Problem~\ref{prob:rob_tar_tracking} with sensing attacks only (\textit{i.e.}, $\alpha_s \geq 0, \alpha_c = 0$): \\
    % $$\Phi_{\mc{V}, \alpha_s}^\star \triangleq \\ \max_{\mb{u}_{\mc{V},t}\in\mc{U}_{\mc{V},t}}\min_{\mc{A}_{s} \subseteq \mc{V}} \Phi(\mc{V}\setminus\mc{A}_{s}, \mc{E}\setminus \emptyset).$$
    % $\mathcal{A}_{s}^\star$ is an optimal removal of the measurements from $\alpha_s$ robots per Problem~\ref{prob:rob_tar_tracking} with sensing attacks only: \\
    % $$\mathcal{A}_{s}^\star \triangleq ~\text{arg} \min_{\mc{A}_{s} \;\subseteq \; \mc{V}} \Phi(\mc{V}\setminus\mc{A}_{s}, \mc{E}\setminus \emptyset).$$  
    
    %  \item $\Phi_{\mc{E},\alpha_c}^\star$ is the optimal value of Problem~\ref{prob:rob_tar_tracking} with communication attacks only (\textit{i.e.}, $\alpha_s = 0, \alpha_c \geq 0$): \\
    % $$\Phi_{\mc{E},\alpha_c}^\star \triangleq \\ \max_{\mb{u}_{\mc{V},t}\in\mc{U}_{\mc{V},t}}\min_{\mc{A}_{c} \subseteq \mc{E}} \Phi(\mc{V}\setminus \emptyset, \mc{E}\setminus\mc{A}_{c}).$$
    %  $\mathcal{A}_{c}^\star$ is an optimal removal of $\alpha_c$ communication links among $N$ robots per Problem~\ref{prob:rob_tar_tracking} with communication attacks only: \\
    % $$\mathcal{A}_{c}^\star \triangleq ~\text{arg} \min_{\mc{A}_{c} \;\subseteq \;\mc{E}} \Phi(\mc{V}\setminus \emptyset, \mc{E}\setminus\mc{A}_{c}).$$  

    \item $\Phi_{\mc{V}, \alpha_s, \mc{E},\alpha_c}^\star$ is the optimal value of Problem~\ref{prob:rob_tar_tracking}: \\
    $$\Phi_{\mc{V}, \alpha_s, \mc{E},\alpha_c}^\star \triangleq \\ \max_{\mb{u}_{\mc{V},t}\in\mc{U}_{\mc{V}}}\min_{\mc{A}_{s} \subseteq \mc{V}, \mc{A}_{c} \subseteq \mc{E}} \Phi(\mc{V}\setminus\mc{A}_{s}, \mc{E}\setminus\mc{A}_{c});$$
    $(\mathcal{A}_{s}^\star, \mathcal{A}_{c}^\star)$ is an optimal removal of the measurements from $\alpha_s$ robots and of $\alpha_c$ communication links among $N$ robots per Problem~\ref{prob:rob_tar_tracking}: \\
    $$(\mathcal{A}_{s}^\star, \mathcal{A}_{c}^\star) \triangleq ~\text{arg} \min_{\mc{A}_{s} \;\subseteq \; \mc{V}, \;\mc{A}_{c} \;\subseteq \;\mc{E}} \Phi(\mc{V}\setminus\mc{A}_{s}, \mc{E}\setminus\mc{A}_{c}).$$  
\end{itemize}

\begin{theorem}[Approximation bound]~\label{thm:ratt_appro} 
\texttt{RATT} returns control inputs $\mb{u}_{\mc{V},t}$ such that 
% if there exist both sensing and communication attacks, \textit{i.e.}, $\alpha_s \geq 0, \alpha_c \geq 0$, and  
\begin{enumerate}
    \item if $\Phi : 2^{\mc{V}, \mc{E}}\mapsto\mathbb{R}$ is non-decreasing, and without loss of generality, normalized and non-negative, then
    \begin{equation}~\label{eqn:appro_non_descrease}
        \frac{\Phi(\mc{V}\setminus\mc{A}_{s}^\star, \mc{E}\setminus\mc{A}_{c}^\star)}{\Phi_{\mc{V}, \alpha_s, \mc{E},\alpha_c}^\star} \geq \min[(1-c_\Phi)^3, \frac{(1-c_\Phi)^2}{\alpha_{c,s}}]; 
    \end{equation}
    
    \item if, additionally, $\Phi$ is submodular, then 
    \begin{equation}~\label{eqn:appro_submodular}
        \frac{\Phi(\mc{V}\setminus\mc{A}_{s}^\star, \mc{E}\setminus\mc{A}_{c}^\star)}{\Phi_{\mc{V}, \alpha_s, \mc{E},\alpha_c}^\star} \geq 
        \min [\frac{1-k_\Phi}{1+k_\Phi}, \frac{1-k_\Phi}{\alpha_{c,s}}].
    \end{equation}    
\end{enumerate}
\end{theorem}

\begin{theorem}[Running time]~\label{thm:ratt_run_time}
\texttt{RATT} runs in $O(|\mc{U}_{\mc{V}}|^2)$ time. 
\end{theorem}

We provide the proofs for Theorem~\ref{thm:ratt_appro} and Theorem~\ref{thm:ratt_run_time} in the appendix.

In Theorem~\ref{thm:ratt_appro}, the approximation bounds in eqs.~\eqref{eqn:appro_non_descrease}, \eqref{eqn:appro_submodular} compare \texttt{RATT}'s selection $\mb{u}_{\mc{V},t}$ against an optimal selection of control inputs that attains the optimal value $\Phi_{\mc{V}, \alpha_s, \mc{E},\alpha_c}^\star$ for Problem~\ref{prob:rob_tar_tracking}. In particular, eqs.~\eqref{eqn:appro_non_descrease}, \eqref{eqn:appro_submodular} imply that \texttt{RATT} guarantees a value for Problem~\ref{prob:rob_tar_tracking} which can be near-optimal for 1) non-decreasing and 2) non-decreasing and submodular functions $\Phi$. For example, when $c_\Phi < 1$ or $k_\Phi < 1$, \texttt{RATT}'s selection is near-optimal in the sense that the approximation bounds are nonzero. Functions with $c_\Phi < 1$ include the mean square error (\textit{e.g.}, the trace of a Kalman filter's covariance matrix)~\cite{chamon2017mean}; Functions with $k_\Phi < 1$ include $\log \det$ of positive-definite matrices (\textit{e.g.}, $\log \det$ of a Kalman filter's covariance matrix)~\cite{jawaid2015submodularity,sharma2015greedy}. In addition, when $\alpha_{c,s} \to 0$ (\textit{e.g.}, no communication attacks $\alpha_c \to 0$), the bound in  eq.~\eqref{eqn:appro_non_descrease} becomes $(1-c_\Phi)^3$ and bound in \eqref{eqn:appro_submodular} becomes $(1-k_\Phi)/(1+k_\Phi)$, which parallels the results in \cite[Theorem 1]{zhou2018resilient} and \cite[Theorem 1]{tzoumas2018resilient} with sensing attacks only.

Furthermore, if $\alpha_{c,s} \to 0$ and $c_\Phi =0$ (or $k_\Phi=0$), \texttt{RATT} becomes exact (\textit{i.e.}, the bounds in eqs.~\eqref{eqn:appro_non_descrease}, \eqref{eqn:appro_submodular} are 1). More broadly, the bound in eq.~\eqref{eqn:appro_non_descrease} increases as $c_\Phi$ decreases. Likewise, the bound in eq.~\eqref{eqn:appro_submodular} increases as $k_\Phi$ decreases. Hence, these curvature-depended bounds make a first step towards classifying the classes of 1) non-decreasing and 2) non-decreasing and submodular functions 
into the category for which Problem~\ref{prob:rob_tar_tracking} can be approximated well (low curvature functions), and the category for which it cannot (high curvature functions). Notably, although it is known for the problem of set function maximization (without attacks) that the bound $1-c_\Phi$ is tight~\cite[Theorem 8.6]{sviridenko2017optimal}, the tightness of $(1-c_\Phi)^3$ in eq.~\eqref{eqn:appro_non_descrease} for Problem~\ref{prob:rob_tar_tracking} is an open problem. 

Theorem~\ref{thm:ratt_run_time} implies that even though \texttt{RATT} goes beyond the objective of set function maximization without attacks or with sensing attacking only, its running time has the same order as that of state-of-the-art non-robust algorithms~\cite{fisher1978analysis,tokekar2014multi} and robust algorithms against sensing attacks only~\cite{zhou2018resilient,tzoumas2018resilient}. In particular, same as these algorithms, \texttt{RATT} terminates with $O(|\mc{U}_{\mc{V}}|^2)$ evaluations of the objective function~$\Phi$. 

\begin{remark}
A myopic algorithm that selects bait control input for each robot independently of all other robots (in contrast to \texttt{RATT}, whose robust maximization step accounts for the control inputs of the other robots during the greedy selection), attains better approximation bounds. That is because, the myopic algorithm is the same as \texttt{RATT} when $N \leq \alpha$ and, in this instance, \texttt{RATT} achieves $(1-c_\Phi)^2$ for non-decreasing functions and $1-k_\Phi$ for non-decreasing and submodular functions (cf. eqs.~\eqref{eq:nd2} and \eqref{eq:sub2} in the appendix). However, being exclusively myopic, this algorithm has worse practical performance than that of the robust maximization (cf. \cite[Remark 1]{zhou2020distributed}).
\end{remark}

%%%%%%%%%%%%%%%%%%%%%%%%%%%%%%%%%%%%%%%%%%
\begin{figure*}
    \centering
	\subfigure[\texttt{RATT} without  1 sensing attack]{\includegraphics[width=0.50\columnwidth]{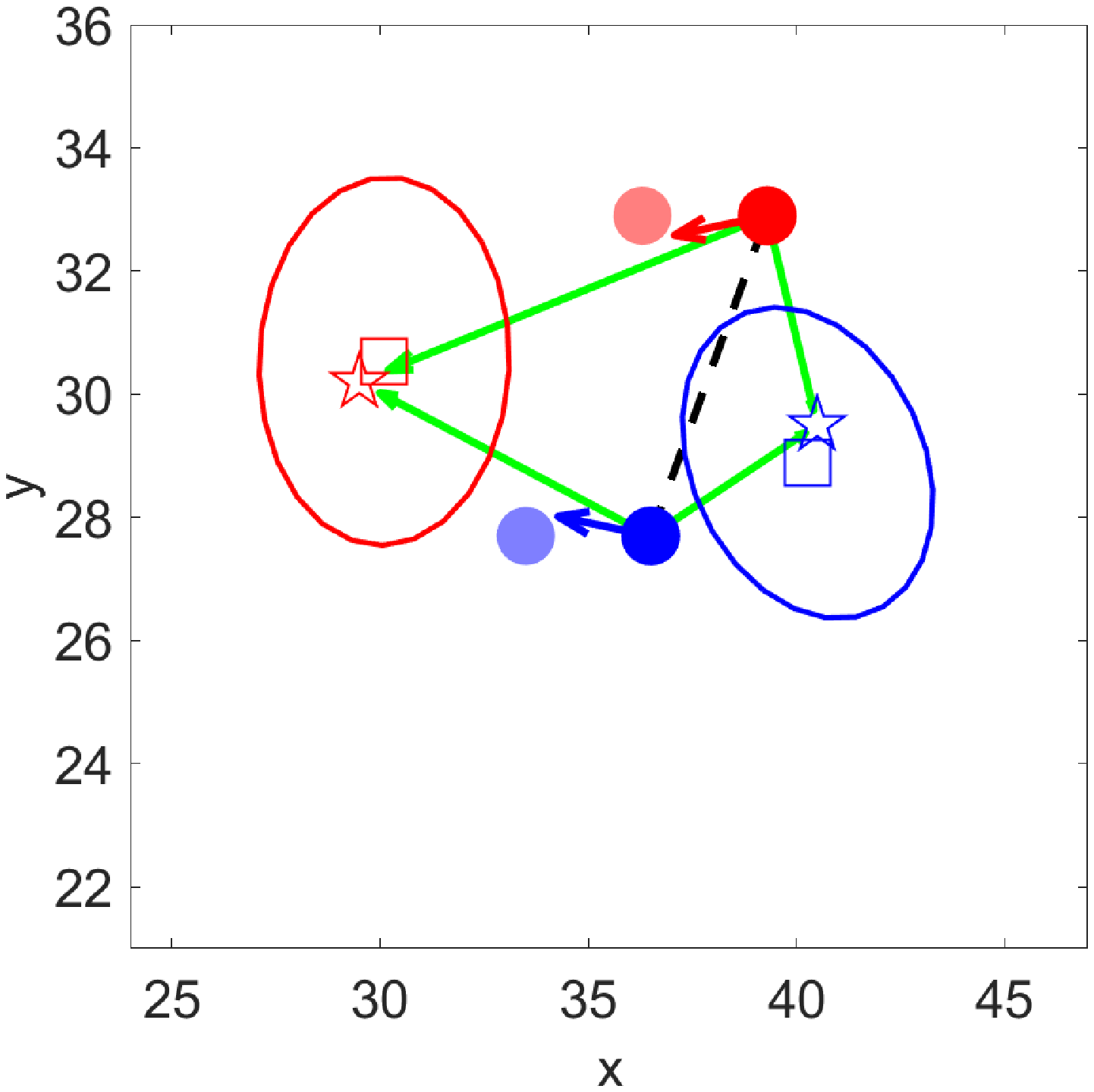}}
	\subfigure[\texttt{NR-OPT} without  1 sensing attack]{\includegraphics[width=0.50\columnwidth]{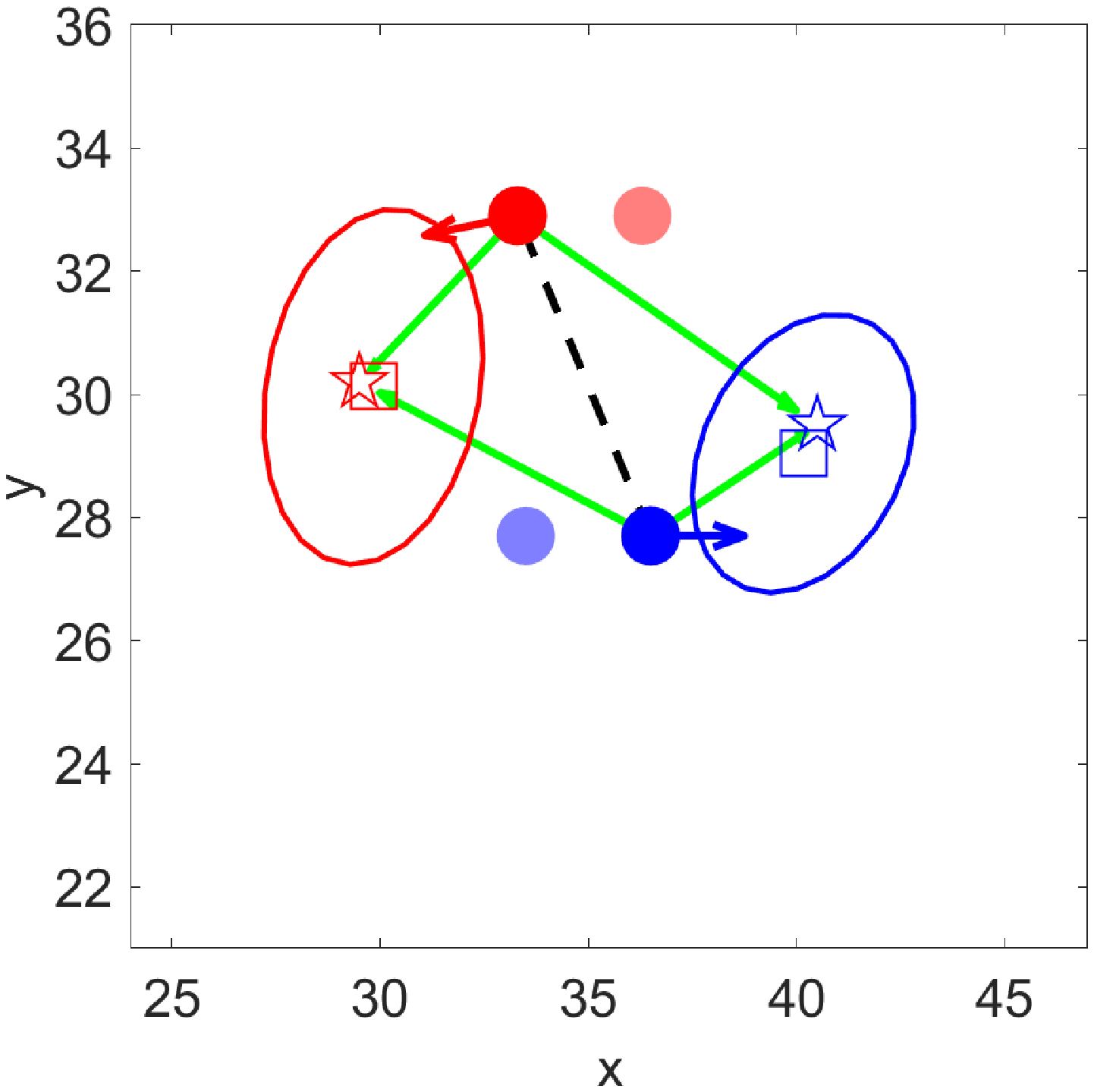}}
	\subfigure[\texttt{RATT} with 1 sensing attack] {\includegraphics[width=0.50\columnwidth]{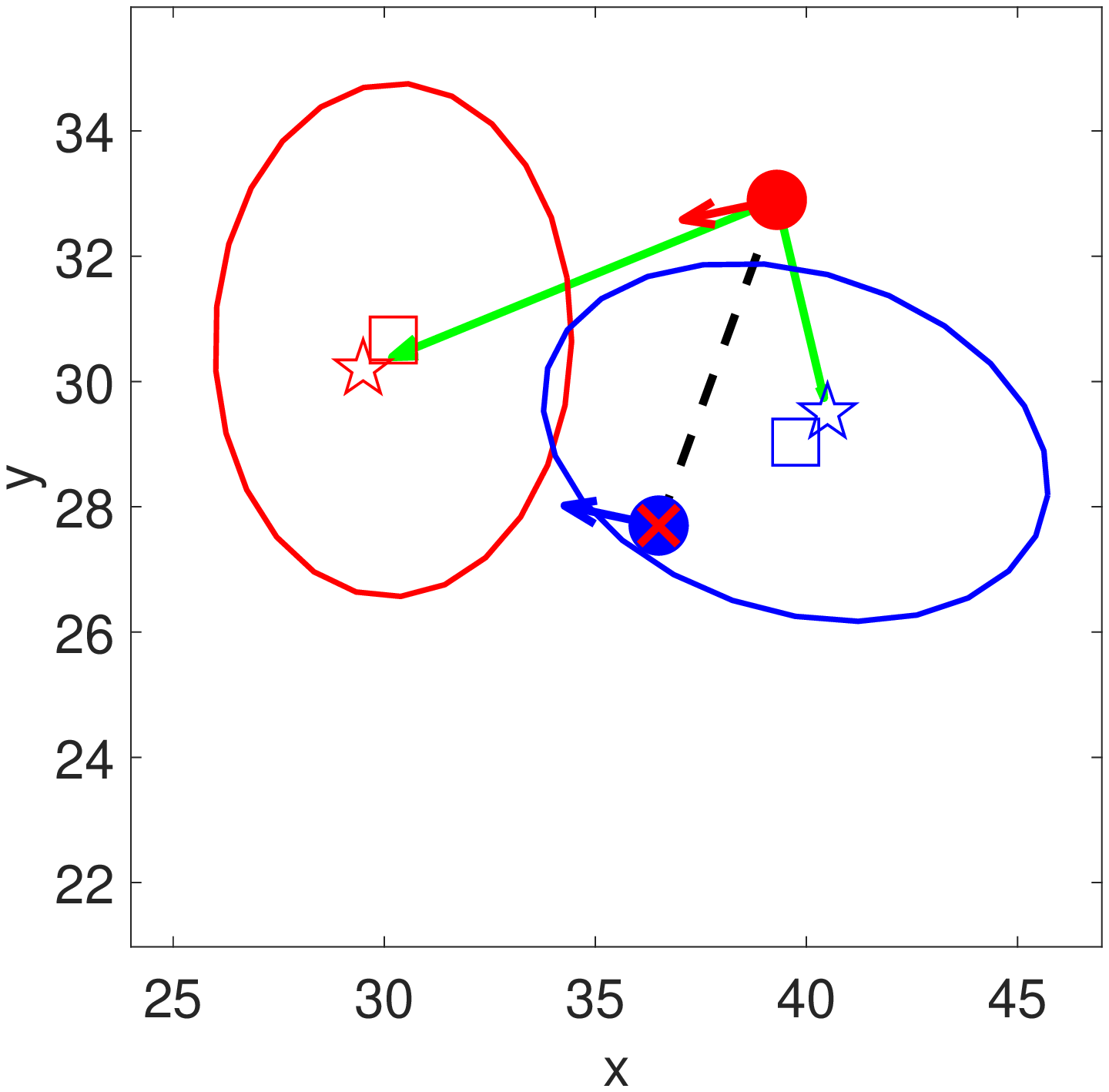}}
	\subfigure[\texttt{NR-OPT} with 1 sensing attack] {\includegraphics[width=0.50\columnwidth]{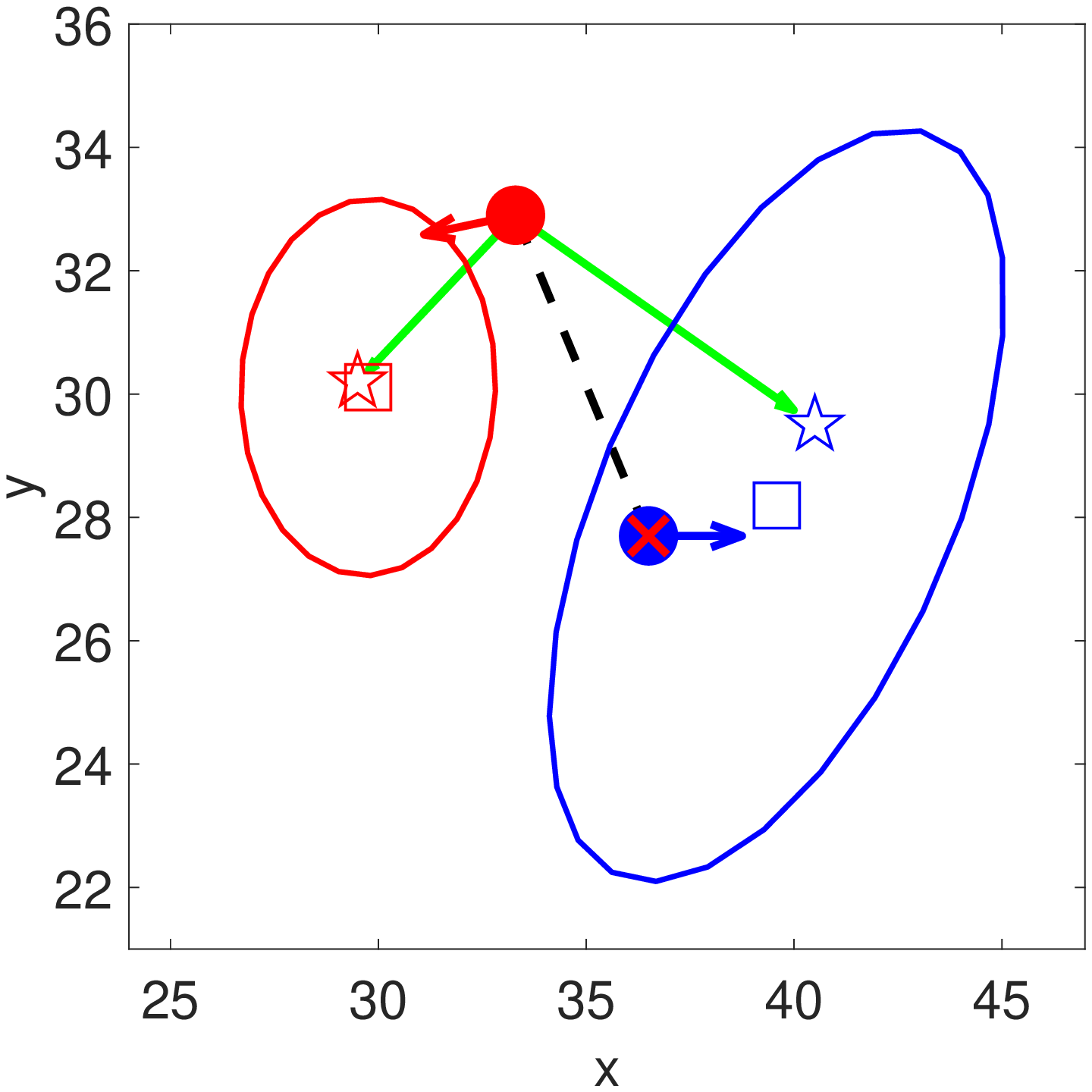}}
\caption{Comparison of 2 robots tracking 2 targets by \texttt{RATT} and \texttt{NR-OPT} without and with1 sensing attack. The transparent dot and the opaque dot represent the positions of the robot at the current time step and at one time step ahead, respectively. The arrow on the robot represents its heading. The pentagram, square, and ellipse indicate the true position, estimated mean position, and covariance of the target, respectively. The green arrow lines represent the sensing links from robots to targets. The black dashed line represents the communication link between robots. The red cross on a robot indicates the sensing attack on the robot. 
% In (a) \& (b), without the attack, the trace of the covariance \& MSE (per target) by \texttt{RATT} and \texttt{NR-OPT} are $1.3937$ \&  $0.7721$ and $1.1466$ \& $0.2955$. In (c) \& (d), with the attack, the trace of the covariance \& MSE by \texttt{RATT} and \texttt{NR-OPT} are $1.3937$ \&  $0.7721$
%  and   $1.1466$ \& $0.2955$. 
}
\label{fig:ql_1sen_atk}
\end{figure*}

%%%%%%%%%%%%%%%%%%%%%%%%%%%%%%%%%%%%%%%%%%
\begin{figure*}
    \centering
	\subfigure[\texttt{RATT} without  1 communication attack]{\includegraphics[width=0.50\columnwidth]{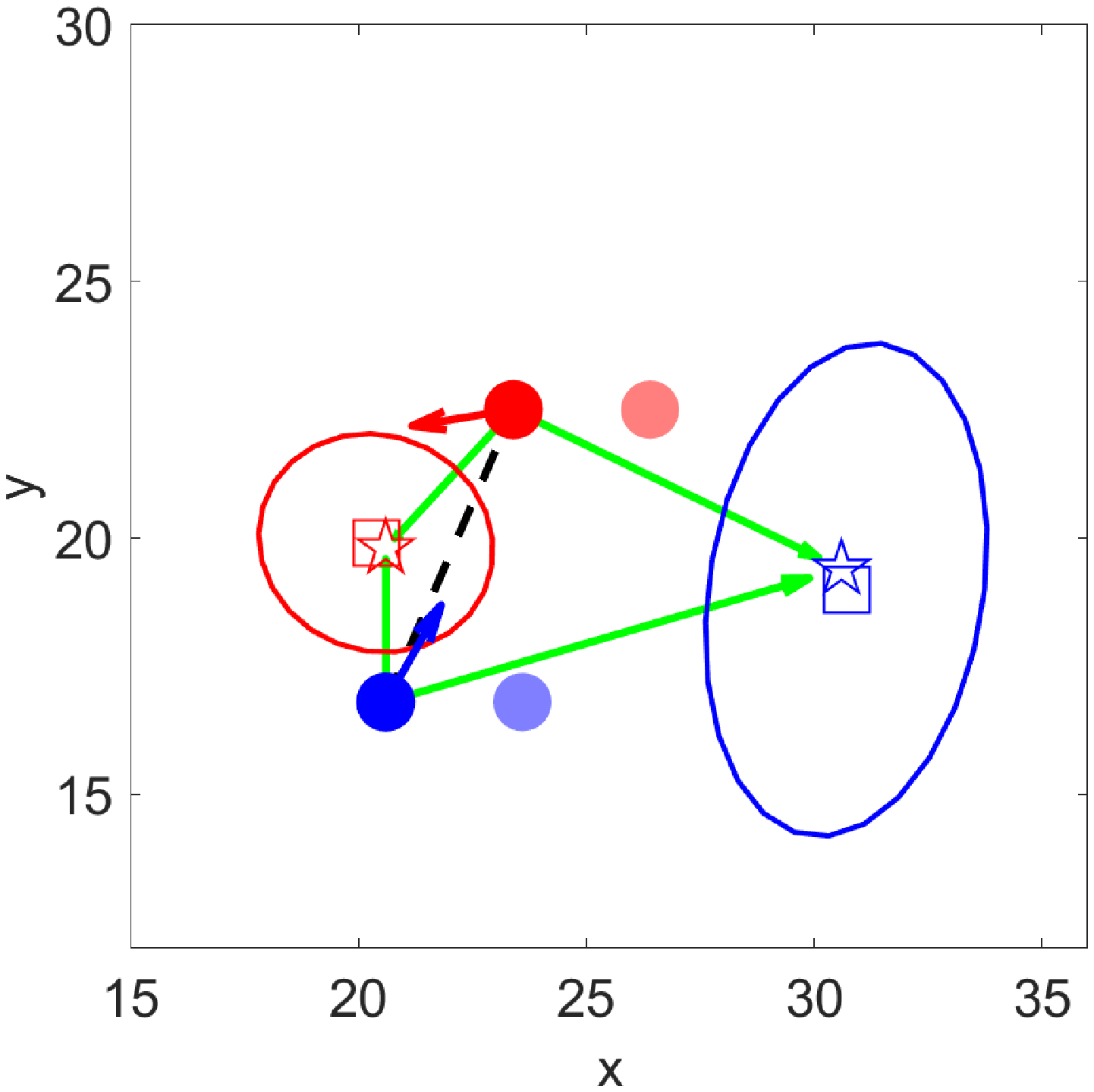}}
	\subfigure[\texttt{NR-OPT} without  1 communication attack]{\includegraphics[width=0.50\columnwidth]{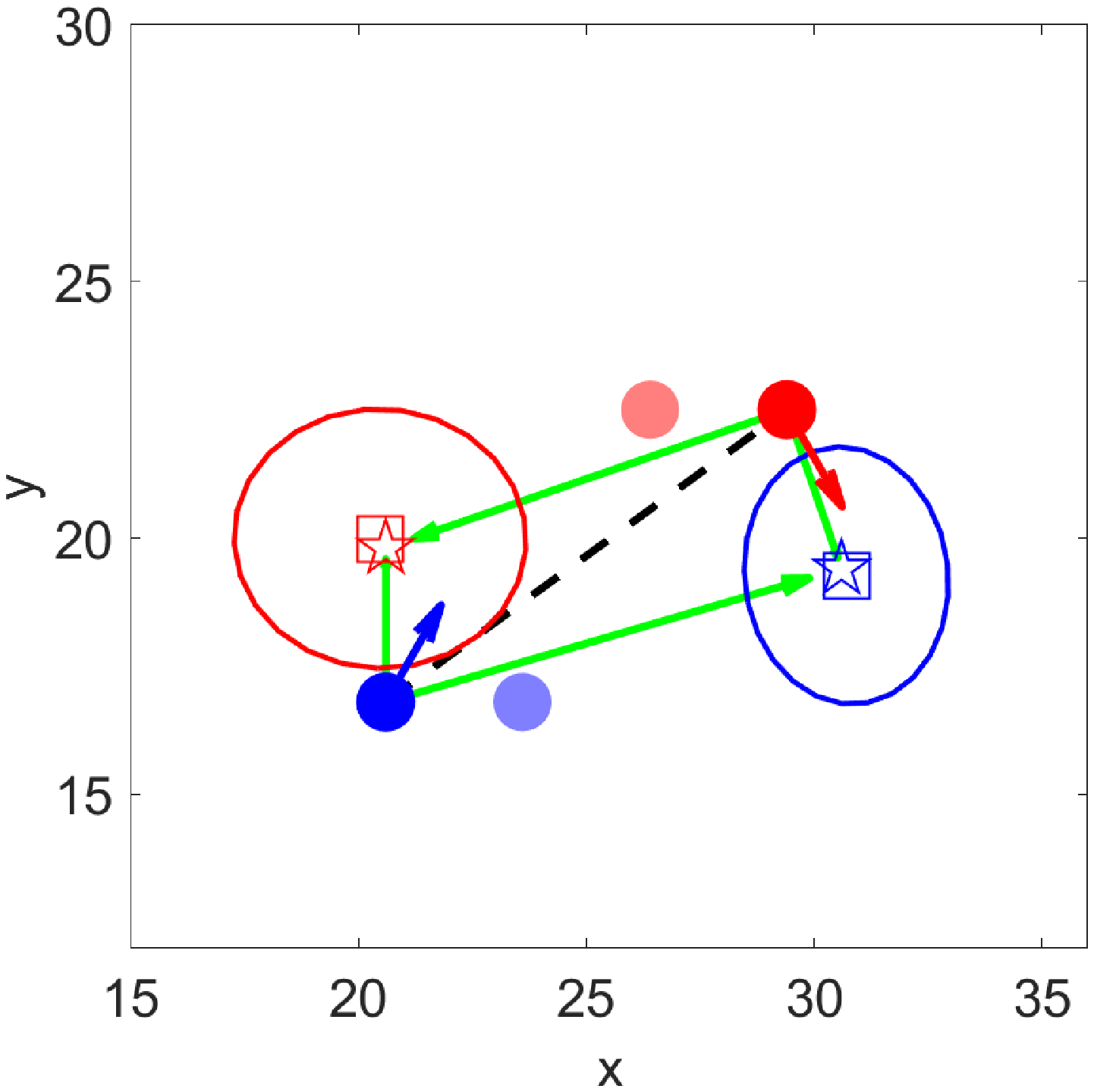}} 
	\subfigure[\texttt{RATT} with 1 communication attack] {\includegraphics[width=0.50\columnwidth]{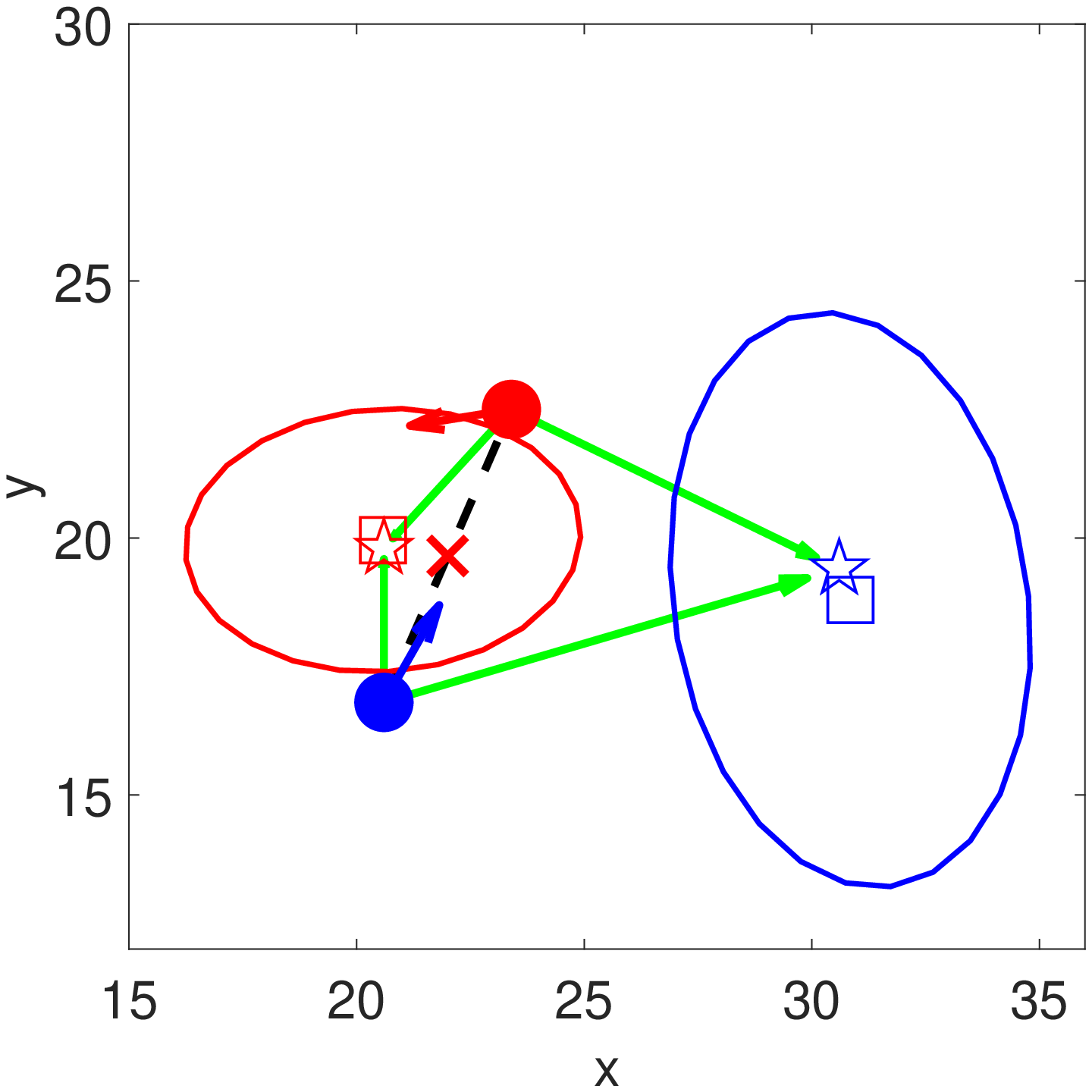}}
	\subfigure[\texttt{NR-OPT} with 1 communication attack] {\includegraphics[width=0.50\columnwidth]{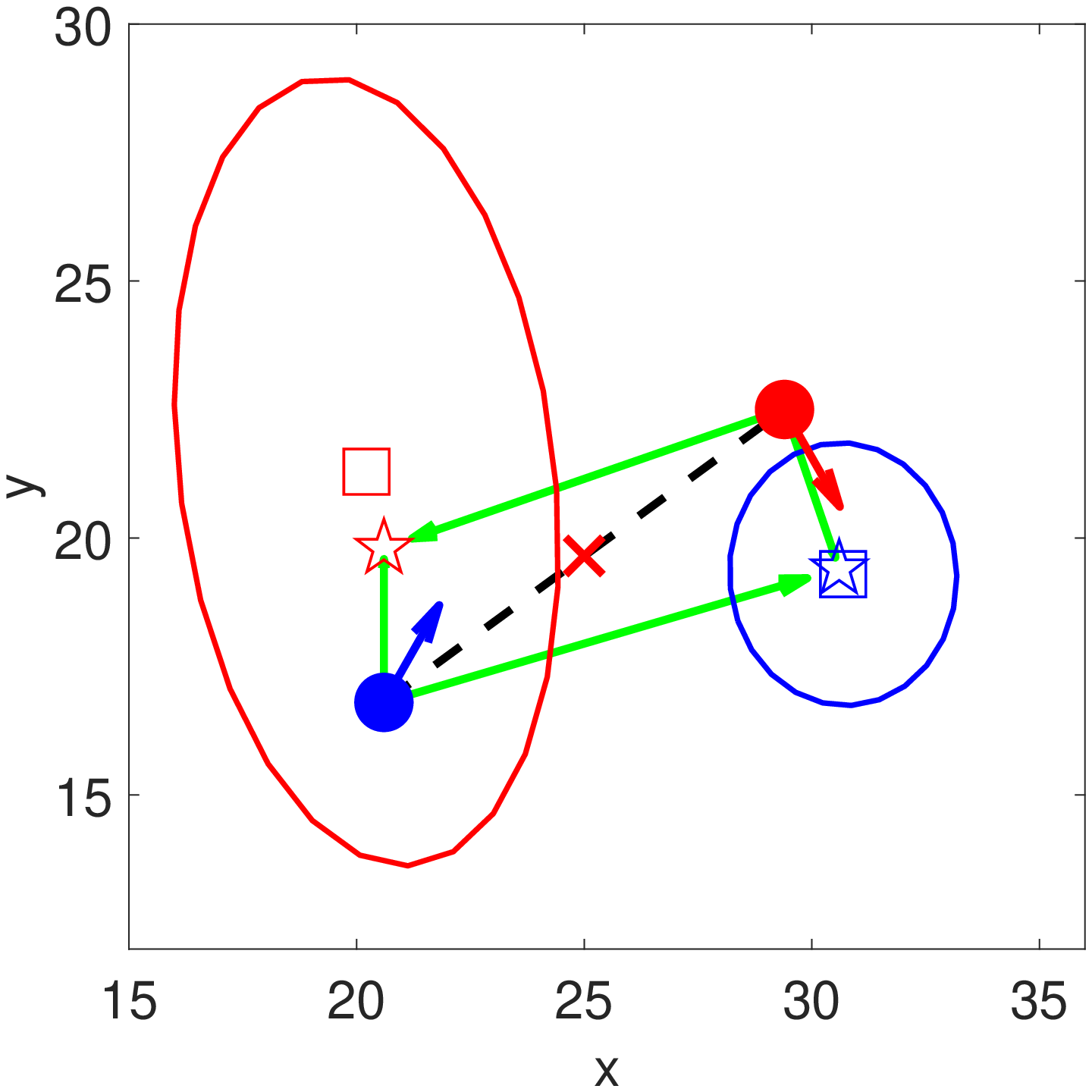}}
\caption{Comparison of 2 robots tracking 2 targets by \texttt{RATT} and \texttt{NR-OPT} without and with1 communication attack. The red cross on a communication link indicates the communication attack on the link. The representations of other depictions follows Fig.~\ref{fig:ql_1sen_atk}.}
\label{fig:ql_1comm_atk}
\end{figure*}

%%%%%%%%%%%%%%%%%%%%%%%%%%%%%%%%%%%%%%%%%%
\begin{figure*}
    \centering
	\subfigure[\texttt{RATT} without  1 sensing attack and 2 communication attacks]{\includegraphics[width=0.50\columnwidth]{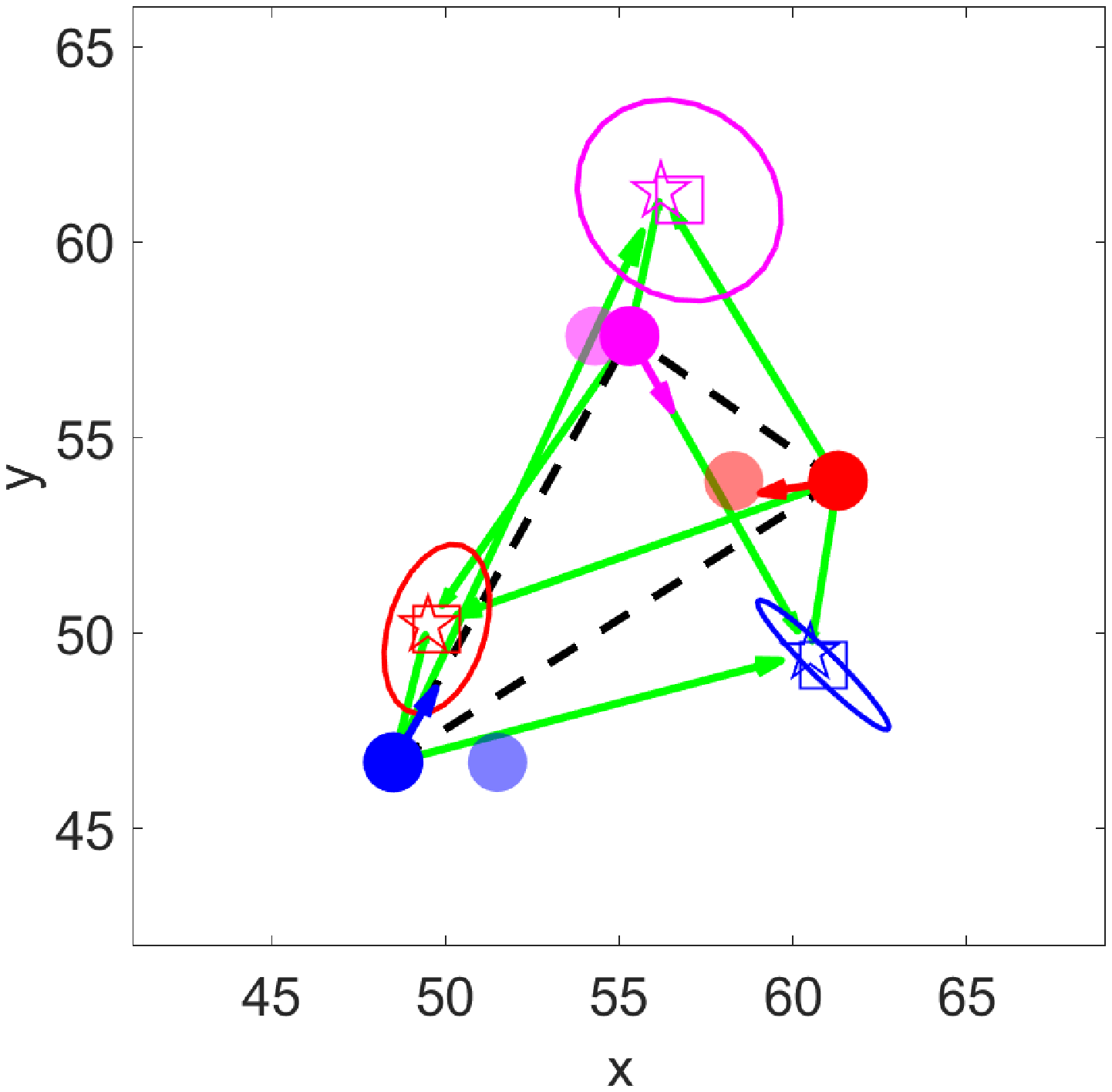}}
	\subfigure[\texttt{NR-OPT} without  1 sensing attack and 2 communication attacks]{\includegraphics[width=0.50\columnwidth]{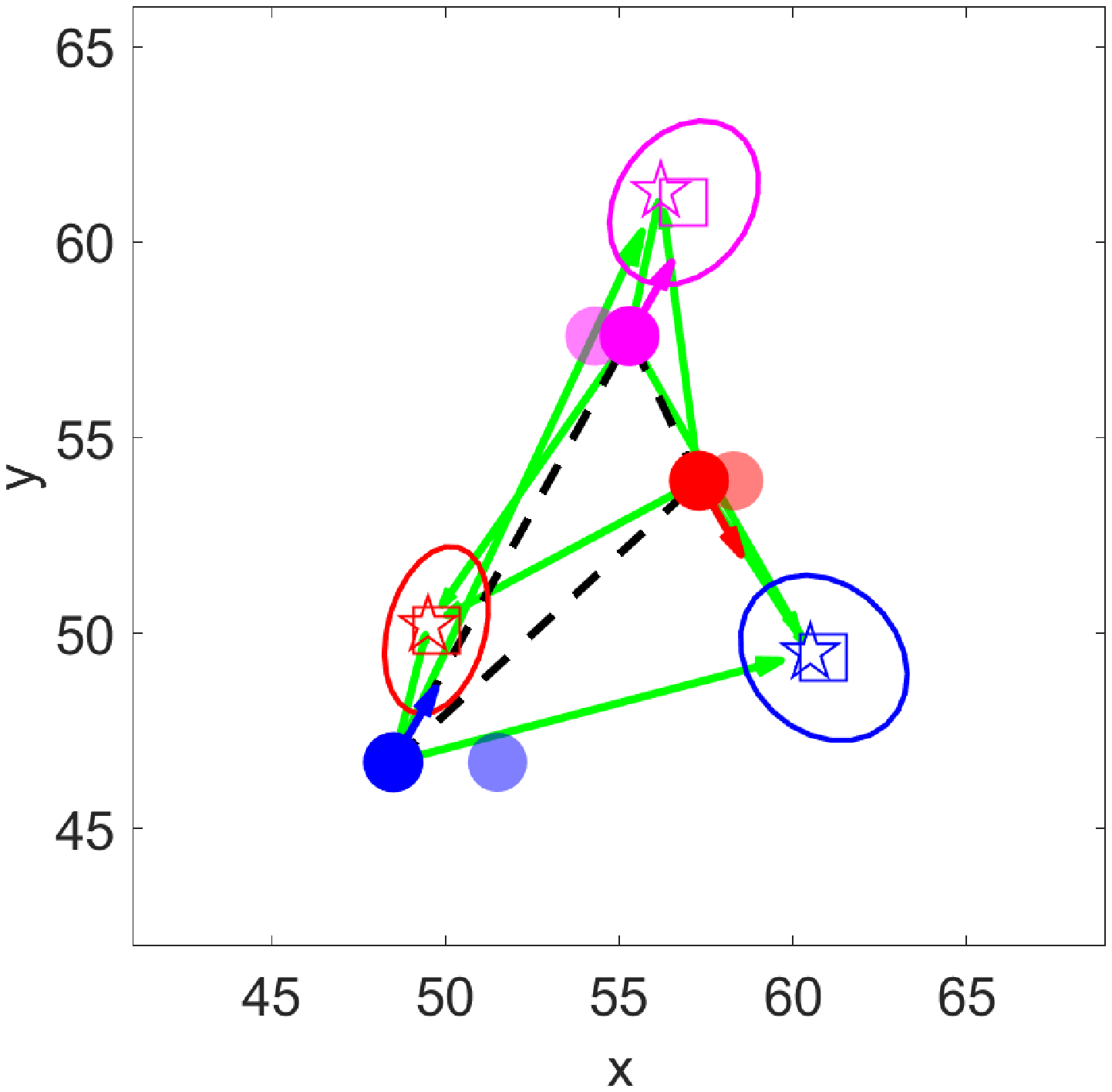}}
	\subfigure[\texttt{RATT} with 1 sensing attack and 2 communication attacks] {\includegraphics[width=0.50\columnwidth]{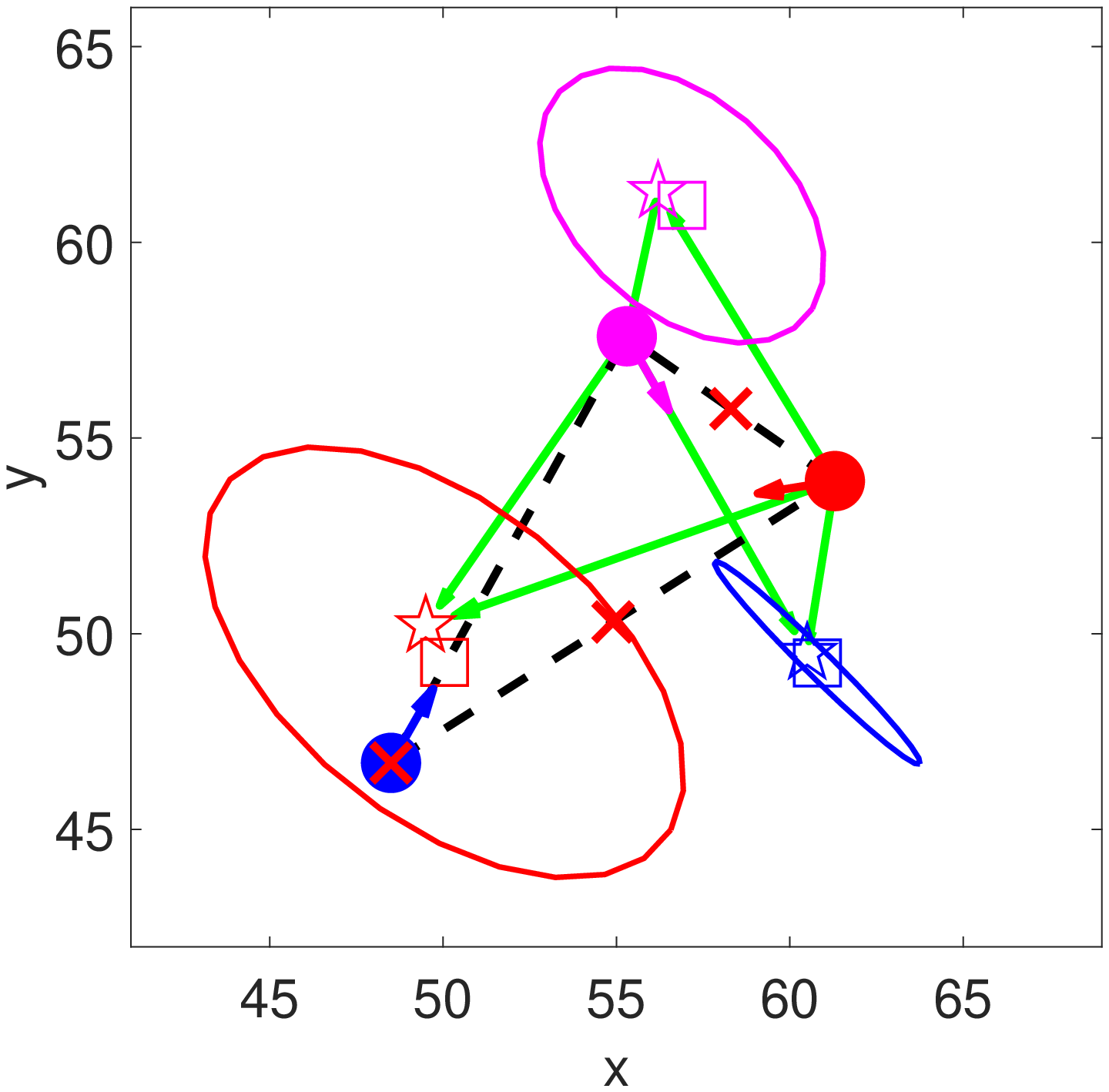}}
	\subfigure[\texttt{NR-OPT} with 1 sensing attack and 2 communication attacks] {\includegraphics[width=0.50\columnwidth]{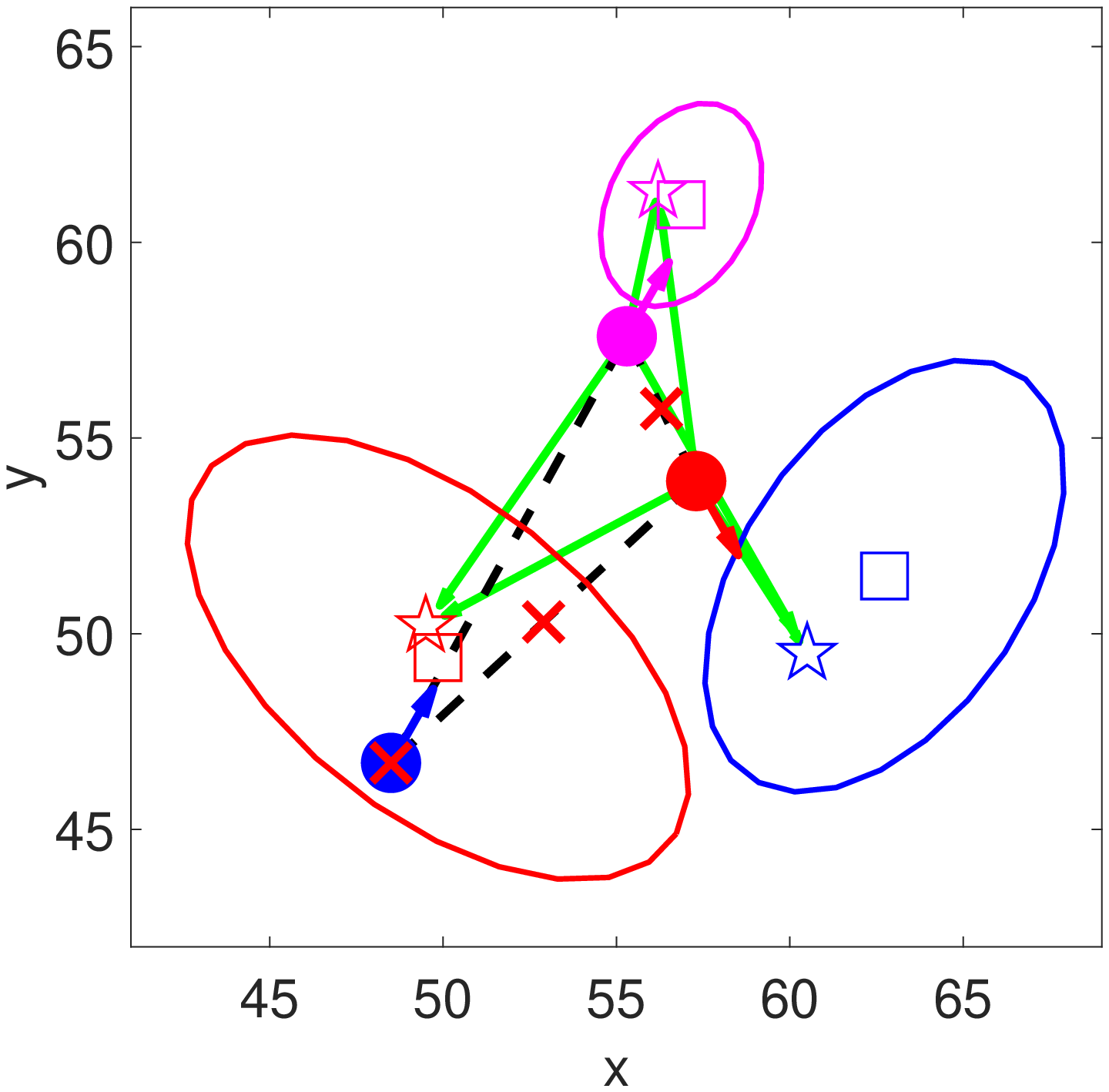}}
\caption{Comparison of 2 robots tracking 2 targets by \texttt{RATT} and \texttt{NR-OPT} without and with 1 sensing attack and 2 communication attacks. The representations of the depictions follows Fig.~\ref{fig:ql_1sen_atk} and Fig.~\ref{fig:ql_1comm_atk}. }
\label{fig:ql_1sen2comm_atk}
\end{figure*}

\section{Numerical Evaluations}~\label{sec:simulation}

We evaluate our algorithm (\texttt{RATT}, Algorithm~\ref{alg:rob_tar_track}) to demonstrate both the necessity for robust target tracking and the benefits of our approach. Particularly, the evaluations demonstrate: (i) \texttt{RATT}'s near-optimal performance, since it performs close to the exhaustive search approach for Problem~\ref{prob:rob_tar_tracking} (which is viable in small-scale scenarios only); (ii) \texttt{RATT} is superior to non-robust heuristics against both worst-case and non-worst-case attacks. We perform the evaluations on a ThinkPad laptop with Intel Core i7 CPU @ 2.7 GHz and 64 GB RAM by using MATLAB 2020a on Ubuntu 20.04. 

% Our MATLAB and Gazebo implementations are available online.\footnote{\url{https://github.com/raaslab/resilient_target_tracking.git}} computational racecourses. 

\textbf{Compared algorithms.} We compare \texttt{RATT} with four other algorithms. The algorithms differ in the way of selecting control inputs for the robots. The first algorithm is an optimal algorithm, which attains the optimal value for Problem~\ref{prob:rob_tar_tracking} by exhaustively searching for all possible robots' control inputs and all possible sensing and communication attacks. Evidently, this algorithm is only viable when the number of robots is small. We refer to this algorithm by \texttt{OPT}. The second algorithm is a non-robust optimal algorithm that ignores the possibility of attacks and uses the exhaustive search to choose control inputs to maximize the team performance (by assuming no attacks). Similar to \texttt{OPT}, this algorithm is viable for small-scale scenarios only. We refer to this algorithm by \texttt{NR-OPT}. The third algorithm is a (non-robust) greedy algorithm that also ignores the possibility of attacks and uses the standard greedy algorithm~\cite{fisher1978analysis} to greedily select the control inputs. We refer to this algorithm by \texttt{Greedy}. The fourth algorithm is a random algorithm that selects randomly (uniformly) the control inputs. We refer to this algorithm by \texttt{Random}.

\textbf{Comparison metrics.} To access $\texttt{RATT}$'s effectiveness against sensing and communication attacks, we compare it with the other four algorithms across multiple scenarios, with varying numbers of robots $N$, varying numbers of targets $N$, and varying numbers of sensing and communication attacks, $\alpha_s$ and $\alpha_c$. We also test the algorithms' performance against varying models of attacks such as the worst-case attack and bounded rational attack. In particular, we consider two performance metrics: the trace of the covariance matrix and the mean squared error (MSE, capturing the gap between the estimated position and true position), averaged over all the targets.

\textbf{Experiment setup.} In \textit{robust multi-target tracking}, a team of mobile robots is tasked to track the locations of multiple moving targets, in the presence of a number of sensing and communication attacks. In what follows, we first specify the robot motion model, target motion model, sensor model, and tracking objective function, as introduced in Section~\ref{subsec:framework}, and then specify the corresponding parameters.  
\paragraph{Robot motion model} Each robot $i \in \mathcal{V}$ moves in a 2D environment with a unicycle motion model: 
% \begin{align} \label{eq:robot_motion_model}
% \begin{split}
%     & x_{i,t+1}^1  =  x_{i,t}^1 + \nu_i \tau \cos(\theta_{i,t}), \\
%     & x_{i,t+1}^2  =  x_{i,t}^2 + \nu_i \tau \sin(\theta_{i,t}), \\
%     & \theta_{i,t+1} = \theta_{i,t} + \tau \omega_i. 
% \end{split}
% \end{align}
\begin{align*} %\label{eq:robot_motion_model}
\begin{split}
{\begin{pmatrix}
x_{i,t+1}^1 \\ x_{i,t+1}^2 \\ \theta_{i,t+1}
\end{pmatrix} = 
\begin{pmatrix}
x_{i,t}^1  \\ x_{i,t}^2 \\ \theta_{i,t}
\end{pmatrix} + 
\begin{pmatrix}
\nu_i \tau \cos(\theta_{i,t})\\
\nu_i \tau \sin(\theta_{i,t})\\
\tau \omega_i
\end{pmatrix},}
\end{split}
\end{align*}
where $\mb{x}_{i}= [x_{i}^1, x_{i}^2, \theta_{i}]^\top$ with $[x_{i}^1, x_{i}^2]^\top$ the position and $\theta_{i}$ the orientation of robot $i$ in the global frame. $[\nu_i, \omega_i]^\top$ denotes the control input (linear and angular velocities) of robot $i$. $\tau$ is the sampling period.  

\paragraph{Target motion model} Each target $i \in \mathcal{T}$ moves in the 2D environment by following a circular motion model with additive white Gaussian noise: 
\begin{align*} %\label{eq:target_motion_model}
\begin{split}
\begin{pmatrix}
y_{j,t+1}^1 \\ y_{j,t+1}^2
\end{pmatrix} = 
\begin{pmatrix}
y_{j,t}^1  \\ y_{j,t}^2
\end{pmatrix} + 
\begin{pmatrix}
\nu_j \cos(\tau \omega_j)\\
\nu_j \sin(\tau \omega_j)
\end{pmatrix} + \mb{w}_{j,t}, 
\end{split}
\end{align*}
where $\mb{y}_{i}= [y_{j}^1, y_{j}^2]^\top$ with $[y_{j}^1, y_{j}^2]^\top$ the position of target $j$ in the global frame. $[\nu_j, \omega_j]^\top$ denotes linear and angular velocities of target $j$. The noise $\mb{w}_{j,t} \sim \mathcal{N}(0, \mb{Q})$ with 
$$
\mb{Q} = \begin{bmatrix}
\sigma_{j}^2  & 0  \\ 0 & \sigma_{j}^2
\end{bmatrix}.
$$

\paragraph{Sensor model} Each robot $i$ observes a target $j$ according to a range and bearing sensor model (as it is the case for depth camera): 
\begin{align*}\label{eq:measure_model_range_bearing}
%  \mb{z}_{i,t}^j =  h_i^j(\mb{x}_{i,t}, \mb{y}_{j,t}) + \mb{v}_{i,t}^j(\mb{x}_{i,t}, \mb{y}_{j,t}). \\
& h_i^j(\mb{x}_{i,t}, \mb{y}_{j,t})  = \begin{bmatrix}r(\mb{x}_{i,t}, \mb{y}_{j,t})\\\gamma(\mb{x}_{i,t}, \mb{y}_{j,t})\end{bmatrix} \\
&\triangleq \begin{bmatrix} \sqrt{(y_{j,t}^2- x_{i,t}^2)^2 + (y_{j,t}^1 - x_{i,t}^1)^2} \\ \text{atan2}(y_{j,t}^2- x_{i,t}^2, y_{j,t}^1 - x_{i,t}^1) - \theta_{i,t}
\end{bmatrix} 
\end{align*}
The measurement noise $\mb{v}_{i,t}^j(\mb{x}_{i,t}, \mb{y}_{j,t}) \sim \mc{N}(0, \mb{R}(\mb{x}_{i,t}, \mb{y}_{j,t}))$ with 
$$
\mb{R}(\mb{x}_{i,t}, \mb{y}_{j,t})) = \begin{bmatrix}
\sigma_{r}^2(\mb{x}_{i,t}, \mb{y}_{j,t})  & 0  \\ 0 & \sigma_{b}^2(\mb{x}_{i,t}, \mb{y}_{j,t})
\end{bmatrix}, 
$$
where $\sigma_{r}^2(\mb{x}_{i,t}, \mb{y}_{j,t})$ and $\sigma_{b}^2(\mb{x}_{i,t}, \mb{y}_{j,t})$ are noise covariances of range and bearing sensors, which grow linearly in range and bearing (between target $j$ and robot $i$). 
% The model also includes a limited sensing range for the sensors, denoted by $r_s$. 
Typically, to implement EKF, we linearize the nonlinear sensor model around the currently predicted position of the target. That is,   
\begin{align*}
& \nabla_{\mb{y}_j} h_i^j(\mb{x}_{i}, \mb{y}_{j}) = \\& \scaleMathLine[1.0]{\frac{1}{r(\mb{x}_{i,t}, \mb{y}_{j,t})} 
\begin{bmatrix} (y_{j}^1 - x_{i}^1) & (y_{j}^2 - x_{i}^2) & 0_{1\times2} \\
-\sin( \theta_i + \gamma(\mb{x}_{i},\mb{y}_{j})) & \cos (\theta_i + \gamma(\mb{x}_{i},\mb{y}_{j})) & 0_{1\times2} \end{bmatrix}.}
\end{align*}

\paragraph{Target tracking objective function} We use an variant of the trace of EKF's posteriori covariance matrix as the target tracking objective function, \textit{i.e.},
\begin{equation*}
    \Phi \triangleq \mathrm{trace}(\Sigma_{\mc{V}, t-1|t})- \mathrm{trace}(\Sigma_{\mc{V}, t|t}),
\end{equation*}
where $\Sigma_{\mc{V}, t-1|t}$ and $\Sigma_{\mc{V}, t|t}$ denote EKF's priori and posteriori covariance matrices, respectively. Notably, the priori covariance matrix is considered as a constant matrix at current time step~\cite{zhou2008optimal,zhou2011multirobot}. Adding the priori covariance matrix is to make the objective function normalized; \textit{e.g.}, if there are no robots' measurements, one has $\Phi = 0$. This objective function is non-decreasing, but not submodular~\cite[Section 4]{jawaid2015submodularity}.

\paragraph{Common parameter specification.} We consider robots and targets are moving inside a $100\times100 m^2$ environment. The available control inputs for each robot $i$ are $\mc{U}_i = \{\pm1,\pm3\} m/s \times \{0, 1, 3\} rad/s$. The velocity of each target $j \in \mc{T}$ is chosen from $\{\pm5/3, \pm5/2, \pm5\} m/s \times \{1/10, 1/20, 1/30\} rad/s$. An initial estimate of each target's position is given to the robots. In addition, the sampling period is set as $\tau = 1s$. 
% The sensing range for each robot is set as $r_s = 50 m$.

% \subsection{Evaluation over one time step}
Next, we evaluate the performance of the aforementioned algorithms through both qualitative and quantitative comparisons. The algorithms are executed over one time step, since we focus on the target tracking quality at one time step ahead (cf. Problem~\ref{prob:rob_tar_tracking}).

\subsection{Qualitative comparison} 
We present qualitative results to show \texttt{RATT}'s benefits against sensing and/or communication attacks by comparing it to \texttt{NR-OPT}, which optimizes the tracking quality by assuming no attacks. Since \texttt{NR-OPT} employs the exhaustive search and is only feasible for small-scale cases, we perform the comparisons with small numbers of robots ($N=2,3$) and small numbers of targets ($M=2,3$). In particular, we consider three robust target tracking scenarios: (i), two robots tracking two targets with one sensing attack (cf. Fig.~\ref{fig:ql_1sen_atk}); (ii), two robots tracking two targets with one communication attack (cf. Fig.~\ref{fig:ql_1comm_atk}); (iii), three robots tracking three targets with one sensing attack and two communication attacks (cf. Fig.~\ref{fig:ql_1sen2comm_atk}). In each scenario, we randomly generate the positions of robots and targets in the environment and set the headings of robots as zero.  We execute the two algorithms with the same initialization, \textit{i.e.}, the same states of robots and targets. Notably, the attacks are worse-case attacks.

\begin{figure*}[th!]
\centering{
% $\alpha_s = \left \lfloor{N/3}\right \rfloor, \alpha_c = \left \lfloor{|\mc{E}|/3}\right \rfloor$
\subfigure[$\alpha_s = 1, \alpha_c = 3 ~(\alpha_{c,s}=1)$]{\includegraphics[width=0.49\columnwidth]{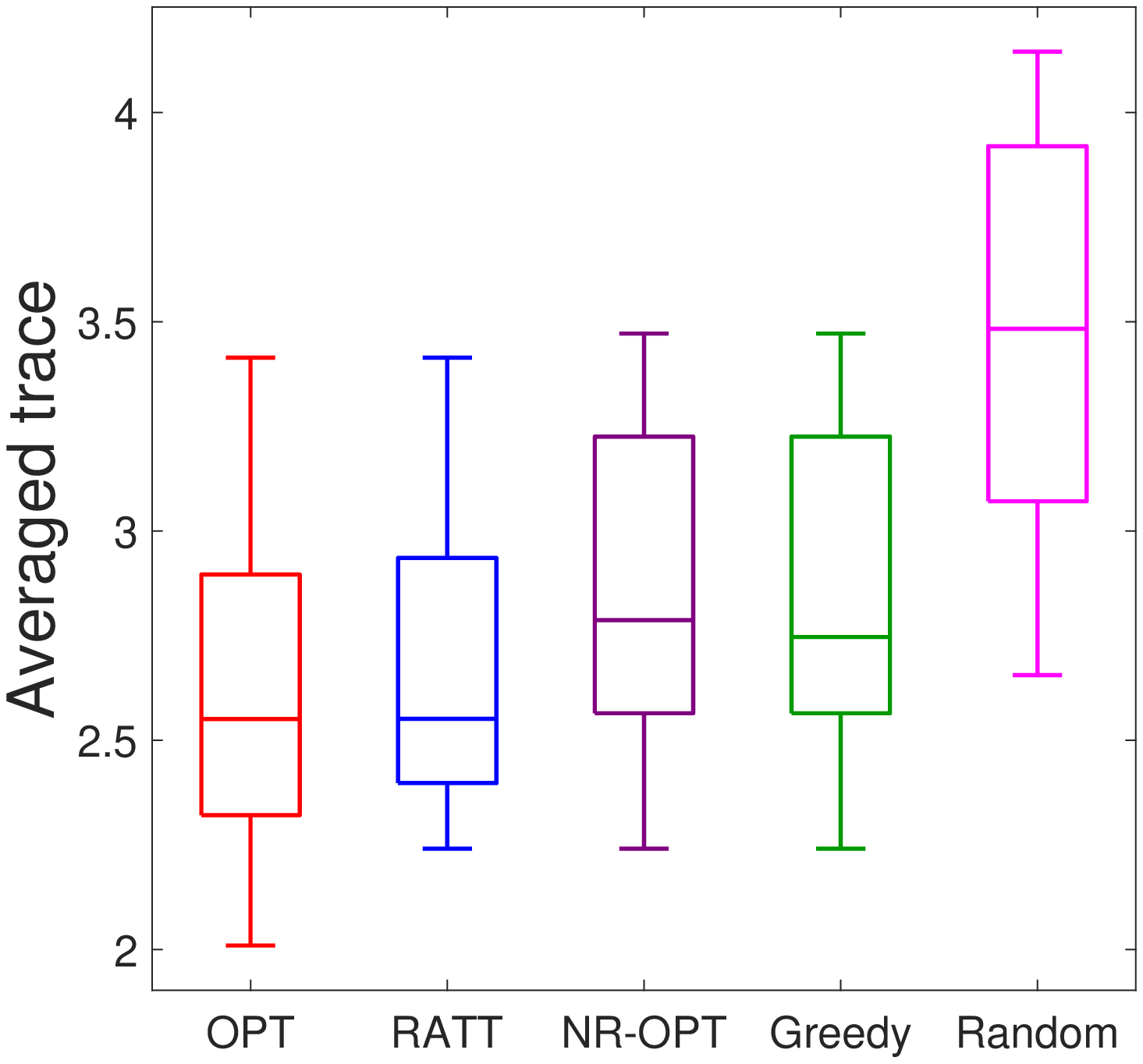}}
% $\alpha_s = \left \lfloor{N/2}\right \rfloor, \alpha_c = \left \lfloor{|\mc{E}|/2}\right \rfloor$
~\subfigure[$\alpha_s = 1, \alpha_c = 4 ~(\alpha_{c,s}=2)$]{\includegraphics[width=0.49\columnwidth]{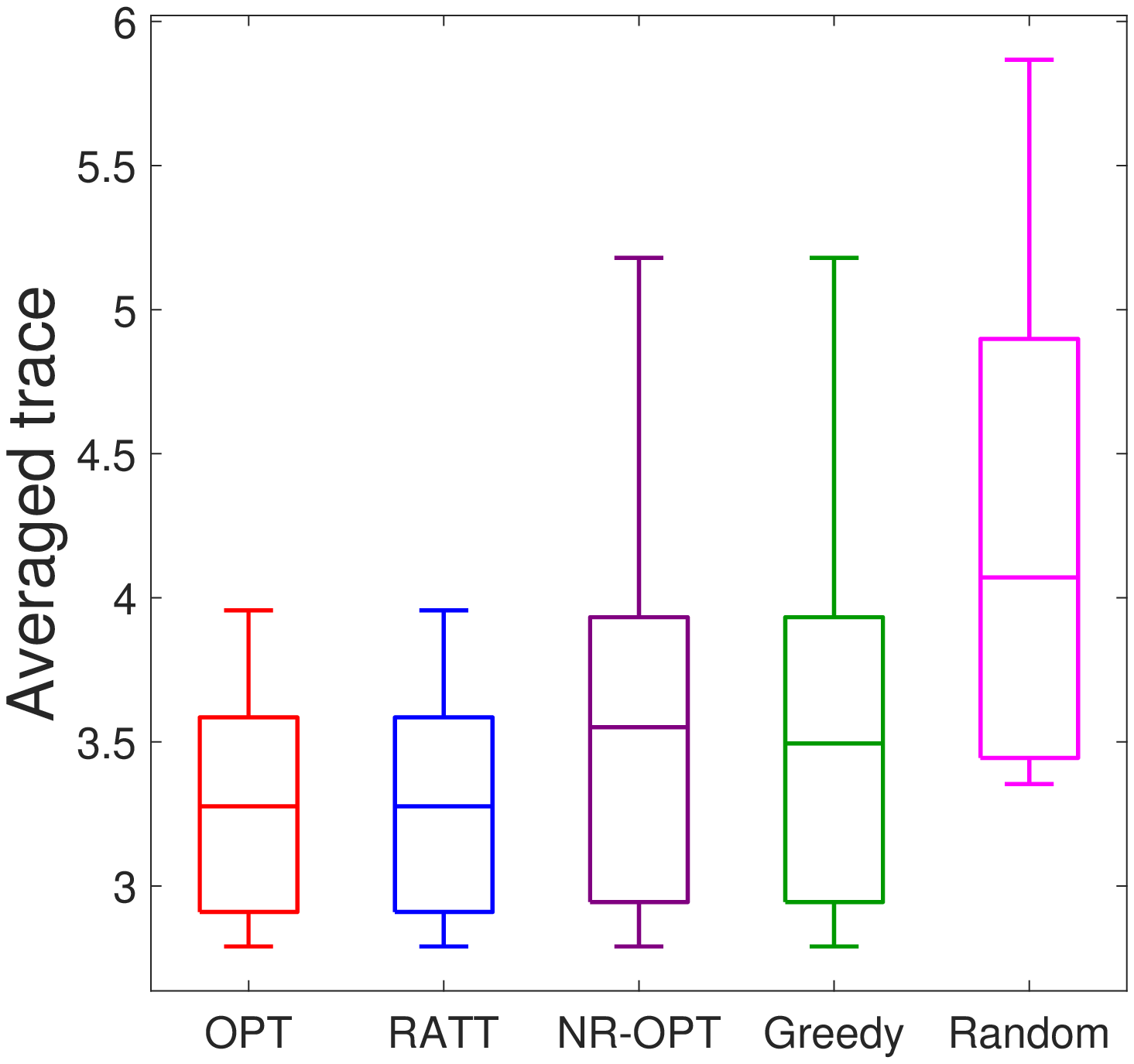}}
% $\alpha_s = \left \lfloor{2N/3}\right \rfloor, \alpha_c = \left \lfloor{|\mc{E}|/3}\right \rfloor$
~\subfigure[$\alpha_s = 2, \alpha_c = 3~(\alpha_{c,s}=1)$]{\includegraphics[width=0.49\columnwidth]{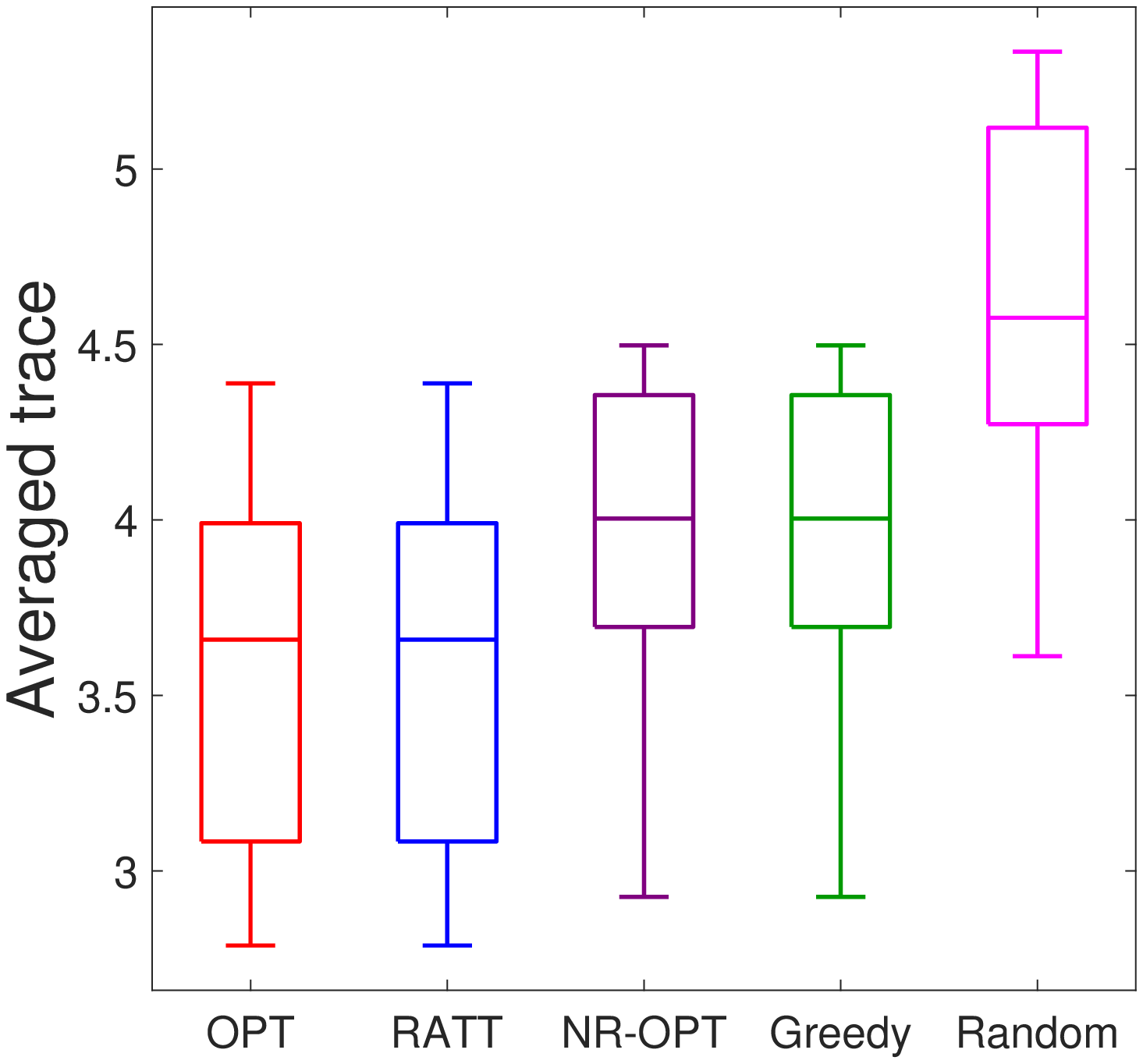}}
% $\alpha_s = \left \lfloor{2N/3}\right \rfloor, \alpha_c = \left \lfloor{2|\mc{E}|/3}\right \rfloor$
~\subfigure[$\alpha_s = 2, \alpha_c = 4~(\alpha_{c,s}=2)$]{\includegraphics[width=0.49\columnwidth]{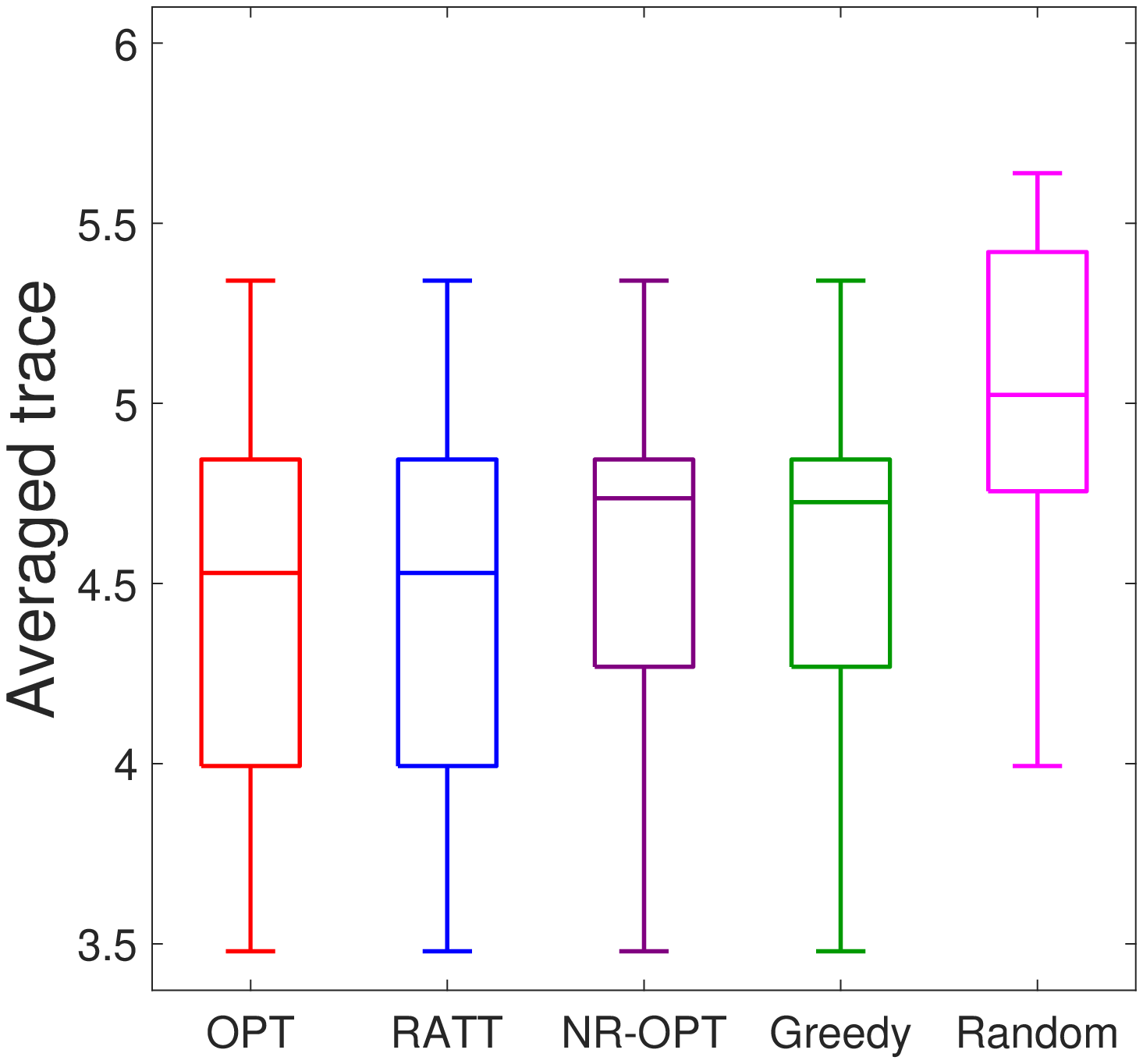}}\\
\subfigure[$\alpha_s = 1, \alpha_c = 3 ~(\alpha_{c,s}=1)$]{\includegraphics[width=0.49\columnwidth]{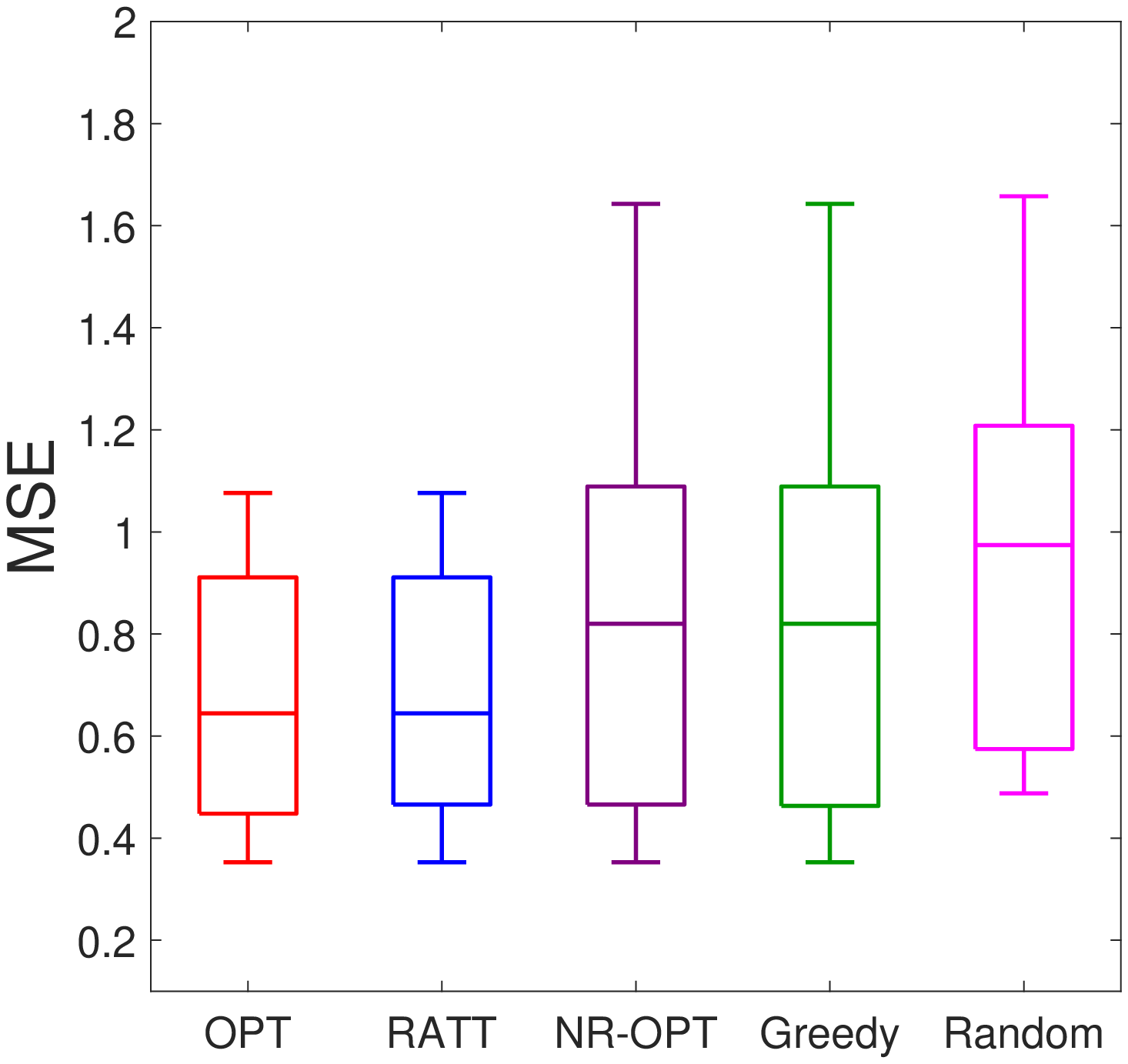}}
~\subfigure[$\alpha_s = 1, \alpha_c = 4 ~(\alpha_{c,s}=2)$]{\includegraphics[width=0.49\columnwidth]{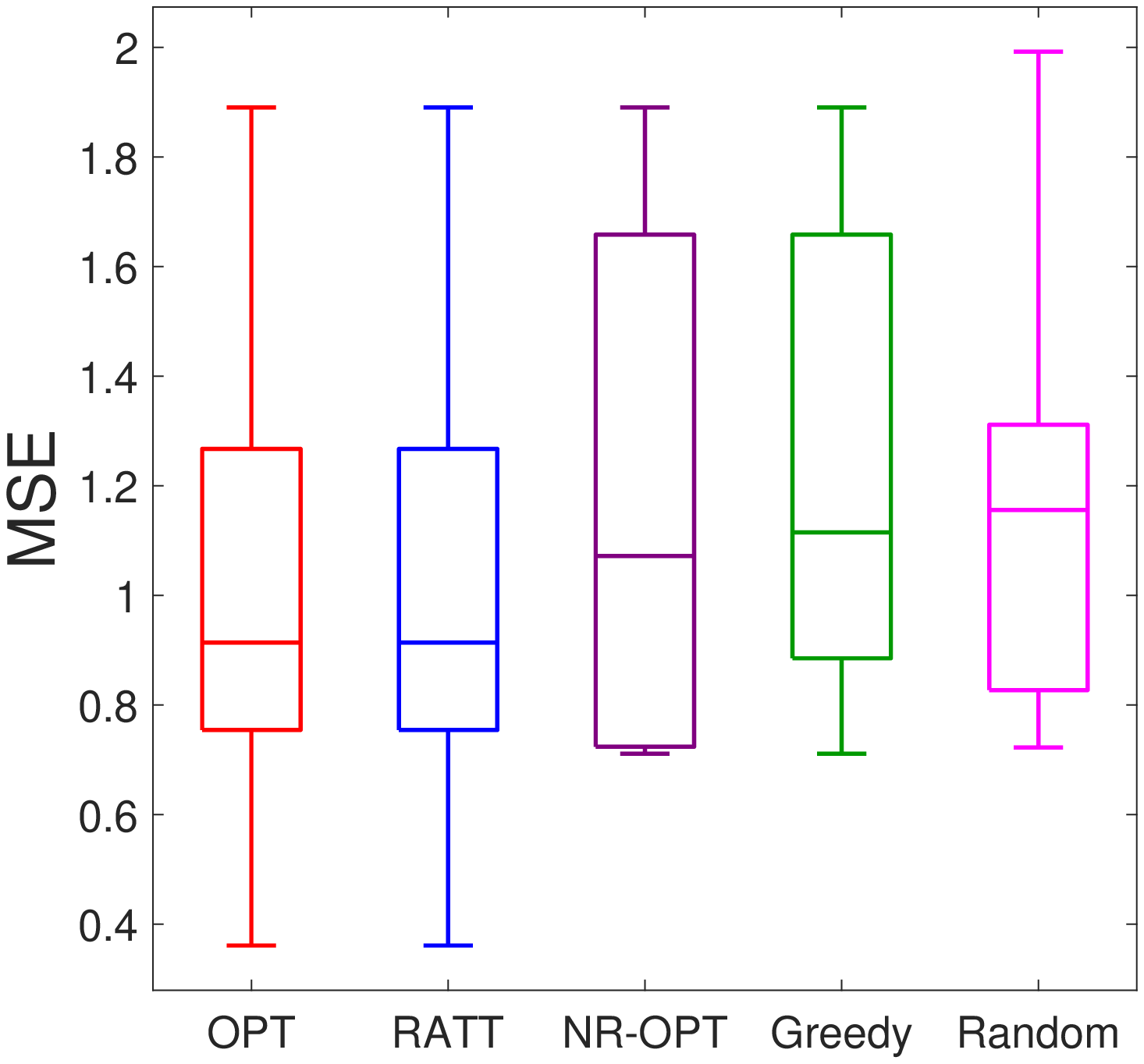}}
~\subfigure[$\alpha_s = 2, \alpha_c = 3~(\alpha_{c,s}=1)$]{\includegraphics[width=0.49\columnwidth]{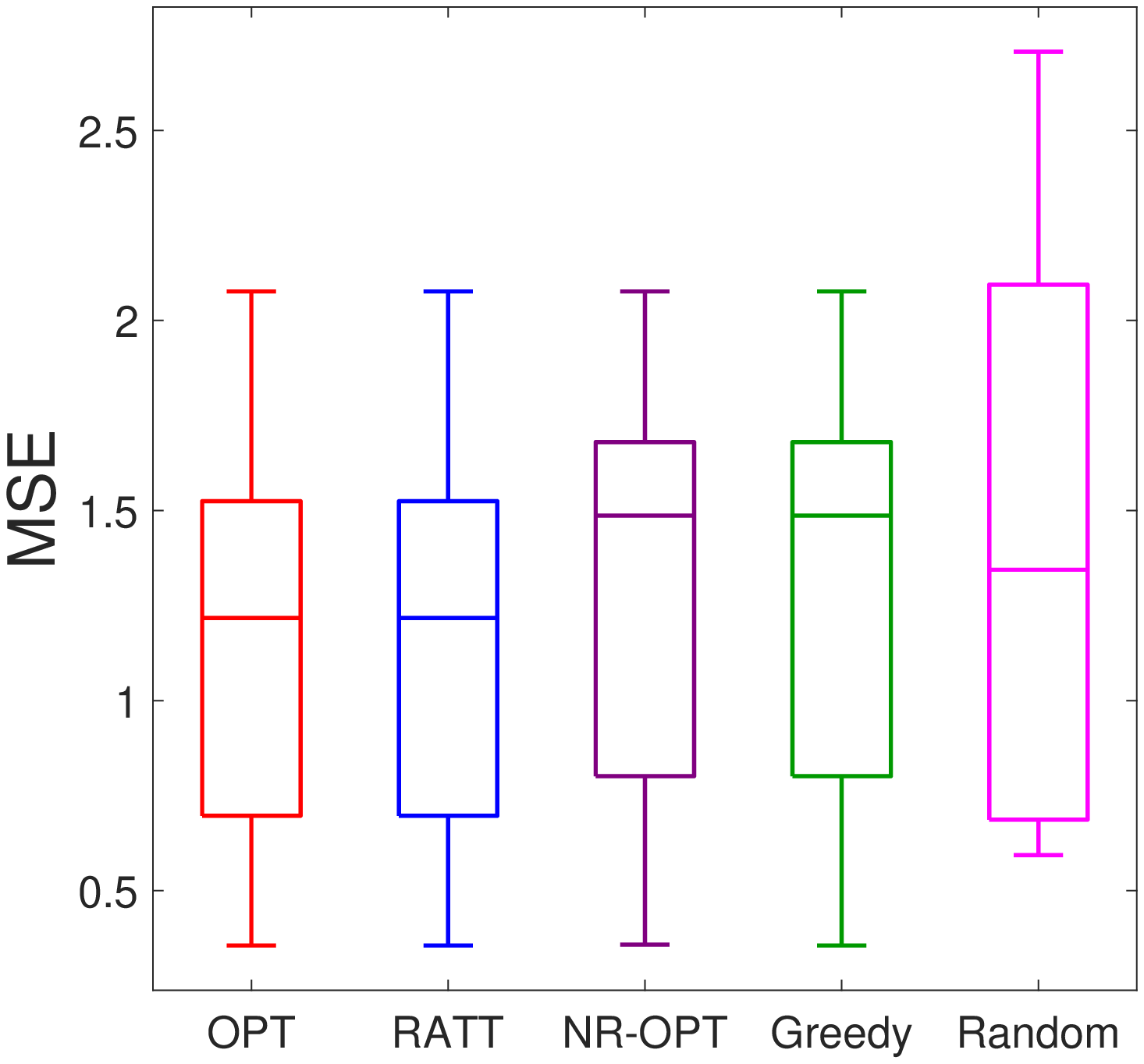}}
~\subfigure[$\alpha_s = 2, \alpha_c = 4~(\alpha_{c,s}=2)$]{\includegraphics[width=0.49\columnwidth]{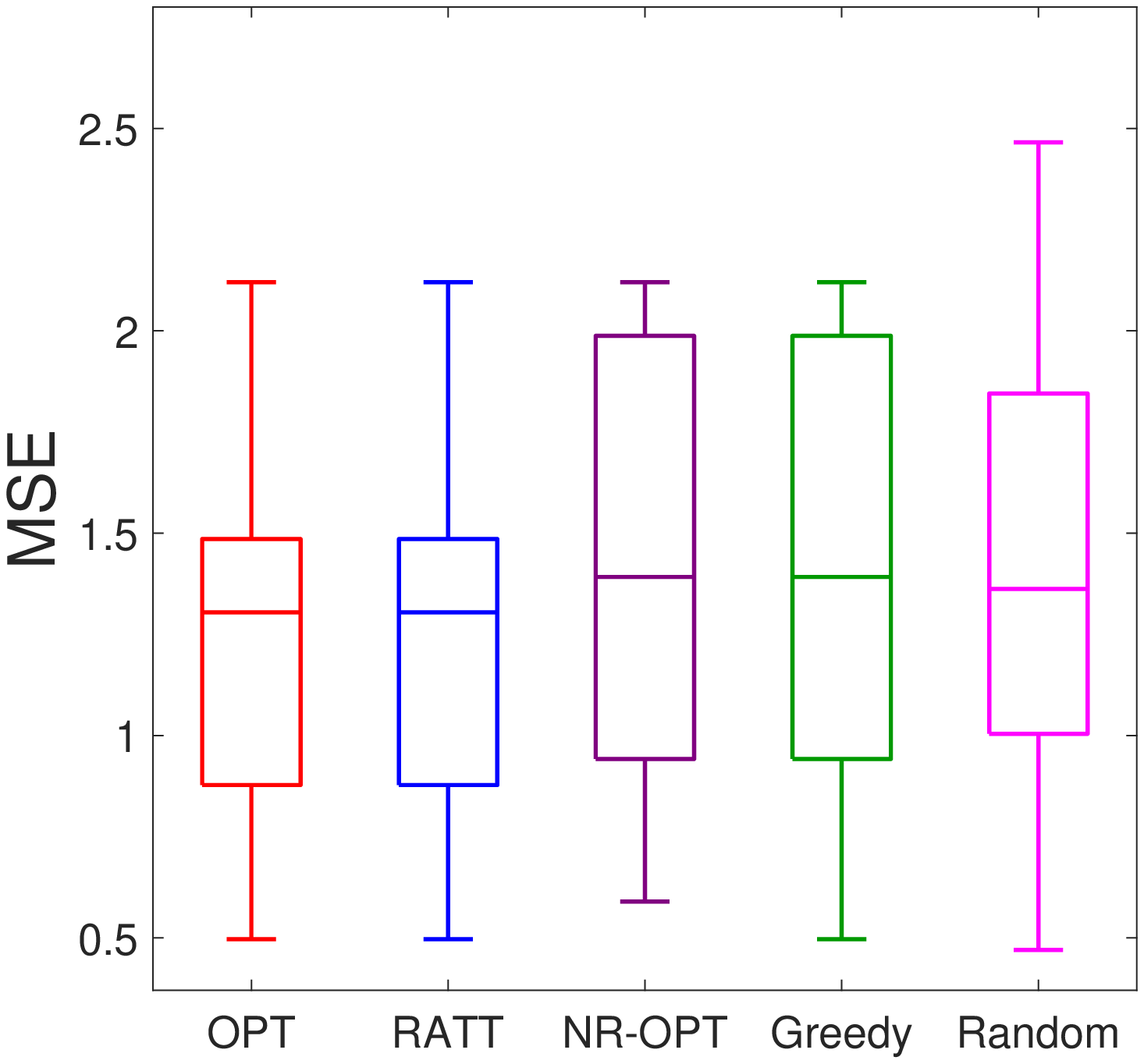}}
\caption{Comparison of the target tracking quality by \texttt{OPT},  \texttt{RATT}, \texttt{NR-OPT}, \texttt{Greedy}, and \texttt{Random} (averaged across 10 Monte Carlo runs): (a)-(d) depict the results of averaged trace of covariance, across four combinations of $\alpha_s$ and $\alpha_c$ values; and (e)-(h) depict corresponding results of MSE. 
\label{fig:small_comp}}}
\end{figure*}

\begin{figure*}
\centering
% \begin{tabular}{c|c|c}
\subfigure[$N=10,M=15$]{\includegraphics[width=0.67\columnwidth]{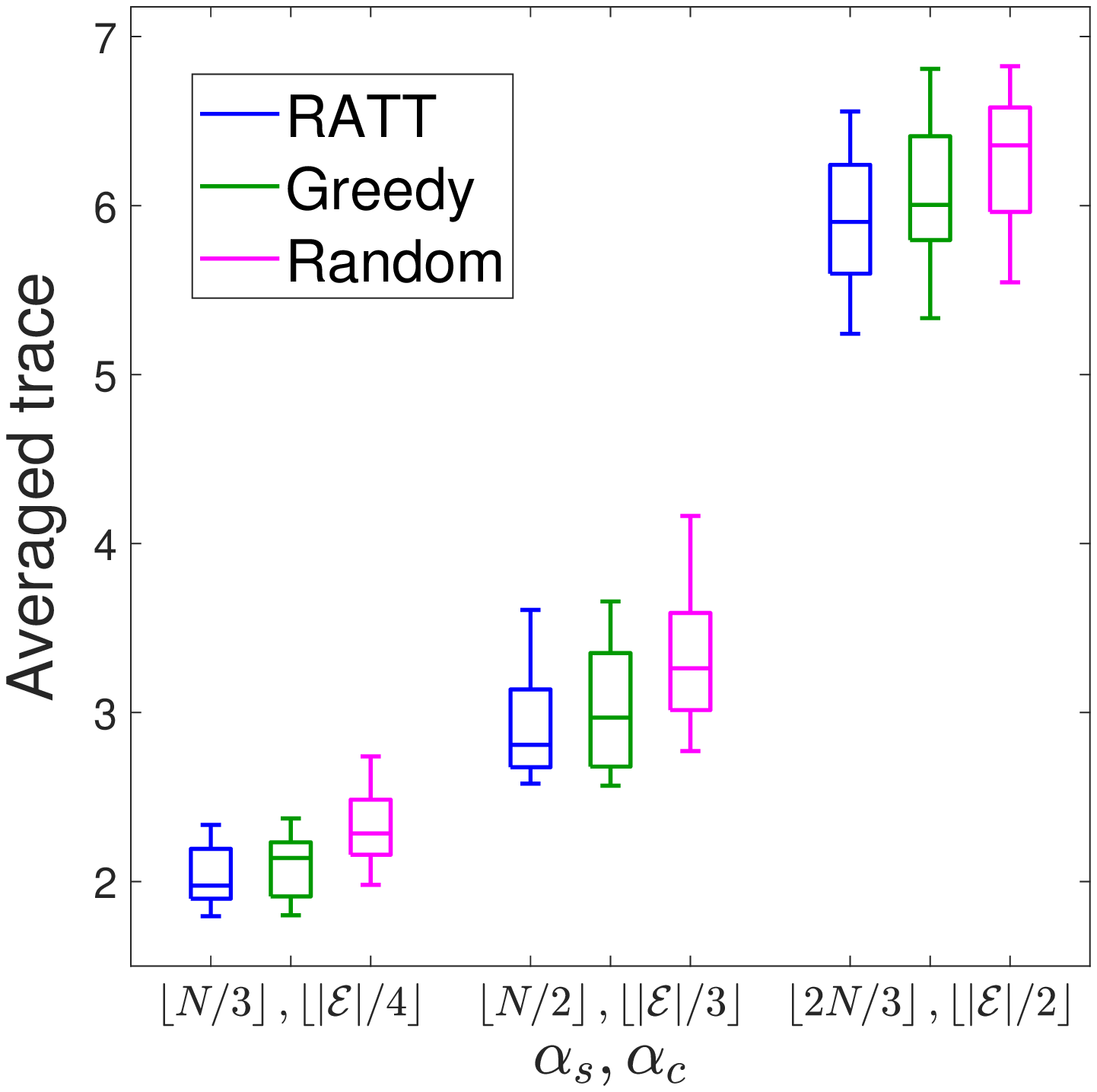}}
\subfigure[$N=20,M=20$]{\includegraphics[width=0.67\columnwidth]{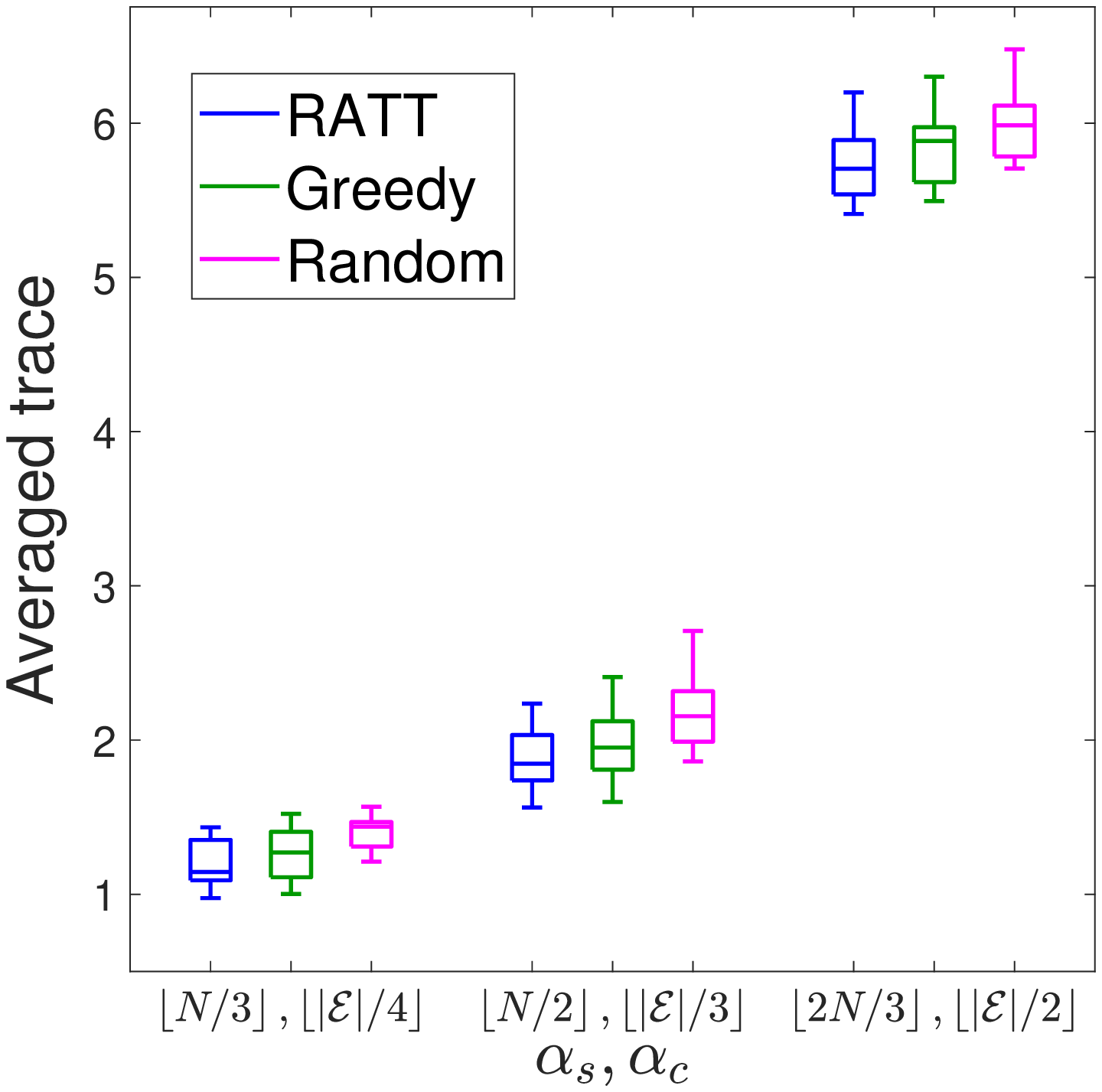}}
\subfigure[$N=30,M=25$]{\includegraphics[width=0.67\columnwidth]{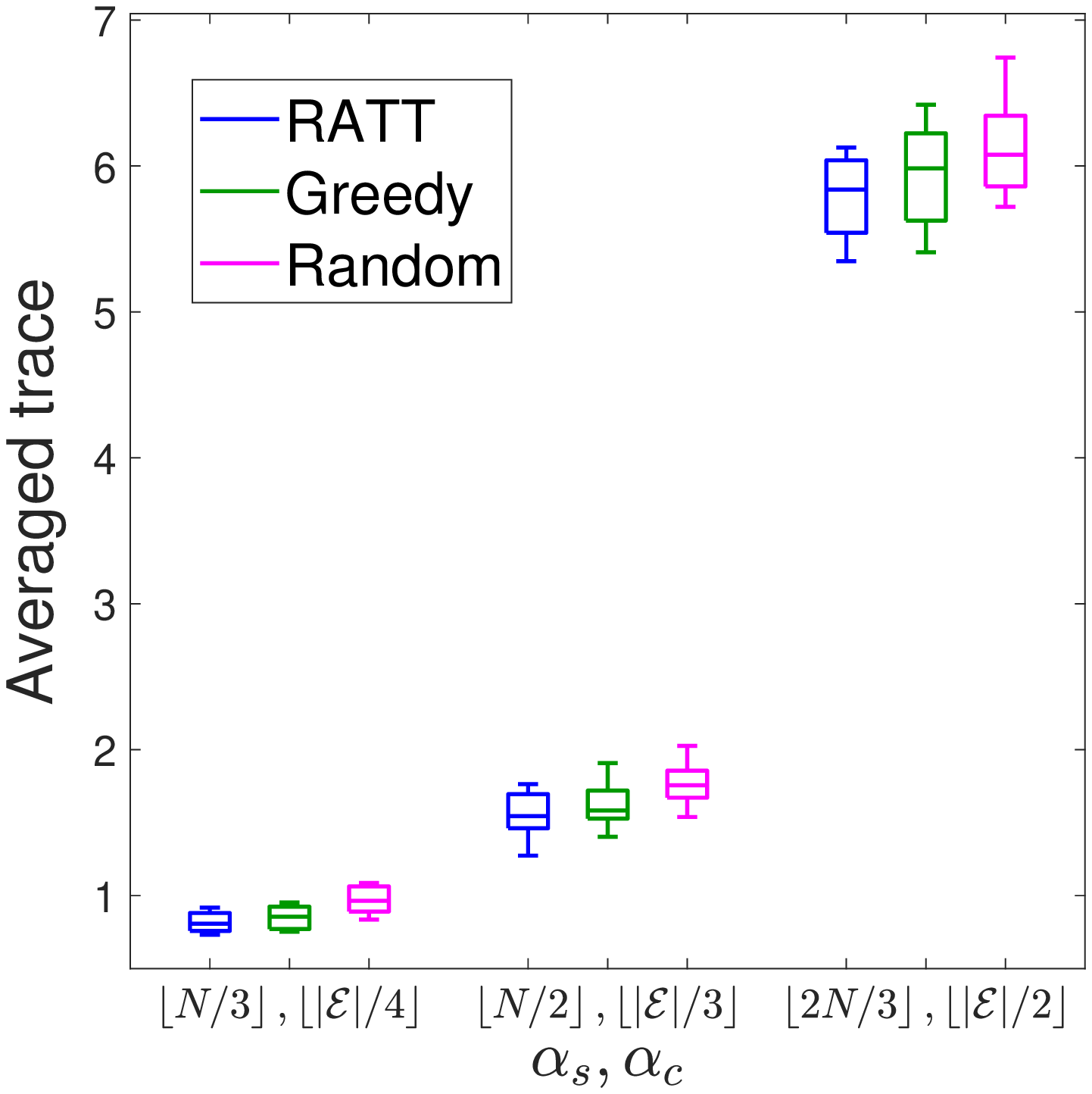}}\\
\subfigure[$N=10,M=15$]{\includegraphics[width=0.67\columnwidth]{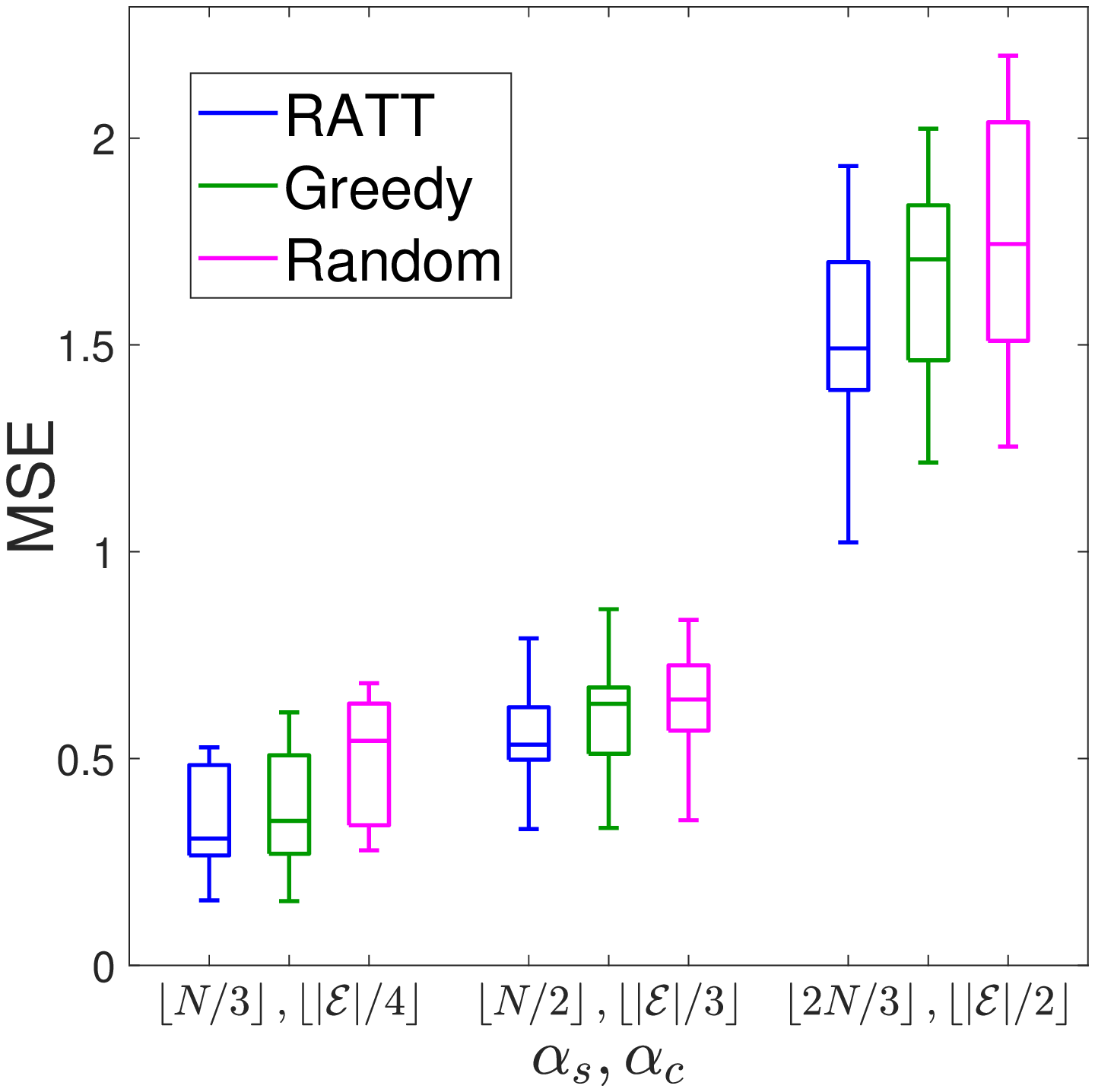}}
\subfigure[$N=20,M=20$]{\includegraphics[width=0.67\columnwidth]{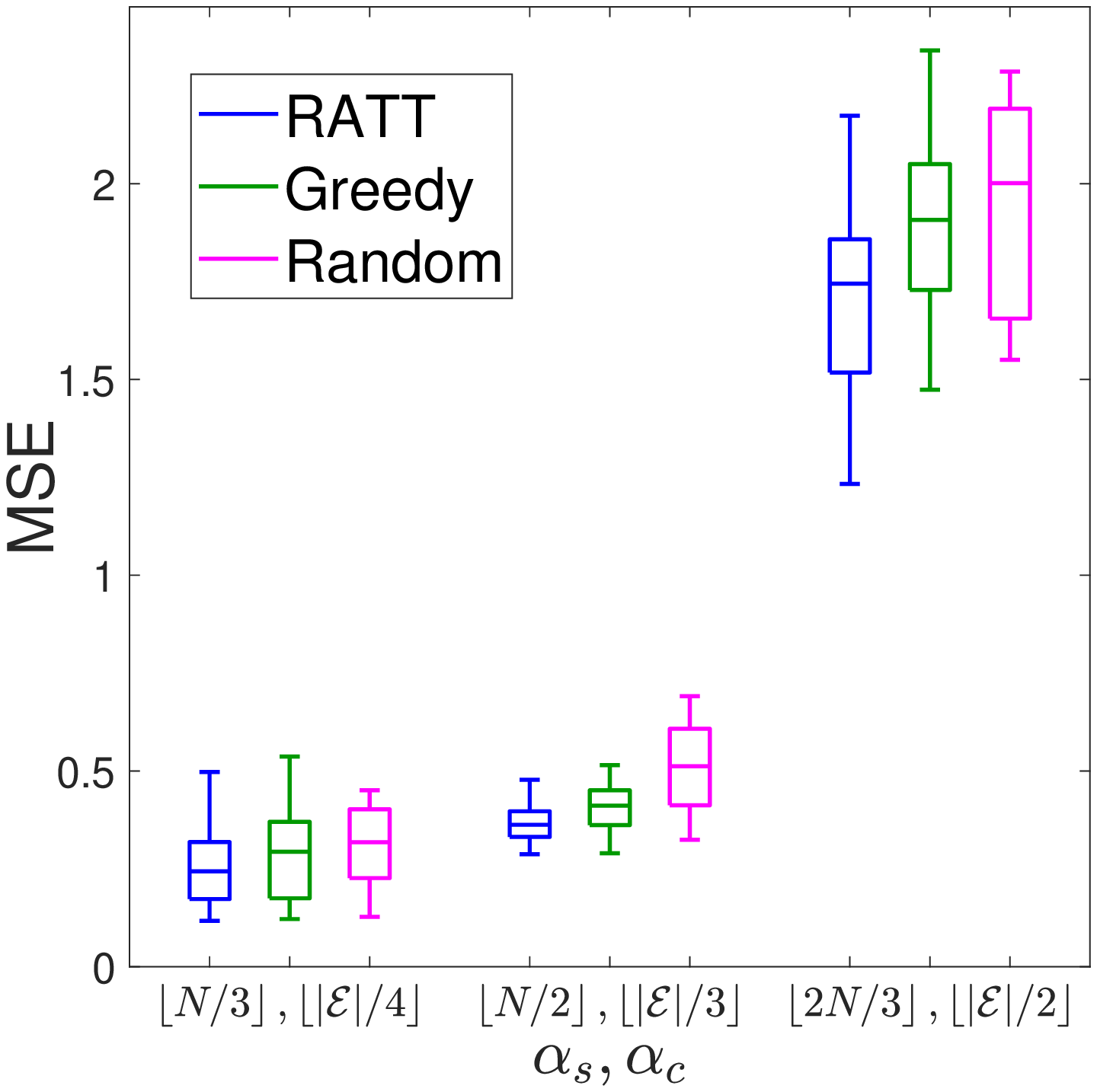}}
\subfigure[$N=30,M=25$]{\includegraphics[width=0.67\columnwidth]{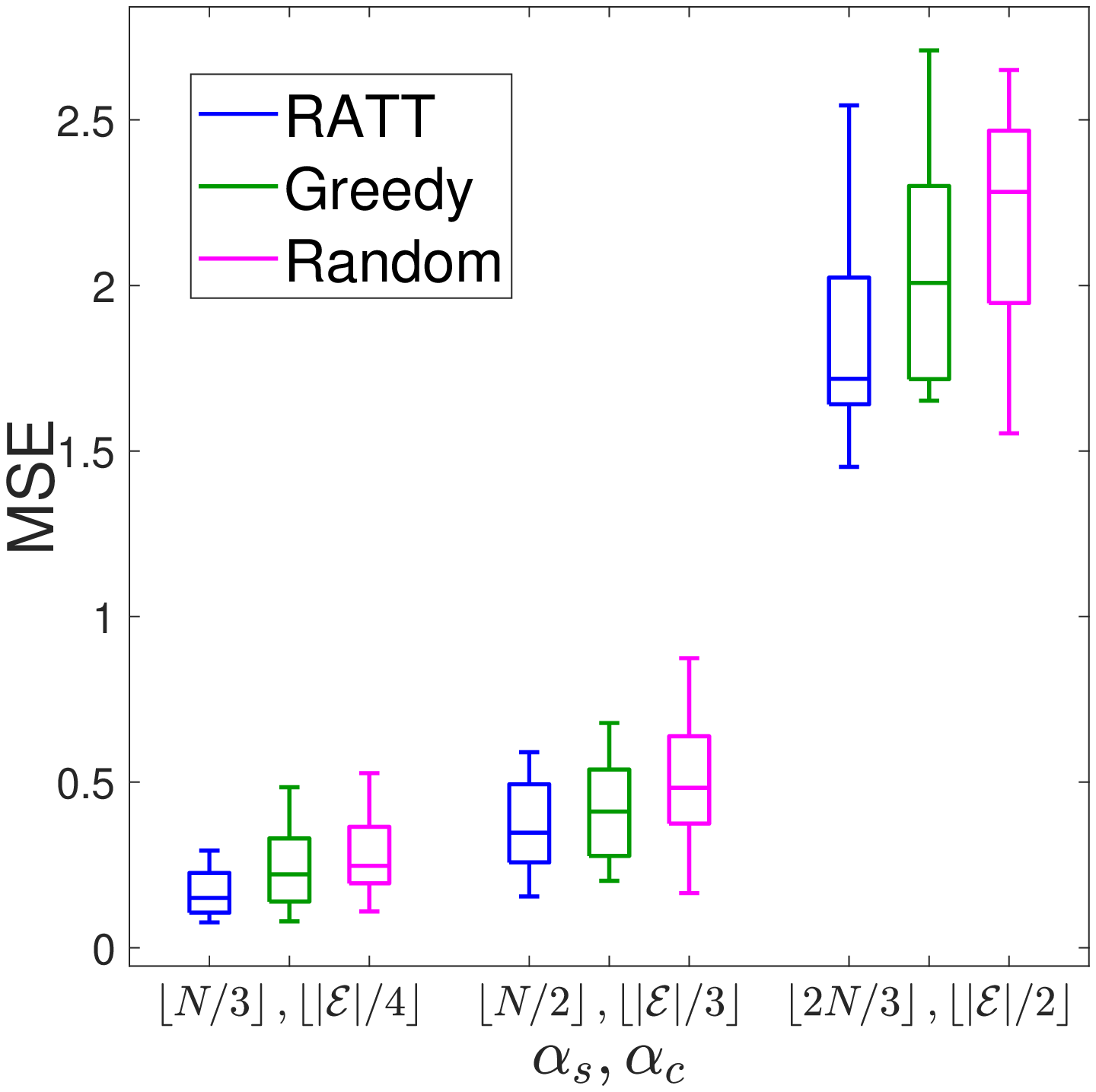}}
% \end{tabular}
\caption{Comparison of the target tracking quality by \texttt{RATT}, \texttt{Greedy}, and \texttt{Random} (averaged across 10 Monte Carlo runs): (a)-(c) depict the results of averaged trace of covariance, across three pairs of $N$ and $M$ values, each with three combinations of $\alpha_s$ and $\alpha_c$ values; and (e)-(h) depict corresponding results of MSE. 
\label{fig:large_comp}}
\end{figure*}

The results for scenario (i) are reported in Fig.~\ref{fig:ql_1sen_atk} where the tracking quality of \texttt{RATT} and \texttt{NR-OPT} is compared without and with one sensing attack. Specifically, without  the sensing attack (Fig.~\ref{fig:ql_1sen_atk}-(a) \& (b)), \texttt{NR-OPT} attains a better tracking quality than \texttt{RATT}, as observed from the covariances (depicted by ellipses, capturing the uncertainty in position estimate) and the gap between the estimated mean positions (depicted by squares) and true positions (depicted by pentagrams) of the targets. This is naturally true, since \texttt{NR-OPT} is an optimal algorithm for non-robust target tracking. However, when there exists a sensing attack, \texttt{NR-OPT} loses its good tracking quality of all targets. For example, as shown in Fig.~\ref{fig:ql_1sen_atk}-(c), the tracking quality of the blue target drops sharply, as the covariance and position estimation gap increases dramatically when the sensing of the blue robot is attacked. While, as observed from Fig.~\ref{fig:ql_1sen_atk}-(d)), \texttt{RATT} shows its superiority against the sensing attack, since it retains a relatively good tracking quality for the two targets, with a slight increase in the estimation gap and a moderate increase in the covariance.

The results for scenarios (ii) and (iii) are reported in Fig.~\ref{fig:ql_1comm_atk} and Fig.~\ref{fig:ql_1sen2comm_atk} where the tracking quality of the two algorithms is compared without and with one communication attack and without and with one sensing and two communication attacks, respectively. Fig.~\ref{fig:ql_1comm_atk} and Fig.~\ref{fig:ql_1sen2comm_atk} exhibit similar results as in Fig.~\ref{fig:ql_1sen_atk}: \texttt{NR-OPT} performs better without the attack, but loses sharply its tracking quality for some target when the attack happens; while \texttt{RATT} maintains a good tracking quality for all targets, even in the presence of the communication attack or both sensing and communication attacks.  

To summarize, the qualitative results in all three scenarios demonstrate the necessity for robust target tracking and the superiority of \texttt{RATT} against sensing and/or communication attacks.

\subsection{Quantitative comparison} 
We present quantitative results to demonstrate \texttt{RATT}'s superiority and robustness against sensing and communication attacks. In particular, we perform both small-scale and large-scale comparisons across various scenarios (with varying $N$, $M$, $\alpha_s$, $\alpha_c$). For each specific scenario, the positions of robots and targets are randomly generated in the environment with robots' headings as zero, across 10 trials. For each trial, all algorithms are executed with the same initialization, \textit{i.e.}, the same states of robots and targets.

\paragraph{Small-scale comparison} We compare the tracking quality of \texttt{RATT} with that of \texttt{OPT}, \texttt{NR-OPT}, \texttt{Greedy}, and \texttt{Random}. Since \texttt{OPT} and \texttt{NR-OPT} are only viable for small-scale cases, we consider small numbers of robots and targets, \textit{e.g.}, $N=M=4$. The algorithms are compared across four combinations of sensing and communication attacks, \textit{i.e.}, $(\alpha_s = 1, \alpha_c = 3), (\alpha_s = 2, \alpha_c = 3), (\alpha_s = 1, \alpha_c = 4), (\alpha_s = 2, \alpha_c = 4)$. 
% Recall that the total number of communication links is $|\mc{E}| = 4\times(4-1)/2 = 6$. 
We evaluate these five algorithms by executing the worst-case attacks.

% $(\alpha_s = \left \lfloor{N/3}\right \rfloor, \alpha_c = \left \lfloor{|\mc{E}|/3}\right \rfloor), (\alpha_s = \left \lfloor{N/2}\right \rfloor, \alpha_c = \left \lfloor{|\mc{E}|/2}\right \rfloor), (\alpha_s = \left \lfloor{2N/3}\right \rfloor, \alpha_c = \left \lfloor{|\mc{E}|/3}\right \rfloor), (\alpha_s = \left \lfloor{2N/3}\right \rfloor, \alpha_c = \left \lfloor{2|\mc{E}|/3}\right \rfloor)$ with $|\mc{E}| = N(N-1)/2$

The comparison results are reported in Fig.~\ref{fig:small_comp}. It is observed that \texttt{RATT} (colored blue) has on average a superior tracking quality to \texttt{NR-OPT} (colored), \texttt{Greedy} (colored green), and \texttt{Random} (colored magenta) in terms of both averaged trace of covariance and MSE. Additionally, \texttt{RATT} performs close to \texttt{OPT} (colored red), an optimal algorithm achieved by exhaustive search.

\paragraph{Large-scale comparison} We compare $\texttt{RATT}$'s tracking quality with that of \texttt{Greedy} and \texttt{Random} in large-scale scenarios, \textit{e.g.}, $N\in[10, 20, 30]$ and $M\in [15, 20, 25]$. \texttt{OPT} is not included in the comparison due to its long evaluation time. For example, when $N=10$, \texttt{OPT} needs to evaluate $|\mc{U}_i|^{10}$ ($12^{10}$) possible cases of robots' control inputs; for each case, its needs to evaluate $\binom{10}{\alpha_s} \times \binom{10\times 9/2}{\alpha_c}$ possible cases of $\alpha_s$ sensing and $\alpha_c$ communication attacks to find the worst-case attack; thus, in total, it takes $12^{10}\times \binom{10}{\alpha_s} \times \binom{10\times 9/2}{\alpha_c} $ to find the optimal control inputs. Evidently, \texttt{OPT} becomes intractable with a large number of robots.  

As computed above, finding the worst-case attack requires the exhaustive search of all possible $\alpha_s$ sensing and $\alpha_c$ communication attacks, a problem of combinatorial complexity, and thus is NP-hard~\cite{orlin2018robust}. Hence, when the number of robots is large, we consider the attacker to use a bounded rational attack approach, instead of executing the worst-case attack. The bounded rational attack sequentially executes two steps to attack robots' sensing and communication: first, it removes the sensing measurements from the top $\alpha_s$ robots (whose maximum individual tracking qualities are among the top $\alpha_s$ ones in the team); second, from the remaining $N-\alpha_s$ robots, it blocks the communications of the top $\alpha_{c,s}$ robots. Evidently, the bounded rational attack takes the same running time as sorting out the top $\alpha=\alpha_s + \alpha_{c,s}$ robots (\textit{i.e.}, $O(N\log(N))$) and thus can be used an evaluation algorithm in large-scale scenarios. 

% a greedy heuristic~\cite{nemhauser1978analysis} to attack robots' sensing and communications, 
% Notably, this attack approximation is \textit{not} considered during the design of the algorithms (\textit{e.g.}, \texttt{OPT} and \texttt{RATT}). It is only used to evaluate the algorithms' performance when computing the worst-case attack is not viable.

With the bounded rational attack, the three algorithms are evaluated across three pairs of numbers of robots and targets, \textit{i.e.}, $(N=10, M =15), (N=20, M =20), (N=30, M =25)$, each pair with three combinations of sensing and communication attacks, \textit{i.e.},  $(\alpha_s = \left \lfloor{N/3}\right \rfloor, \alpha_c = \left \lfloor{|\mc{E}|/4}\right \rfloor), (\alpha_s = \left \lfloor{N/2}\right \rfloor, \alpha_c = \left \lfloor{|\mc{E}|/3}\right \rfloor)$, and $(\alpha_s = \left \lfloor{2N/3}\right \rfloor, \alpha_c = \left \lfloor{|\mc{E}|/2}\right \rfloor)$.

The comparison results are reported in Fig.~\ref{fig:large_comp}. We observe that \texttt{RATT} (colored blue) is again superior to both \texttt{Greedy} (colored green) and \texttt{Rand} (colored magenta) with regard to averaged trace of covariance and MSE. Although \texttt{RATT} is designed to protect against worst-case attacks, it retains its superiority against bounded rational attacks.  

Overall, in the quantitative evaluations above, \texttt{RATT} achieves a superior and close-to-optimal performance against worst-case attacks, and remains superior even against non-worst-case attacks, \textit{i.e.}, bounded rational attack.  

% \subsection{Evaluation over multiple steps} 6 robots
% \subsubsection{Qualitative comparison} 4 robots

\section{Conclusion} ~\label{sec:conclusion}
We worked towards protecting critical multi-robot target tracking from sensing and communication attacks or failures. Particularly, we introduced the first robust planning framework for multi-robot tracking against any fixed numbers of worst-case sensing and communication attacks/failures (Problem~\ref{prob:rob_tar_tracking}). We provided the first robust algorithm, \texttt{RATT} (Algorithm~\ref{alg:rob_tar_track}), for Problem~\ref{prob:rob_tar_tracking}, and proved its suboptimality guarantees using the notations of curvature for set functions. With extensive empirical evaluations, we demonstrate the need for robust target tracking and \texttt{RATT}'s robustness and superiority against both worst-case and non-worst-case attacks. 

A future direction is to extend the proposed framework to the decentralized settings~\cite{zhou2020distributed,liu2021distributed}. With decentralized communication, robustifying the team's connectivity becomes essential to secure against communication attacks/failures. A second avenue is to investigate the proposed framework in an online resilient fashion, \textit{e.g.}, the receding-horizon planning~\cite{schlotfeldt2021resilient}.

% \section*{Acknowledgments}
 \appendices 
\section*{Appendix}~\label{sec:appendix}
We first prove \texttt{RATT}'s approximation bound (Theorem~\ref{thm:ratt_appro}), and then, its running time (Theorem~\ref{thm:ratt_run_time}). 

\subsection{Proof of Theorem~\ref{thm:ratt_appro}}
We prove Theorem~\ref{thm:ratt_appro}, \textit{i.e.}, \texttt{RATT}'s approximation bound by capitalizing on the results in \cite{tzoumas2018resilient} which proves the approximation bound of a robust algorithm against sensing attacks. Here, we focus on the proof of \texttt{RATT}'s approximation bound against sensing and communication attacks. Particularly, we first prove eq.~\eqref{eqn:appro_non_descrease} where $\Phi$ is merely a non-decreasing function, and then, prove eq.~\eqref{eqn:appro_submodular} where $\Phi$ is a non-decreasing and submodular function. We start with describing necessary notations:

% We also use the following notations. 
% To this end, we first prove \texttt{RATT}'s approximation bounds with sensing attacks only (Lemma~\eqref{lem:ratt_appro_sen}) and communication attacks only (Lemma~\eqref{lem:ratt_appro_comm}), respectively. Then, building on that, we prove \texttt{RATT}'s approximation bound with both sensing and communication attacks (Theorem~\ref{thm:ratt_appro}). 

\noindent \textbf{Notation.} We introduce notations to indicate that subsets of robots whose control inputs are computed by different approaches. Specifically, we let $\mc{V}^b$ and $\mc{V}^g$  denote the subsets of robots whose control inputs are computed using the bait selection and the greedy algorithm, respectively. For simplicity, we name $\mc{V}^b$ and $\mc{V}^g$ as \textit{bait robots} and \textit{greedy robots}, respectively. Notably, $\mc{V} = \mc{V}^b \cup \mc{V}^g$, $|\mc{V}^b| = \alpha$, and $|\mc{V}^g| = N- \alpha$ with $\alpha = \alpha_s + \alpha_{c,s}$. For a subset of robots $\mc{W} \subseteq \mc{V}$, we denote $\mb{u}_{\mc{W}}^b$\footnote{For clarify, we drop the subscript for time indices.} and $\mb{u}_{\mc{W}}^g$ as its control inputs selected by the bait selection and the greedy algorithm, respectively. In addition, we let $\mb{u}_{\mc{W}}$ and $\mb{u}_{\mc{W}}^\star$ denote the control inputs selected by \texttt{RATT} and the optimal algorithm for $\mc{W}$. 
% Corresponding, we use $\Phi(\mb{u}_{\mc{W}})$, $\Phi(\mb{u}^b_{\mc{W}})$, $\Phi(\mb{u}^g_{\mc{W}})$,  and $\Phi(\mb{u}_{\mc{W}}^\star)$ to denote the tracking quality using \texttt{RATT}, the bait selection, the greedy algorithm, and the optimal algorithm without attacks in subset $\mc{W}$.

Consider that the $\alpha_c$ communication attacks separate the robot team into $K$ subgroups, each denoted as $\mc{V}_k, k\in \{1,\cdots, K\}$. Clearly, $\mc{V} = \bigcup_{k=1}^{K} \mc{V}_k$. Similarly, in each subgroup $\mc{V}_k$, we use $\mc{V}_k^b$ and $\mc{V}_k^g$ to denote the bait robots and greedy robots, respectively. Notably, $\mc{V}_k = \mc{V}_k^b \cup \mc{V}_k^g$,  $\mc{V}^b = \bigcup_{k=1}^{K} \mc{V}_k^b$, and  $\mc{V}^g = \bigcup_{k=1}^{K} \mc{V}_k^g$. 

% By the notations above, $\Phi(\mb{u}_{\mc{V}_k})$ and $\Phi(\mb{u}_{\mc{V}_k}^\star)$ denote respectively the tracking quality using \texttt{RATT} and the optimal algorithm in each subgroup $\mc{V}_k$; $\Phi(\mb{u}_{\mc{V}_k}) = \Phi(\mb{u}^b_{\mc{V}_k^b}, \mb{u}^g_{\mc{V}_k^g})$. 

% \vspace{3pt}
% \subsubsection{Proof of Lemma~\eqref{lem:ratt_appro_sen}} The proof directly follows the proof of Theorem 1 (eqs. (7), (8)) in \cite{tzoumas2018resilient}. We refer the reader to \cite[Appendix C]{tzoumas2018resilient} for more details. 

% \vspace{3pt}
% \subsubsection{Proof of Lemma~\eqref{lem:ratt_appro_comm}}

Recall that the worst-case communication attacks tend to separate the robot team into more, and smaller subgroups. On this basis, \texttt{RATT} first utilizes \texttt{CAA} (Algorithm~\eqref{alg:comm_appro}) to compute the number of robots in the largest subgroup as $n_{\max}$ and approximates $\alpha_c$ communication attacks as $\alpha_{c,s}= N-n_{\max}$ sensing attacks. Indeed, by \texttt{CAA} and the behavior of the worst-case communication attacks, the number of subgroups and the number of robots in each subgroup after any number of the worst-case communication attacks can be exactly determined. 

% In Lemma~\eqref{lem:ratt_appro_comm}, we consider communication attacks only. 
Then, \texttt{RATT} selects baits control inputs for $\alpha = \alpha_s + \alpha_{c,s}$ robots and the greedy control inputs for the remaining $N - \alpha$ robots by the robust maximization. Notably, $|\mc{V}^b| = |\bigcup_{k=1}^{K}  \mc{V}_k^b| = \alpha$ and $|\mc{V}^g| = |\bigcup_{k=1}^{K}  \mc{V}_k^g| = N - \alpha$.  Also recall that the team's tracking quality is the tracking quality of the subgroup that performs the best (Section~\eqref{subsec:framework}-(f)). Hence, with the worse-case sensing and communication attacks, the team's tracking quality using \texttt{RATT} can be written as:
\begin{equation}~\label{eq:team_subgroup}
    \Phi(\mc{V}\setminus\mc{A}_{s}^\star, \mc{E}\setminus\mc{A}_{c}^\star) = \max_{k\in \{1,\cdots, K\}}\Phi(\mb{u}_{\mc{V}_k \setminus \mc{A}_k^\star}), 
\end{equation}
where $\mc{A}_k^\star$ denotes the worse-case sensing attacks in subgroup $\mc{V}_k$. For clarify, we omit the dependence of tracking quality $\Phi$ on the communication links in each subgroup $\mc{V}_k$, since the robots can communicate in the same subgroup. 

Correspondingly, with the worse-case sensing and communication attacks, the optimal team's tracking quality is: 
\begin{equation}~\label{eq:opt_team_subgroup}
    \Phi_{\mc{V}, \alpha_s, \mc{E},\alpha_c}^\star = \max_{k\in \{1,\cdots, K\}}\Phi(\mb{u}^\star_{\mc{V}_k \setminus \mc{A}_k^\star}). 
\end{equation}

Thus, to evaluate \texttt{RATT}'s approximation bound $\Phi(\mc{V}\setminus\mc{A}_{s}^\star, \mc{E}\setminus\mc{A}_{c}^\star)/\Phi_{\mc{V}, \alpha_s, \mc{E},\alpha_c}^\star$, we need to find out how well it does in each subgroup, \textit{i.e.},  $\Phi(\mb{u}_{\mc{V}_k \setminus \mc{A}_k^\star})/\Phi(\mb{u}^\star_{\mc{V}_k \setminus \mc{A}_k^\star})$.  The challenge is, even though we know the number of robots in each subgroup after $\alpha_c$ worst-case communication attacks, we do not know exactly the robot kinds (\textit{e.g.}, bait and/or greedy robots) and the number of each kind in it. Also, we do not know exactly how $\alpha_s$ worst-case sensing attacking are distributed in these subgroups. To address these issues, we first explore possible \textit{patterns} of robot kinds after the communication attacks, and then, under each possible pattern, we investigate possible \textit{distributions} of the sensing attacks in the subgroups. Based on these possible patterns and distributions, we first prove 
eq.~\eqref{eqn:appro_non_descrease} and then prove eq.~\eqref{eqn:appro_submodular}. 
% we prove Lemma~\eqref{lem:ratt_appro_comm} by comparing \texttt{RATT}'s tracking quality (\textit{i.e.}, eq.~\eqref{eq:team_subgroup}) with the optimal tracking quality (eq.~\eqref{eq:opt_team_subgroup}) in each subgroup. 

\vspace{5pt}
\subsubsection{Proof of \texorpdfstring{eq.~\eqref{eqn:appro_non_descrease}}{}}

We first consider $N > \alpha$. In this instance, $|\mc{V}^g| = N- \alpha > 0$, and thus, there exist some greedy robots. We then consider two patterns of robot kinds after the communication attacks in the following.  
% and case 2: $N \leq \alpha$.
% \vspace{3pt}
% \textbf{Case 1:} When $N > \alpha$, the number of greedy robots is larger than or equal to that of bait robots, \textit{i.e.}, $|\mc{V}^g|\geq |\mc{V}^b|$. In this case, there exist  in general. 

\begin{itemize}
    \item \textbf{Pattern 1:} A subgroup contains $\alpha_s$ bait robots and all $N-\alpha$ greedy robots, denoted as $\mc{V}_1$. Notably, $\mc{V}_1^g = \mc{V}^g$. Other subgroups contain the remaining $\alpha_{c,s}$ bait robots, each denoted as $\mc{V}_{k}, k\in\{2,\cdots, K\}$. Notably, $\bigcup_{k=2}^{K}\mc{V}_k = \mc{V}^b$.  Under Pattern 1, we consider three possible distributions of the $\alpha_s$ worst-case sensing attacks. An example is shown in Fig.~\ref{fig:p1_3d}.
    
    \begin{figure*}[th!]
    \centering{
    \subfigure[Pattern 1, Distribution 1]
    {\includegraphics[width=0.6\columnwidth]{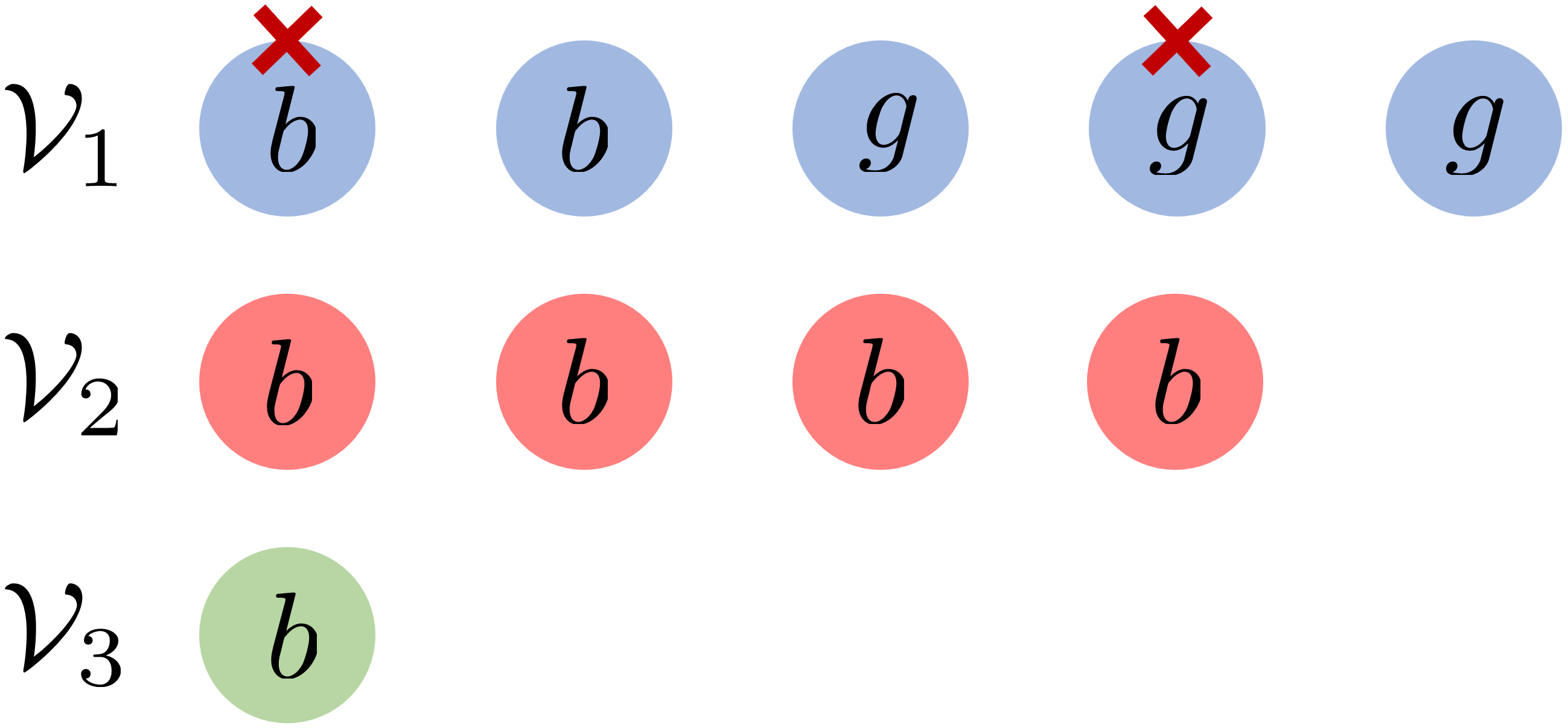}}~~~~
    \subfigure[Pattern 1, Distribution 2] 
    {\includegraphics[width=0.6\columnwidth]{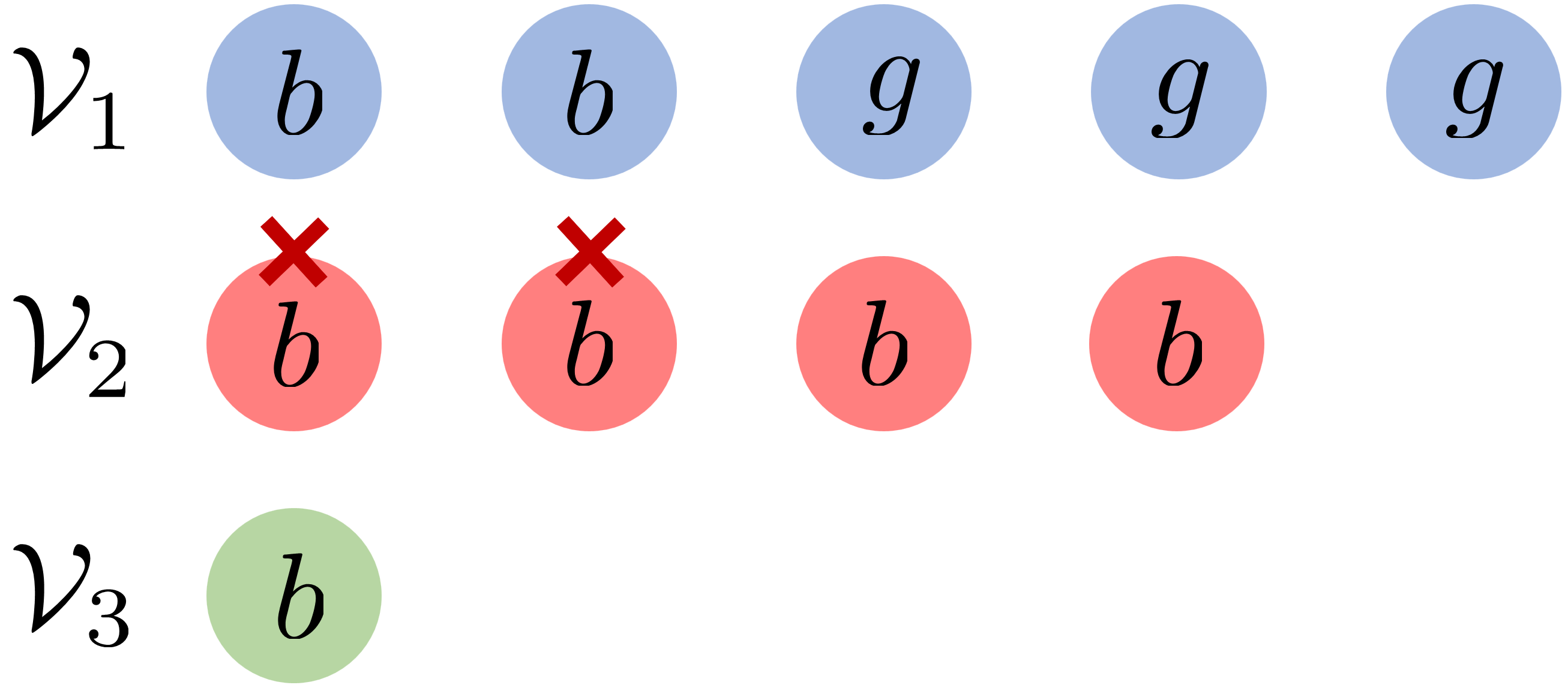}}~~~~
    \subfigure[Pattern 1, Distribution 3] 
    {\includegraphics[width=0.6\columnwidth]{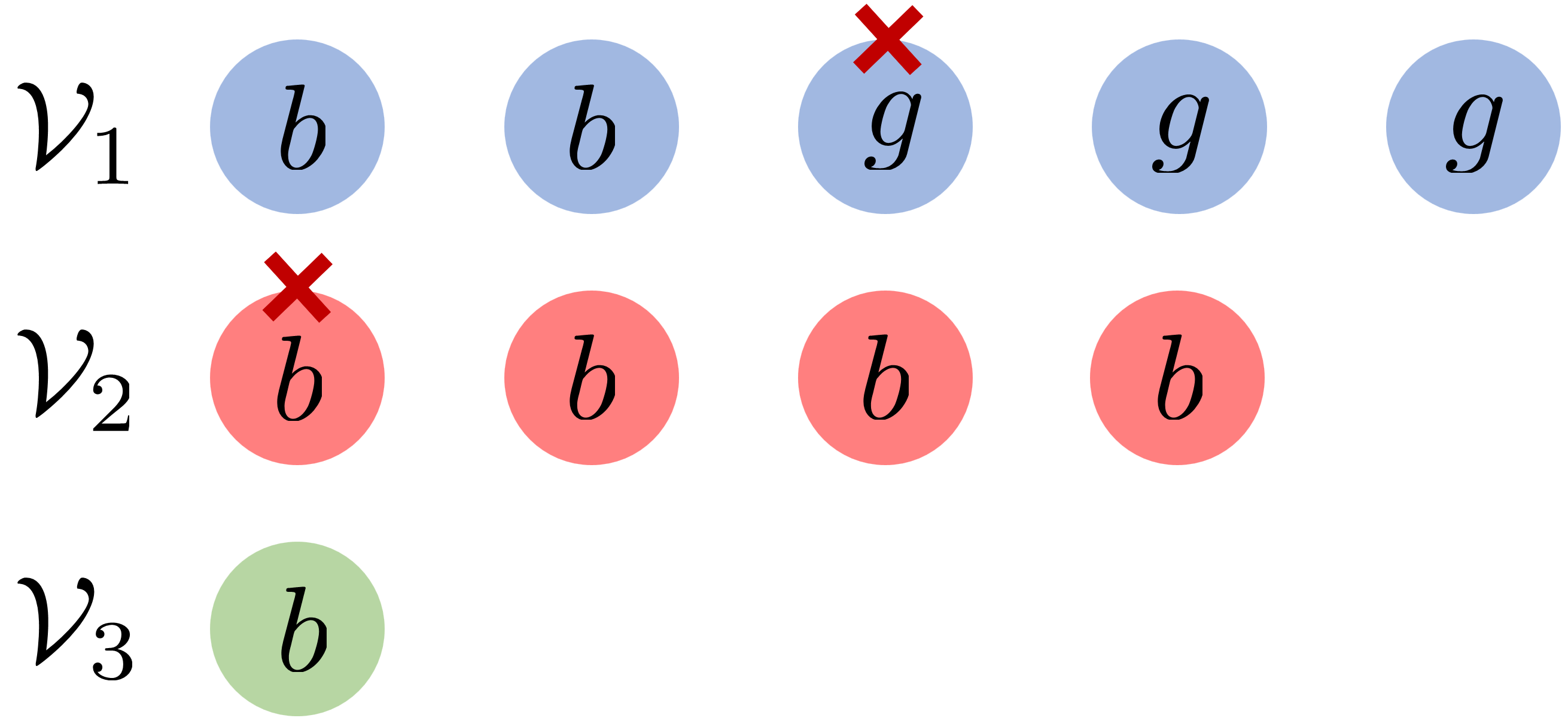}}
    \caption{An illustration of Pattern 1 and 3 possible distributions. Consider a team of $N = 10$ robots encounters $\alpha_s = 2$ worst-case sensing attacks (red crosses) and $\alpha_c = 29$ worst-case sensing attacks. By \texttt{CAA} (Algorithm~\ref{alg:comm_appro}), after the $29$ worst-case communication attacks, the robot team is separated into 3 subgroups, $\mc{V}_1$,  $\mc{V}_2$, $\mc{V}_3$. The robots in the same subgroup have the same color. The communication links in each subgroup are omitted for clarity. The number of robots in the largest clique ($\mc{V}_1$) is $n_{\max} =5$. The approximated number of sensing attacks is $N-n_{\max} =5$. Then the total number of sensing attacks $\alpha = \alpha_s + \alpha_{c,s} = 7$. By \texttt{RATT} (Algorithm~\ref{alg:rob_tar_track}), there are $\alpha = 7$ bait robots (represented by ``$b$'') and $N-\alpha = 3$ greedy robots (represented by ``$g$''). (a), The $\alpha_s =2$ sensing attacks are in $\mc{V}_1$. (b), The $\alpha_s =2$ sensing attacks are in $\mc{V}_2$.  (c), One sensing attack is in $\mc{V}_1$ and the other sensing attack is in $\mc{V}_2$.  
    \label{fig:p1_3d}}
    }
    \end{figure*}
    
    \vspace{3pt} 
    \textbf{-- Distribution 1:} The $\alpha_s$ sensing attacks are in subgroup $\mc{V}_1$ (see Fig.~\ref{fig:p1_3d}-(a)). With Distribution 1, we evaluate \texttt{RATT}'s approximation bound in each subgroup $\mc{V}_k$ as follows. 
    
    \vspace{2pt}
    \textit{i):} In subgroup $\mc{V}_1$, one has $\mc{V}_1 = \mc{V}_1^b \cup \mc{V}_1^g$. Notably, $|\mc{V}_1^b| = \alpha_s$, $|\mc{V}_1^g| = N - \alpha$, and $|\mc{A}_1^\star| = \alpha_s$. We evaluate \texttt{RATT}'s approximation bound, $\Phi(\mb{u}_{\mc{V}_1 \setminus \mc{A}_1^\star})/\Phi(\mb{u}^\star_{\mc{V}_1 \setminus \mc{A}_1^\star})$, in  $\mc{V}_1$ by following \cite[Proof of ineq. (8)]{tzoumas2018resilient}. Particularly, 
    % To this end, we use the additional notations as follows: 
    
    \begin{align}
        & \Phi(\mb{u}_{\mc{V}_1 \setminus \mc{A}_1^\star}) = \Phi(\mb{u}_{(\mc{V}_1^b \setminus \mc{A}_1^\star) \cup (\mc{V}_1^g \setminus \mc{A}_1^\star)}), ~\label{eq:nd_p1d1v1_1}\\
        & \geq (1-c_\Phi) \sum_{i \in (\mc{V}_1^b \setminus \mc{A}_1^\star) \cup (\mc{V}_1^g \setminus \mc{A}_1^\star)} \Phi(\mb{u}_{i}), ~\label{eq:nd_p1d1v1_2}\\
        & = (1-c_\Phi) [\sum_{i \in \mc{V}_1^b \setminus \mc{A}_1^\star} \Phi(\mb{u}^b_{i}) + \sum_{i \in \mc{V}_1^g \setminus \mc{A}_1^\star} \Phi(\mb{u}^g_{i})], ~\label{eq:nd_p1d1v1_3}\\
        & \geq (1-c_\Phi) [\sum_{i \in \mc{V}_1^g\setminus (\mc{V}_1^g \setminus \mc{A}_1^\star) } \Phi(\mb{u}^g_{i}) + \sum_{i \in \mc{V}_1^g \setminus \mc{A}_1^\star} \Phi(\mb{u}^g_{i})], ~\label{eq:nd_p1d1v1_4}\\
        & = (1-c_\Phi) \sum_{i \in \mc{V}_1^g} \Phi(\mb{u}^g_{i}), ~\label{eq:nd_p1d1v1_5}\\
        & \geq (1-c_\Phi)^2  \Phi(\mb{u}^g_{\mc{V}_1^g}), ~\label{eq:nd_p1d1v1_6} \\    
        & \geq (1-c_\Phi)^3  \Phi(\mb{u}^\star_{\mc{V}_1^g}), ~\label{eq:nd_p1d1v1_7} \\
        % \forall \mc{W}\subseteq \mc{V}_1, |\mc{W}| = |\mc{V}_1^g|, ~\label{eq:nd_p1d1v1_7} \\
        & \geq (1-c_\Phi)^3 \Phi(\mb{u}^\star_{\mc{V}_1 \setminus \mc{A}_1^\star}). ~\label{eq:nd_p1d1v1_8}
    \end{align}
    Eqs.~\eqref{eq:nd_p1d1v1_1} - \eqref{eq:nd_p1d1v1_8} hold for the following reasons. Eq.~\eqref{eq:nd_p1d1v1_1} holds since $\mc{V}_1 = \mc{V}_1^b + \mc{V}_1^g$. Ineq.~\eqref{eq:nd_p1d1v1_2} follows from eq.~\eqref{eq:nd_p1d1v1_1} due to \cite[Lemma 4]{tzoumas2018resilient}, which is derived based on the definition of the total curvature (eq.~\eqref{eq:total_curvature}). Eq.~\eqref{eq:nd_p1d1v1_3} follows from ineq.~\eqref{eq:nd_p1d1v1_2} due to $(\mc{V}_1^b \setminus \mc{A}_1^\star) \cap (\mc{V}_1^b \setminus \mc{A}_1^\star) = \emptyset$. Ineq.~\eqref{eq:nd_p1d1v1_4} holds because: first, since $\mc{V}_1^b$ are bait robots and $\mc{V}_1^g$ are greedy robots,  $\Phi(\mb{u}^b_{i}) \geq \Phi(\mb{u}^g_{i'})$ for any $i\in\mc{V}_1^b\setminus \mc{A}_1^\star$ and $i'\in\mc{V}_1^g \setminus \mc{A}_1^\star$; second, since $\mc{V}_1^b\setminus \mc{A}_1^\star$ and $\mc{V}_1^g\setminus \mc{A}_1^\star$ denote respectively the set of non-attacked robots in $\mc{V}_1^b$ and the set of non-attacked robots in $\mc{V}_1^g$ after the worst-case sensing attacks $\mc{A}_1^\star$, one has $|\mc{V}_1^b \cap \mc{A}_1^\star| + |\mc{V}_1^g \cap \mc{A}_1^\star| = \alpha_s$, and thus $|\mc{V}_1^b\setminus \mc{A}_1^\star| = \alpha_s - |\mc{V}_1^b \cap \mc{A}_1^\star| = |\mc{V}_1^g \cap \mc{A}_1^\star| = |\mc{V}_1^g\setminus (\mc{V}_1^g \setminus \mc{A}_1^\star)|$. Eq.~\eqref{eq:nd_p1d1v1_5} follows naturally from ineq.~\eqref{eq:nd_p1d1v1_4}. Ineq.~\eqref{eq:nd_p1d1v1_6} follows from eq.~\eqref{eq:nd_p1d1v1_5} due to \cite[Corollary 1]{tzoumas2018resilient}, which is derived by using the property of the total curvature (eq.~\eqref{eq:total_curvature}). Ineq.~\eqref{eq:nd_p1d1v1_7} follows from ineq.~\eqref{eq:nd_p1d1v1_6} due to \cite[Theorem 8.1]{sviridenko2017optimal}, which implies that the greedy algorithm, introduced in \cite[Section 2]{fisher1978analysis}, achieves $1-c_\Phi$ approximation bound for optimizing non-decreasing functions. Ineq.~\eqref{eq:nd_p1d1v1_8} follows from ineq.~\eqref{eq:nd_p1d1v1_7} due to \cite[Lemma 9]{tzoumas2018resilient}. 
    
    \vspace{2pt}
    \textit{ii):} In each of the other subgroups $\mc{V}_{k}, k\in\{2,\cdots, K\}$, there exist no sensing attacks ($\mc{A}_k^\star = \emptyset$) and the robots' control inputs are bait control inputs. Hence,
    \begin{align}
        \Phi(\mb{u}_{\mc{V}_k \setminus \mc{A}_k^\star}) &= \Phi(\mb{u}^b_{\mc{V}_k}) ~\label{eq:nd_p1d1vk_1}\\
        & \geq (1-c_\Phi) \sum_{i \in \mc{V}_k} \Phi(\mb{u}^b_{i}), ~\label{eq:nd_p1d1vk_2}\\
        & = (1-c_\Phi) \sum_{i \in \mc{V}_k} \Phi(\mb{u}_{i}^\star), ~\label{eq:nd_p1d1vk_3} \\
        & \geq (1-c_\Phi)^2 \Phi(\mb{u}^\star_{\mc{V}_k}). ~\label{eq:nd_p1d1vk_4} \\
        & = (1-c_\Phi)^2 \Phi(\mb{u}^\star_{\mc{V}_k\setminus \mc{A}_k^\star}). ~\label{eq:nd_p1d1vk_5}
    \end{align}
    Eqs.~\eqref{eq:nd_p1d1vk_1} - \eqref{eq:nd_p1d1vk_5} hold for the following reasons. Eq.~\eqref{eq:nd_p1d1vk_1} holds since $\mc{A}_k^\star = \emptyset$ and $\mc{V}_k = \mc{V}_k^b$. 
    Ineq.~\eqref{eq:nd_p1d1vk_2} follows from eq.~\eqref{eq:nd_p1d1vk_1} due to \cite[Lemma 4]{tzoumas2018resilient}.  Eq.~\eqref{eq:nd_p1d1vk_3} follows from ineq.~\eqref{eq:nd_p1d1vk_2} since each bait control input $\mb{u}_{i}^b$ gives the maximum individual tracking quality. Ineq.~\eqref{eq:nd_p1d1vk_4} follows from eq.~\eqref{eq:nd_p1d1vk_3}  due to \cite[Corollary 1]{tzoumas2018resilient} and $\mc{A}_k^\star = \emptyset$. Ineq.~\eqref{eq:nd_p1d1vk_5} follows from ineq.~\eqref{eq:nd_p1d1vk_4} due to $\mc{A}_k^\star = \emptyset$. 
    
    \vspace{2pt}
    To sum up, in Pattern 1, Distribution 1, by eqs.~\eqref{eq:nd_p1d1v1_1} - \eqref{eq:nd_p1d1v1_8} and eqs.~\eqref{eq:nd_p1d1vk_1}-\eqref{eq:nd_p1d1vk_5}, one has
    \begin{align*}
            &\max_{k\in \{1,\cdots, K\}}\Phi(\mb{u}_{\mc{V}_k\setminus \mc{A}_k^\star}) \\
            &\geq \min[(1-c_\Phi)^2, (1-c_\Phi)^3] \max_{k\in \{1,\cdots, K\}}\Phi(\mb{u}^\star_{\mc{V}_k\setminus \mc{A}_k^\star}).
    \end{align*}
    Therefore, along with eqs.~\eqref{eq:team_subgroup} and \eqref{eq:opt_team_subgroup}, one has
    \begin{equation}\label{eq:nd_p1_d1}
        \Phi(\mc{V}\setminus\emptyset, \mc{E}\setminus\mc{A}_{c}^\star) \geq (1-c_\Phi)^3 \Phi_{\mc{V}, \alpha_s, \mc{E},\alpha_c}^\star. 
    \end{equation}

    \vspace{3pt} 
    \textbf{-- Distribution 2:} The $\alpha_s$ sensing attacks are in subgroups $\{\mc{V}_{k}\}_{k=2}^{K}$ (see Fig.~\ref{fig:p1_3d}-(b)). 
    With Distribution 2, we evaluate \texttt{RATT}'s approximation bound in each subgroup $\mc{V}_k$ as follows. 
    
    \vspace{2pt}
    \textit{i):} In subgroup $\mc{V}_1$, one has $\mc{V}_1 = \mc{V}_1^b \cup \mc{V}_1^g$. Notably, $|\mc{V}_1^b| = \alpha_s$,  $|\mc{V}_1^g| = N - \alpha$, and $\mc{A}_1^\star = \emptyset$. Hence, 
    
    \begin{align}
        & \Phi(\mb{u}_{\mc{V}_1 \setminus \mc{A}_1^\star}) = \Phi(\mb{u}_{\mc{V}_1^b \cup \mc{V}_1^g}), ~\label{eq:nd_p1d2v1_1}\\
        & \geq (1-c_\Phi) \sum_{i \in \mc{V}_1^b  \cup \mc{V}_1^g} \Phi(\mb{u}_{i}), ~\label{eq:nd_p1d2v1_2}\\
        & = (1-c_\Phi) [\sum_{i \in \mc{V}_1^b} \Phi(\mb{u}^b_{i}) + \sum_{i \in \mc{V}_1^g} \Phi(\mb{u}^g_{i})], ~\label{eq:nd_p1d2v1_3}\\
        & = (1-c_\Phi) [\sum_{i \in \mc{V}_1^b} \Phi(\mb{u}^\star_{i}) + \sum_{i \in \mc{V}_1^g} \Phi(\mb{u}^g_{i})], ~\label{eq:nd_p1d2v1_4}\\
        & \geq (1-c_\Phi)^2 [\Phi(\mb{u}^\star_{\mc{V}_1^b}) + \Phi(\mb{u}^g_{\mc{V}_1^g})],  ~\label{eq:nd_p1d2v1_5}\\
        & \geq (1-c_\Phi)^2 \Phi(\mb{u}^\star_{\mc{V}_1^b}) +  (1-c_\Phi)^3 \Phi(\mb{u}^\star_{\mc{V}_1^g}),  ~\label{eq:nd_p1d2v1_6}\\
        & \geq (1-c_\Phi)^3 [\Phi(\mb{u}^\star_{\mc{V}_1^b}) +  \Phi(\mb{u}^\star_{\mc{V}_1^g})], ~\label{eq:nd_p1d2v1_7}\\
        & \geq (1-c_\Phi)^3 \Phi(\mb{u}^\star_{\mc{V}_1}), ~\label{eq:nd_p1d2v1_8} \\
        & = (1-c_\Phi)^3 \Phi(\mb{u}^\star_{\mc{V}_1 \setminus \mc{A}_1^\star}). ~\label{eq:nd_p1d2v1_9}
    \end{align}
    Eqs.~\eqref{eq:nd_p1d2v1_1} - \eqref{eq:nd_p1d2v1_9} hold for the following reasons. Eq.~\eqref{eq:nd_p1d2v1_1} holds since $\mc{A}_1^\star = \emptyset$ and $\mc{V}_1 = \mc{V}_1^b \cup \mc{V}_1^g$.  Ineq.~\eqref{eq:nd_p1d2v1_2} follows from eq.~\eqref{eq:nd_p1d2v1_1} due to \cite[Lemma 4]{tzoumas2018resilient}. Eq.~\eqref{eq:nd_p1d2v1_3} naturally follows from ineq.~\eqref{eq:nd_p1d2v1_2} due to $ \mc{V}_1^b \cap \mc{V}_1^g = \emptyset$. Eq.~\eqref{eq:nd_p1d2v1_4} follows from eq.~\eqref{eq:nd_p1d2v1_3} since each bait control input $\mb{u}_{i}^b$ gives the maximum individual tracking quality. Ineq.~\eqref{eq:nd_p1d2v1_5} follows from eq.~\eqref{eq:nd_p1d2v1_4} due to \cite[Corollary 1]{tzoumas2018resilient}. Ineq.~\eqref{eq:nd_p1d2v1_6} follows from ineq.~\eqref{eq:nd_p1d2v1_5} due to \cite[Theorem 8.1]{sviridenko2017optimal}, which implies that the greedy algorithm~\cite[Section 2]{fisher1978analysis} achieves $1-c_\Phi$ approximation bound for optimizing non-decreasing functions, and thus $\Phi(\mb{u}^g_{\mc{V}_1^g}) \geq (1-c_\Phi) \Phi(\mb{u}^\star_{\mc{V}_1^g})$. Ineq.~\eqref{eq:nd_p1d2v1_7} follows from ineq.~\eqref{eq:nd_p1d2v1_6} due to $1-c_\Phi \in [0,1]$. Ineq.~\eqref{eq:nd_p1d2v1_8} follows from ineq.~\eqref{eq:nd_p1d2v1_7} due to $\mc{V}_1^b, \mc{V}_1^g \subseteq \mc{V}_1$, $\mc{V}_1^b \cap \mc{V}_1^g = \emptyset$, and thus $\Phi(\mb{u}^\star_{\mc{V}_1^b}) +  \Phi(\mb{u}^\star_{\mc{V}_1^g}) \geq \Phi(\mb{u}^\star_{\mc{V}_1})$. Eq.~\eqref{eq:nd_p1d2v1_9} follows from ineq.~\eqref{eq:nd_p1d2v1_8} due to $\mc{A}_1^\star = \emptyset$. 
    
    \vspace{2pt}
    \textit{ii):} In the other subgroups $\{\mc{V}_{k}\}_{k=2}^{K}$, even though there exist $\alpha_s$ sensing attacks, the control inputs of the non-attacked robots are still bait control inputs. Hence, in each subgroup $\mc{V}_{k}$, one has, 
    \begin{align}
        \Phi(\mb{u}_{\mc{V}_k \setminus \mc{A}_k^\star}) &= \Phi(\mb{u}^b_{\mc{V}_k\setminus \mc{A}_k^\star}) ~\label{eq:nd_p1d2vk_1}\\
        & \geq (1-c_\Phi) \sum_{i \in \mc{V}_k\setminus \mc{A}_k^\star} \Phi(\mb{u}^b_{i}), ~\label{eq:nd_p1d2vk_2}\\
        & = (1-c_\Phi) \sum_{i \in \mc{V}_k\setminus \mc{A}_k^\star} \Phi(\mb{u}_{i}^\star), ~\label{eq:nd_p1d2vk_3} \\
        & \geq (1-c_\Phi)^2 \Phi(\mb{u}^\star_{\mc{V}_k\setminus \mc{A}_k^\star}). ~\label{eq:nd_p1d2vk_4}
    \end{align}
    Eqs.~\eqref{eq:nd_p1d2vk_1} - \eqref{eq:nd_p1d2vk_4} hold for the following reasons. Eq.~\eqref{eq:nd_p1d2vk_1} holds since $\mc{V}_k = \mc{V}_k^b$. 
    Ineq.~\eqref{eq:nd_p1d2vk_2} follows from eq.~\eqref{eq:nd_p1d2vk_1} due to \cite[Lemma 4]{tzoumas2018resilient}.  Eq.~\eqref{eq:nd_p1d2vk_3} follows from ineq.~\eqref{eq:nd_p1d2vk_2} since each bait control input $\mb{u}_{i}^b$ gives the maximum individual tracking quality. Ineq.~\eqref{eq:nd_p1d2vk_4} follows from eq.~\eqref{eq:nd_p1d2vk_3}  due to \cite[Corollary 1]{tzoumas2018resilient}.
    
    \vspace{2pt}
    To sum up, in Pattern 1, Distribution 2, by eqs.~\eqref{eq:nd_p1d2v1_1} - \eqref{eq:nd_p1d2v1_9} and eqs.~\eqref{eq:nd_p1d2vk_1}-\eqref{eq:nd_p1d2vk_4}, one has
    \begin{align*}
            &\max_{k\in \{1,\cdots, K\}}\Phi(\mb{u}_{\mc{V}_k\setminus \mc{A}_k^\star}) \\
            &\geq \min[(1-c_\Phi)^2, (1-c_\Phi)^3] \max_{k\in \{1,\cdots, K\}}\Phi(\mb{u}^\star_{\mc{V}_k\setminus \mc{A}_k^\star}).
    \end{align*}
    Therefore, along with eqs.~\eqref{eq:team_subgroup} and \eqref{eq:opt_team_subgroup}, one has
    \begin{equation}\label{eq:nd_p1_d2}
        \Phi(\mc{V}\setminus\emptyset, \mc{E}\setminus\mc{A}_{c}^\star) \geq (1-c_\Phi)^3 \Phi_{\mc{V}, \alpha_s, \mc{E},\alpha_c}^\star. 
    \end{equation}    
    
    \vspace{3pt} 
    \textbf{-- Distribution 3:} Part of the $\alpha_s$ sensing attacks (\textit{i.e.}, $\alpha_s^1$, $\alpha_s^1 < \alpha_s$) is in subgroup $\mc{V}_1$ and the remaining part (\textit{i.e.}, $\alpha_s - \alpha_s^1$) is in subgroups $\{\mc{V}_{k}\}_{k=2}^{K}$ (see Fig.~\ref{fig:p1_3d}-(c)). 
    
    \vspace{2pt}
    \textit{i):} In subgroup $\mc{V}_1$, one has $\mc{V}_1 = \mc{V}_1^b \cup \mc{V}_1^g$. Notably, $|\mc{V}_1^b| = \alpha_s$,  $|\mc{V}_1^g| = N - \alpha$, and $|\mc{A}_1^\star| = \alpha_s^1$. Since the number of sensing attacks $\alpha_{s}^1$ in $\mc{V}_1$ is less than $\alpha_s$, there must exist a subset of $\alpha_s - \alpha_{s}^1$ robots that does not encounter sensing attacks in $\mc{V}_1^b$. We denote this subset as $\mc{V}_1^{b1}\subseteq \mc{V}_1^{b}$ with $|\mc{V}_1^{b1}| = \alpha-\alpha_s^1$. In addition, denote the other robots in $\mc{V}_1^{b1}$ as $\mc{V}_1^{b2}= \mc{V}_1^{b}\setminus \mc{V}_1^{b1}$. Notably, $|\mc{V}_1^{b2}| = \alpha_s^1$. Then, we evaluate \texttt{RATT}'s approximation bound, $\Phi(\mb{u}_{\mc{V}_1 \setminus \mc{A}_1^\star})/\Phi(\mb{u}^\star_{\mc{V}_1 \setminus \mc{A}_1^\star})$, in  $\mc{V}_1$ as follows.
    
    \begin{align}
        & \Phi(\mb{u}_{\mc{V}_1 \setminus \mc{A}_1^\star}) = \Phi(\mb{u}_{(\mc{V}_1^b \setminus \mc{A}_1^\star) \cup (\mc{V}_1^g \setminus \mc{A}_1^\star)}), ~\label{eq:nd_p1d3v1_1}\\
        & \geq (1-c_\Phi) \sum_{i \in (\mc{V}_1^b \setminus \mc{A}_1^\star) \cup (\mc{V}_1^g \setminus \mc{A}_1^\star)} \Phi(\mb{u}_{i}), ~\label{eq:nd_p1d3v1_2}\\
        & = (1-c_\Phi) [\sum_{i \in \mc{V}_1^{b1}} \Phi(\mb{u}^b_{i}) + \nonumber\\ & \hspace{2cm} \sum_{i \in \mc{V}_1^{b2} \setminus \mc{A}_1^\star} \Phi(\mb{u}^b_{i}) + \sum_{i \in \mc{V}_1^g \setminus \mc{A}_1^\star} \Phi(\mb{u}^g_{i})], ~\label{eq:nd_p1d3v1_3}\\
        & \geq (1-c_\Phi)^2 \Phi(\mb{u}^\star_{\mc{V}_1^{b1}}) + (1-c_\Phi)^3 \Phi(\mb{u}^\star_{\mc{V}_1^{g}}),  ~\label{eq:nd_p1d3v1_4}\\
        & \geq (1-c_\Phi)^3 [\Phi(\mb{u}^\star_{\mc{V}_1^{b1}}) + \Phi(\mb{u}^\star_{\mc{V}_1^{g}})],  ~\label{eq:nd_p1d3v1_5}\\ 
        & \geq (1-c_\Phi)^3 \Phi(\mb{u}^\star_{\mc{V}_1^{b1} \cup \mc{V}_1^{g}}),  ~\label{eq:nd_p1d3v1_6}\\ 
        & \geq (1-c_\Phi)^3 \Phi(\mb{u}^\star_{\mc{V}_1 \setminus \mc{A}_1^\star}). ~\label{eq:nd_p1d3v1_7}
    \end{align}
    Eqs.~\eqref{eq:nd_p1d3v1_1} - \eqref{eq:nd_p1d3v1_7} hold for the following reasons. Eq.~\eqref{eq:nd_p1d3v1_1} holds since $\mc{V}_1 = \mc{V}_1^b + \mc{V}_1^g$. Ineq.~\eqref{eq:nd_p1d3v1_2} follows from eq.~\eqref{eq:nd_p1d3v1_1} due to \cite[Lemma 4]{tzoumas2018resilient}. Eq.~\eqref{eq:nd_p1d3v1_3} follows from ineq.~\eqref{eq:nd_p1d3v1_2} due to the definitions of $\mc{V}_1^{b1}$ and $\mc{V}_1^{b2}$. Ineq.~\eqref{eq:nd_p1d3v1_4} follows from eq.~\eqref{eq:nd_p1d3v1_3} because: first, $\sum_{i \in \mc{V}_1^{b1}} \Phi(\mb{u}^b_{i}) \geq (1-c_\Phi) \Phi(\mb{u}^\star_{\mc{V}_1^{b1}})$ by following the steps in eqs.~\eqref{eq:nd_p1d1vk_1} - \eqref{eq:nd_p1d1vk_4}; second, $\sum_{i \in \mc{V}_1^{b2} \setminus \mc{A}_1^\star} \Phi(\mb{u}^b_{i}) + \sum_{i \in \mc{V}_1^g \setminus \mc{A}_1^\star} \Phi(\mb{u}^g_{i}) \geq  (1-c_\Phi)^2 \Phi(\mb{u}^\star_{\mc{V}_1^{g}})$ by following the steps in eqs.~\eqref{eq:nd_p1d1v1_3} - \eqref{eq:nd_p1d1v1_7} since $|\mc{V}_1^{b1}\setminus \mc{A}_1^\star| = |\mc{V}_1^g \setminus (\mc{V}_1^g \setminus \mc{A}_1^\star)|$. Ineq.~\eqref{eq:nd_p1d3v1_5} follows from ineq.~\eqref{eq:nd_p1d3v1_4} due to $1-c_\Phi \in [0,1]$. Ineq.~\eqref{eq:nd_p1d3v1_6} follows from ineq.~\eqref{eq:nd_p1d3v1_5} due to $\mc{V}_1^{b1}, \mc{V}_1^{g} \subseteq \mc{V}_1^{b1} \cup \mc{V}_1^{g}$ and $\mc{V}_1^{b1} \cap \mc{V}_1^{g} = \emptyset$. Ineq.~\eqref{eq:nd_p1d3v1_7} follows from ineq.~\eqref{eq:nd_p1d3v1_6} due to \cite[Lemma 9]{tzoumas2018resilient}. 
    
   \vspace{2pt}
    \textit{ii):} In the other subgroups $\{\mc{V}_{k}\}_{k=2}^{K}$, even though there exist $\alpha_s - \alpha_s^1$ sensing attacks, the control inputs of the non-attacked robots are still bait control inputs. This scenario is the same as Pattern 1, Distribution 2-b). Hence, in each subgroup $\mc{V}_{k}$, one has, 
    \begin{align}
        \Phi(\mb{u}_{\mc{V}_k \setminus \mc{A}_k^\star}) \geq (1-c_\Phi)^2 \Phi(\mb{u}^\star_{\mc{V}_k\setminus \mc{A}_k^\star}). ~\label{eq:nd_p1d3vk}
    \end{align}
    
    \vspace{2pt}
    To sum up, in Pattern 1, Distribution 3, by eqs.~\eqref{eq:nd_p1d3v1_1} - \eqref{eq:nd_p1d3v1_7} and eq.~\eqref{eq:nd_p1d3vk}, one has
    \begin{align*}
            &\max_{k\in \{1,\cdots, K\}}\Phi(\mb{u}_{\mc{V}_k\setminus \mc{A}_k^\star}) \\
            &\geq \min[(1-c_\Phi)^2, (1-c_\Phi)^3] \max_{k\in \{1,\cdots, K\}}\Phi(\mb{u}^\star_{\mc{V}_k\setminus \mc{A}_k^\star}).
    \end{align*}
    Therefore, along with eqs.~\eqref{eq:team_subgroup} and \eqref{eq:opt_team_subgroup}, one has
    \begin{equation}\label{eq:nd_p1_d3}
        \Phi(\mc{V}\setminus\emptyset, \mc{E}\setminus\mc{A}_{c}^\star) \geq (1-c_\Phi)^3 \Phi_{\mc{V}, \alpha_s, \mc{E},\alpha_c}^\star. 
    \end{equation}    
    
    \vspace{3pt}
    Overall, in Pattern 1, by eqs.~\eqref{eq:nd_p1_d1}, \eqref{eq:nd_p1_d2}, and \eqref{eq:nd_p1_d3}, one has:
    \begin{equation}\label{eq:nd_p1}
        \Phi(\mc{V}\setminus\emptyset, \mc{E}\setminus\mc{A}_{c}^\star) \geq (1-c_\Phi)^3 \Phi_{\mc{V}, \alpha_s, \mc{E},\alpha_c}^\star.
    \end{equation}
    
    \begin{figure*}[th!]
    \centering{
    \subfigure[Pattern 2, Distribution 1]
    {\includegraphics[width=0.56\columnwidth]{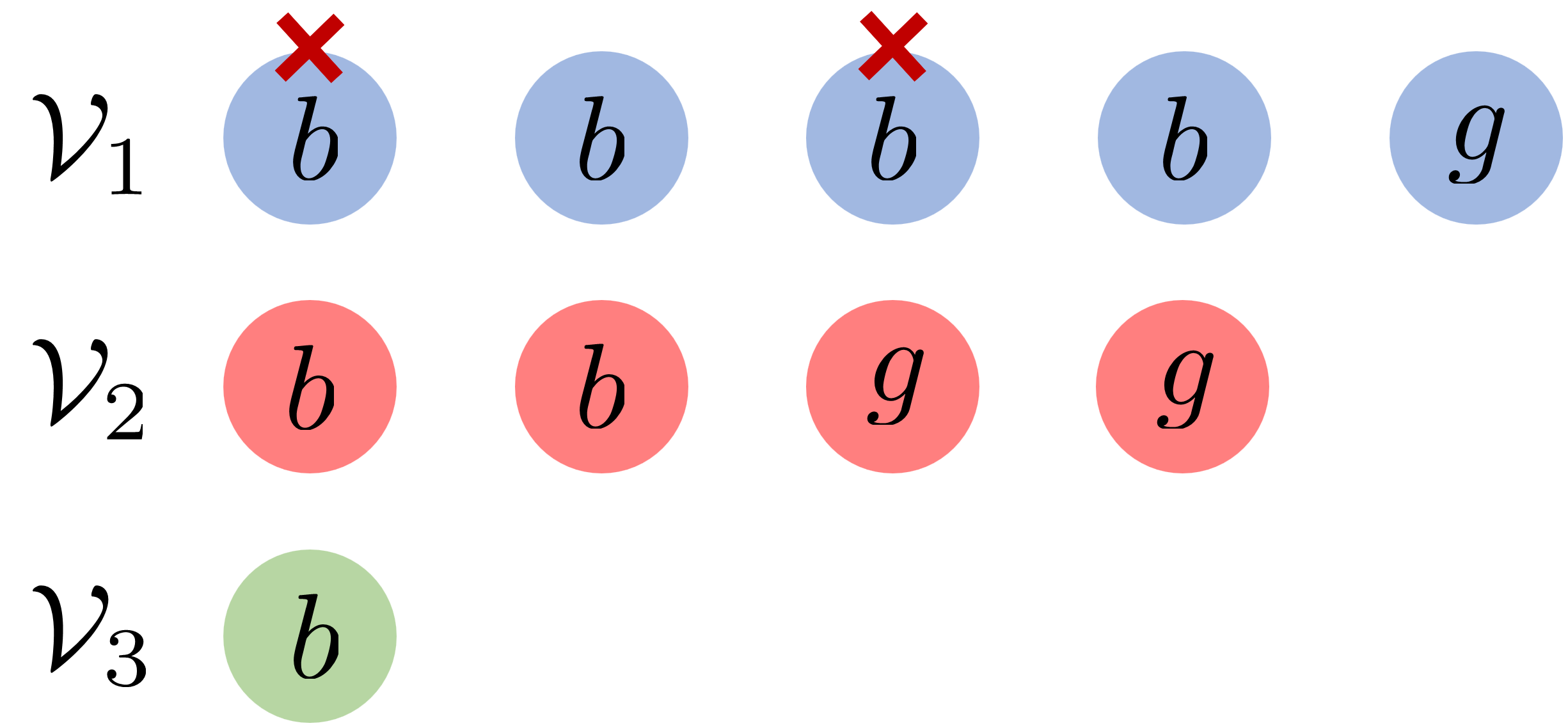}}~~~~
    \subfigure[Pattern 2, Distribution 2] 
    {\includegraphics[width=0.70\columnwidth]{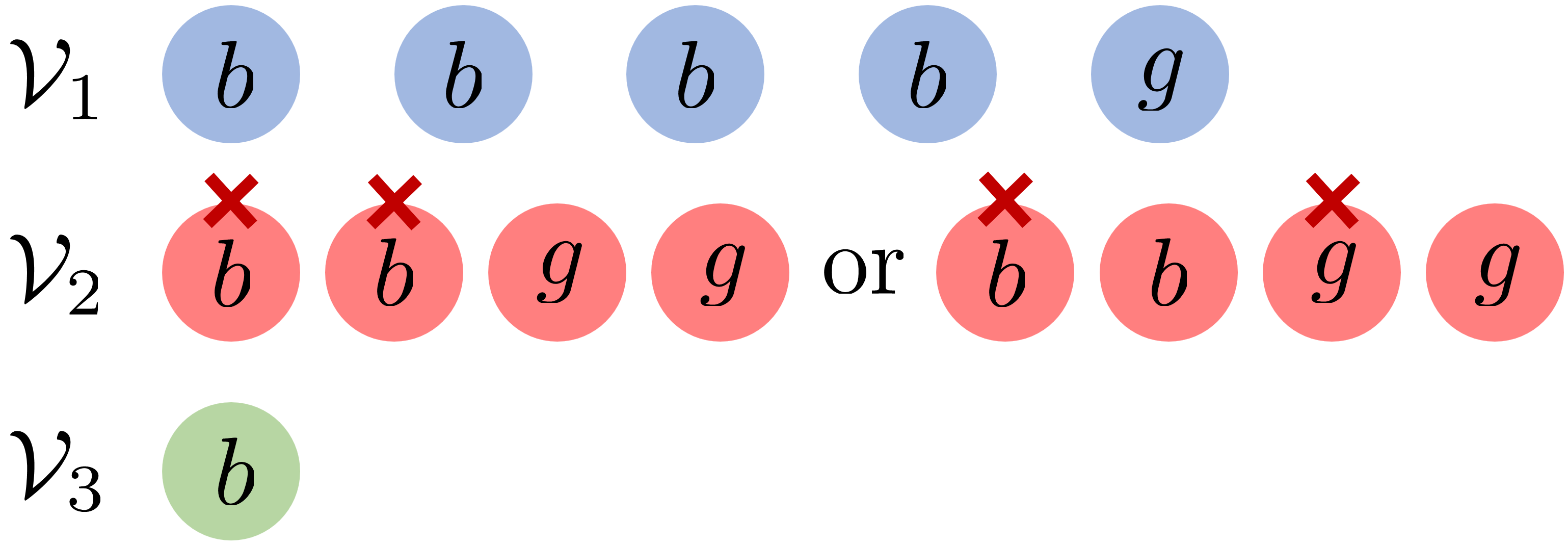}}~~~~
    \subfigure[Pattern 2, Distribution 3] 
    {\includegraphics[width=0.55\columnwidth]{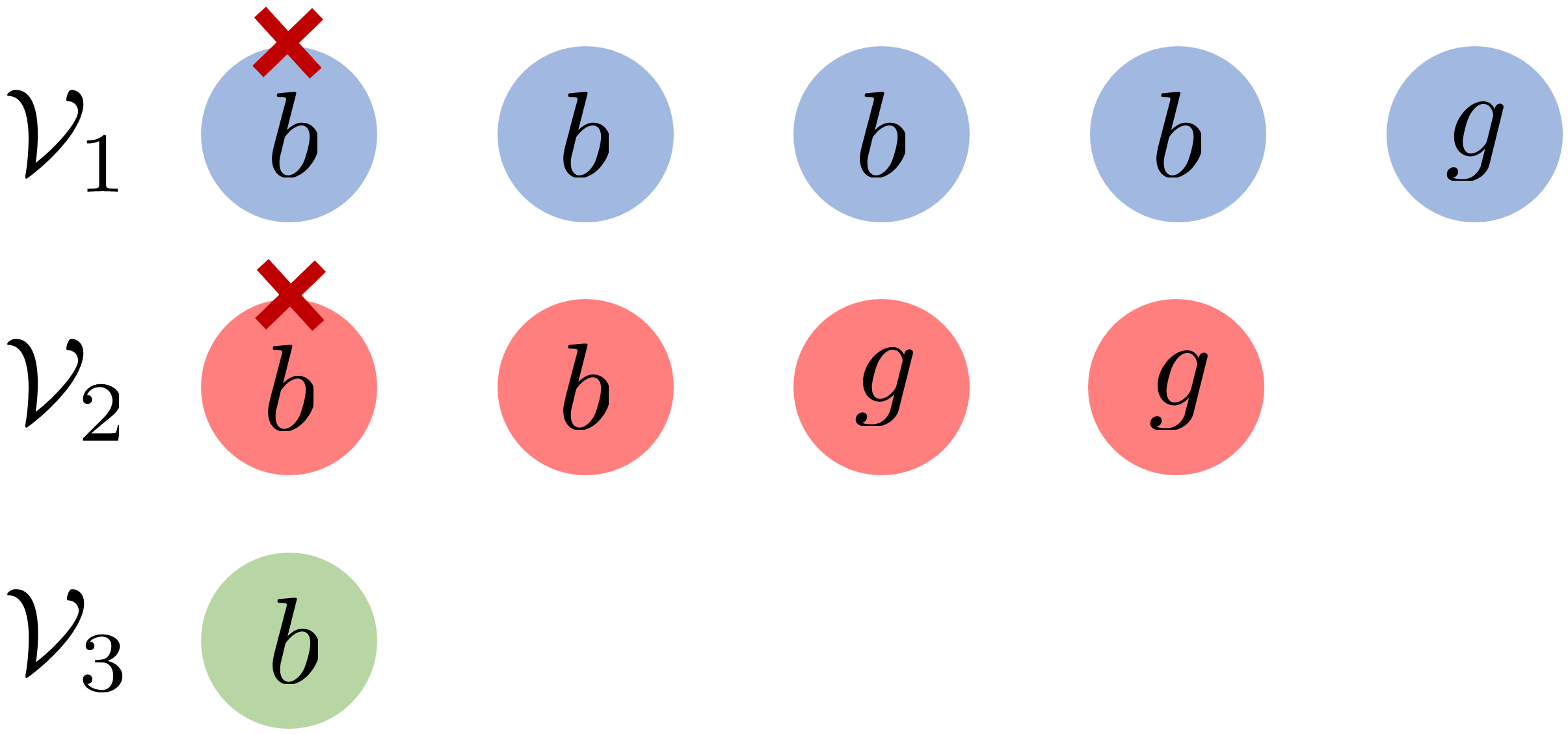}}
    \caption{An illustration of Pattern 2 and 3 possible distributions. Consider a team of $N = 10$ robots encounters $\alpha_s = 2$ worst-case sensing attacks (red crosses) and $\alpha_c = 29$ worst-case sensing attacks. By \texttt{CAA} (Algorithm~\ref{alg:comm_appro}), after the $29$ worst-case communication attacks, the robot team is separated into 3 subgroups, $\mc{V}_1$,  $\mc{V}_2$, $\mc{V}_3$. The robots in the same subgroup have the same color. The communication links in each subgroup are omitted for clarity. The number of robots in the largest clique ($\mc{V}_1$) is $n_{\max} =5$. The approximated number of sensing attacks is $N-n_{\max} =5$. Then the total number of sensing attacks $\alpha = \alpha_s + \alpha_{c,s} = 7$. By \texttt{RATT} (Algorithm~\ref{alg:rob_tar_track}), there are $\alpha = 7$ bait robots (represented by ``$b$'') and $N-\alpha = 3$ greedy robots (represented by ``$g$''). (a), The $\alpha_s =2$ sensing attacks are in $\mc{V}_1$. (b), The $\alpha_s =2$ sensing attacks are in $\mc{V}_2$.  (c), One sensing attack is in $\mc{V}_1$ and the other sensing attack is in $\mc{V}_2$.    
    \label{fig:p2_3d}}
    }
    \end{figure*}

    \item \textbf{Pattern 2:} Two subgroups, denoted by $\mc{V}_1$ and $\mc{V}_2$, contain all $N-\alpha$ greedy robots and some bait robots. That is, $|\mc{V}_1^g| + |\mc{V}_2^g| = N-\alpha$ and $|\mc{V}_1^b| + |\mc{V}_2^b| \leq \alpha = \alpha_s + \alpha_{c,s}$. In addition, $|\mc{V}_1^b| = \alpha_s + |\mc{V}_2^g|$, $|\mc{V}_2| \leq \alpha_{c,s}$, and $|\mc{V}_2^b|>0$. Other subgroups contain the remaining $\alpha - (|\mc{V}_1^b| + |\mc{V}_2^b|)$ bait robots, denoted as $\{\mc{V}_{k}\}_{k=3}^K$. 
    
    Notably, in subgroups $\{\mc{V}_{k}\}_{k=3}^K$, since all the robots have bait control inputs, $\texttt{RATT}$ has the same approximation bound $(1-c_\Phi)^2$ whether there exist sensing attacks (see eqs.~\eqref{eq:nd_p1d2vk_1} - \eqref{eq:nd_p1d2vk_4}) or not (see eqs.~\eqref{eq:nd_p1d1vk_1} - \eqref{eq:nd_p1d1vk_5}). Thus, we focus on the distribution of the sensing attacks in subgroups $\mc{V}_1$ and $\mc{V}_2$. To this end, we consider three possible distributions of the $\alpha_s$ worst-case sensing attacks under Pattern 2. An example is shown in Fig.~\ref{fig:p2_3d}.

    \vspace{3pt} 
    \textbf{-- Distribution 1:} The $\alpha_s$ sensing attacks are in subgroup $\mc{V}_1$ (see Fig.~\ref{fig:p2_3d}-(a)). With Distribution 1, we evaluate \texttt{RATT}'s approximation bound in each subgroup $\mc{V}_k$ as follows. 
        
    \vspace{2pt}
    \textit{i):} In subgroup $\mc{V}_1$, one has $|\mc{A}_1^\star| = \alpha_s$ and $\mc{V}_1 = \mc{V}_1^b \cup \mc{V}_1^g$ with $|\mc{V}_1^b| = \alpha_s + |\mc{V}_2^g|$. The subset of $|\mc{V}_2^g|$ bait robots in $\mc{V}_1^b$ can be substituted for the greedy robots $\mc{V}_2^g$ that is separated into $\mc{V}_2$, since $\Phi(\mb{u}^b_{i}) \geq \Phi(\mb{u}^g_{i'})$ for any $i\in\mc{V}_1^b$ and $i'\in\mc{V}_2^g$. Then this scenario turns out to be Pattern 1, Distribution 1-a), and thus, one has 
        \begin{align}
        \Phi(\mb{u}_{\mc{V}_1 \setminus \mc{A}_1^\star}) 
        \geq (1-c_\Phi)^3 \Phi(\mb{u}^\star_{\mc{V}_1 \setminus \mc{A}_1^\star}). ~\label{eq:nd_p2d1v1}
    \end{align}
    
    \vspace{2pt}
    \textit{ii):}  In subgroup $\mc{V}_2$, one has $|\mc{V}_2| \leq \alpha_{c,s}$, $|\mc{V}_2^b| > 0$, and $\mc{A}_2^\star = \emptyset$. Denote $i_2^{\star} = \text{arg} \max_{i\in \mc{V}_2^b} \Phi(\mb{u}^b_i)$. Then one has:
    \begin{align}
        & \Phi(\mb{u}_{\mc{V}_2\setminus \mc{A}_2^\star}) = \Phi(\mb{u}_{\mc{V}_2}) ~\label{eq:nd_p2d1v2_1}\\
        & \geq (1-c_\Phi) \sum_{i \in \mc{V}_2} \Phi(\mb{u}_{i}), ~\label{eq:nd_p2d1v2_2}\\
        & = (1-c_\Phi) [\Phi(\mb{u}^b_{i_2^{\star}}) + \sum_{i \in \mc{V}_2\setminus i_2^{\star}} \Phi(\mb{u}_{i})], ~\label{eq:nd_p2d1v2_3}\\ 
        & \geq (1-c_\Phi) \frac{1}{|\mc{V}_2|} [\sum_{i \in \mc{V}_2^b} \Phi(\mb{u}^b_{i}) + \sum_{i \in \mc{V}_2^g} \Phi(\mb{u}^b_{i})], ~\label{eq:nd_p2d1v2_4}\\
        & = (1-c_\Phi) \frac{1}{|\mc{V}_2|} \sum_{i \in \mc{V}_2} \Phi(\mb{u}^b_{i}), ~\label{eq:nd_p2d1v2_5}\\
         & \geq (1-c_\Phi) \frac{1}{|\mc{V}_2|} \sum_{i \in \mc{V}_2} \Phi(\mb{u}^\star_{i}), ~\label{eq:nd_p2d1v2_6}\\
         & \geq  (1-c_\Phi)^2 \frac{1}{|\mc{V}_2|} \Phi(\mb{u}^\star_{\mc{V}_2}), ~\label{eq:nd_p2d1v2_7} \\
         & \geq  (1-c_\Phi)^2 \frac{1}{\alpha_{c,s}} \Phi(\mb{u}^\star_{\mc{V}_2}). ~\label{eq:nd_p2d1v2_8} \\
         & =  (1-c_\Phi)^2 \frac{1}{\alpha_{c,s}} \Phi(\mb{u}^\star_{\mc{V}_2\setminus \mc{A}_2^\star}). ~\label{eq:nd_p2d1v2_9}
    \end{align}
    Eqs.~\eqref{eq:nd_p2d1v2_1} - \eqref{eq:nd_p2d1v2_9} hold for the following reasons. Eq.~\eqref{eq:nd_p2d1v2_1} holds since $\mc{A}_2^\star = \emptyset$. 
    Ineq.~\eqref{eq:nd_p2d1v2_2} follows from eq.~\eqref{eq:nd_p2d1v2_1} due to \cite[Lemma 4]{tzoumas2018resilient}. Eq.~\eqref{eq:nd_p2d1v2_3} follows from ineq.~\eqref{eq:nd_p2d1v2_2} since $\mc{V}_2 = i_2^{\star}  \cup \mc{V}_2\setminus i_2^{\star}$ and $i_2^{\star} \in \mc{V}_2^b$.
    Ineq.~\eqref{eq:nd_p2d1v2_4} follows from eq.~\eqref{eq:nd_p2d1v2_3} because: first, by the definition of $i_2^{\star}$, one has $\Phi(\mb{u}^b_{i_2^{\star}}) \geq \Phi(\mb{u}^b_{i})$ for all $i\in\mc{V}_2^b$; second, since the control inputs of $\mc{V}_2^g$ are not selected as bait control inputs,  $\Phi(\mb{u}^b_{i}) \geq \Phi(\mb{u}^b_{i'})$ for any $i\in\mc{V}_2^b$ and $i'\in\mc{V}_2^g$, and thus $\Phi(\mb{u}^b_{i_2^{\star}}) \geq \Phi(\mb{u}^b_{i})$ for any $i\in\mc{V}_2^g$ as well; therefore $|\mc{V}_2|\Phi(\mb{u}^b_{i_2^{\star}}) \geq \sum_{i \in \mc{V}_2^b} \Phi(\mb{u}^b_{i}) + \sum_{i \in \mc{V}_2^g} \Phi(\mb{u}^b_{i})$. Eq.~\eqref{eq:nd_p2d1v2_5} follows from ineq.~\eqref{eq:nd_p2d1v2_4} since $\mc{V}_1 = \mc{V}_1^b + \mc{V}_1^g$.  Ineq.~\eqref{eq:nd_p2d1v2_6} follows from eq.~\eqref{eq:nd_p2d1v2_5} since the bait control input gives the maximum individual tracking quality. Ineq.~\eqref{eq:nd_p2d1v2_7} follows from  ineq.~\eqref{eq:nd_p2d1v2_6} due to \cite[Corollary 1]{tzoumas2018resilient}. Ineq.~\eqref{eq:nd_p2d1v2_8} follows from ineq.~\eqref{eq:nd_p2d1v2_7} due to $|\mc{V}_2| \leq \alpha_{c,s}$. Ineq.~\eqref{eq:nd_p2d1v2_9} follows from ineq.~\eqref{eq:nd_p2d1v2_8} due to $\mc{A}_2^\star = \emptyset$.

    \vspace{2pt}
    To sum up, in Pattern 2, Distribution 1, by eq.~\eqref{eq:nd_p2d1v1} and eqs.~\eqref{eq:nd_p2d1v2_1}-\eqref{eq:nd_p2d1v2_9}, one has
    \begin{align*}
            &\max_{k\in \{1,\cdots, K\}}\Phi(\mb{u}_{\mc{V}_k\setminus \mc{A}_k^\star}) \\
            &\geq \min[(1-c_\Phi)^3, \frac{(1-c_\Phi)^2}{\alpha_{c,s}}] \max_{k\in \{1,\cdots, K\}}\Phi(\mb{u}^\star_{\mc{V}_k\setminus \mc{A}_k^\star}).
    \end{align*}
    Therefore, along with eqs.~\eqref{eq:team_subgroup} and \eqref{eq:opt_team_subgroup}, one has
    \begin{align}\label{eq:nd_p2_d1}
        &\Phi(\mc{V}\setminus\emptyset, \mc{E}\setminus\mc{A}_{c}^\star) \geq 
   \min[(1-c_\Phi)^3, \frac{(1-c_\Phi)^2}{\alpha_{c,s}}] \Phi_{\mc{V}, \alpha_s, \mc{E},\alpha_c}^\star. 
    \end{align}
    
    \vspace{3pt} 
    \textbf{-- Distribution 2:} The $\alpha_s$ sensing attacks are in subgroup $\mc{V}_2$ (see Fig.~\ref{fig:p2_3d}-(b)). With Distribution 2, we evaluate \texttt{RATT}'s approximation bound in each subgroup $\mc{V}_k$ as follows. 
    
    \vspace{2pt}
    \textit{i):} In subgroup $\mc{V}_1$, one has $\mc{V}_1 = \mc{V}_1^b \cup \mc{V}_1^g$ with $|\mc{V}_1^b| = \alpha_s + |\mc{V}_2^g|$. The subset of $|\mc{V}_2^g|$ bait robots in $\mc{V}_1^b$ can be substituted for the greedy robots $\mc{V}_2^g$ that is separated into $\mc{V}_2$, since $\Phi(\mb{u}^b_{i}) \geq \Phi(\mb{u}^g_{i'})$ for any $i\in\mc{V}_1^b$ and $i'\in\mc{V}_2^g$. In addition, there exist no attacks ($\mc{A}_1^\star = \emptyset$). Then this scenario turns out to be Pattern 1, Distribution 2-a), and thus, one has 
        \begin{align}
        \Phi(\mb{u}_{\mc{V}_1 \setminus \mc{A}_1^\star}) 
        \geq (1-c_\Phi)^3 \Phi(\mb{u}^\star_{\mc{V}_1 \setminus \mc{A}_1^\star}). ~\label{eq:nd_p2d2v1}
    \end{align}
    
    \vspace{2pt}
    \textit{ii):}  In subgroup $\mc{V}_2$, one has $|\mc{V}_2| \leq \alpha_{c,s}$, $|\mc{V}_2^b| > 0$, and $|\mc{A}_2^\star| = \alpha_s$. If $\mc{V}_2 \setminus \mc{A}_2^\star$ contains greedy robots only, then $\mc{V}_2 \setminus \mc{A}_2^\star \subseteq \mc{V}_2^g$ and it is \textit{dominated} by $\mc{V}_1\setminus \mc{A}_1^\star$, \textit{i.e.}, $\Phi(\mb{u}_{\mc{V}_1\setminus \mc{A}_1^\star}) \geq \Phi(\mb{u}_{\mc{V}_2\setminus \mc{A}_2^\star})$. That is because, $\mc{A}_1^\star = \emptyset$, $|\mc{V}_1^b| > |\mc{V}_2^g| \geq \mc{V}_2 \setminus \mc{A}_2^\star$, and $\Phi(\mb{u}^b_{i}) \geq \Phi(\mb{u}^g_{i'})$ for any $i\in\mc{V}_1^b$ and $i'\in\mc{V}_2^g$. 
    
   If $\mc{V}_2 \setminus \mc{A}_2^\star$ contains both bait and greedy robots, \texttt{RATT} has the same approximation bound as in Pattern 2, Distribution 1-b) by following eqs.~\eqref{eq:nd_p2d1v2_1} - \eqref{eq:nd_p2d1v2_9}.  Thus, in total, one has 
    \begin{align}
         \Phi(\mb{u}_{\mc{V}_2\setminus \mc{A}_2^\star}) \geq (1-c_\Phi)^2 \frac{1}{\alpha_{c,s}} \Phi(\mb{u}^\star_{\mc{V}_2\setminus \mc{A}_2^\star}). ~\label{eq:nd_p2d2v2}
    \end{align}

    \vspace{2pt}
    To sum up, in Pattern 2, Distribution 2, by eq.~\eqref{eq:nd_p2d2v1} and eq.~\eqref{eq:nd_p2d2v2}, one has
    \begin{align*}
            &\max_{k\in \{1,\cdots, K\}}\Phi(\mb{u}_{\mc{V}_k\setminus \mc{A}_k^\star}) \\
            &\geq \min[(1-c_\Phi)^3, \frac{(1-c_\Phi)^2}{\alpha_{c,s}}] \max_{k\in \{1,\cdots, K\}}\Phi(\mb{u}^\star_{\mc{V}_k\setminus \mc{A}_k^\star}).
    \end{align*}
    Therefore, along with eqs.~\eqref{eq:team_subgroup} and \eqref{eq:opt_team_subgroup}, one has
    \begin{align}\label{eq:nd_p2_d2}
        &\Phi(\mc{V}\setminus\emptyset, \mc{E}\setminus\mc{A}_{c}^\star) \geq 
   \min[(1-c_\Phi)^3, \frac{(1-c_\Phi)^2}{\alpha_{c,s}}] \Phi_{\mc{V}, \alpha_s, \mc{E},\alpha_c}^\star. 
    \end{align}

    \vspace{3pt} 
    \textbf{-- Distribution 3:} Part of the $\alpha_s$ sensing attacks (\textit{i.e.}, $\alpha_s^1$, $\alpha_s^1 < \alpha_s$) is in subgroup $\mc{V}_1$ and the remaining part (\textit{i.e.}, $\alpha_s - \alpha_s^1$) is in subgroups $\{\mc{V}_{k}\}_{k=2}^{K}$ (see Fig.~\ref{fig:p2_3d}-(c)). 
 
    \vspace{2pt}
    \textit{i):} In subgroup $\mc{V}_1$, one has $|\mc{A}_1^\star| = \alpha_s^1$ and $\mc{V}_1 = \mc{V}_1^b \cup \mc{V}_1^g$ with $|\mc{V}_1^b| = \alpha_s + |\mc{V}_2^g|$. The subset of $|\mc{V}_2^g|$ bait robots in $\mc{V}_1^b$ can be substituted for the greedy robots $\mc{V}_2^g$ that is separated into $\mc{V}_2$, since $\Phi(\mb{u}^b_{i}) \geq \Phi(\mb{u}^g_{i'})$ for any $i\in\mc{V}_1^b$ and $i'\in\mc{V}_2^g$. Then this scenario turns out to be Pattern 1, Distribution 3-a), and thus, one has 
        \begin{align}
        \Phi(\mb{u}_{\mc{V}_1 \setminus \mc{A}_1^\star}) 
        \geq (1-c_\Phi)^3 \Phi(\mb{u}^\star_{\mc{V}_1 \setminus \mc{A}_1^\star}). ~\label{eq:nd_p2d3v1}
    \end{align}
    
    \vspace{2pt}
    \textit{ii):}  In subgroup $\mc{V}_2$, one has $|\mc{V}_2| \leq \alpha_{c,s}$, $|\mc{V}_2^b| > 0$, and $|\mc{A}_2^\star| = \alpha_s - \alpha_s^1$. If $\mc{V}_2 \setminus \mc{A}_2^\star$ contains greedy robots only, then $\mc{V}_2 \setminus \mc{A}_2^\star \subseteq \mc{V}_2^g$ and it is \textit{dominated} by $\mc{V}_1\setminus \mc{A}_1^\star$, \textit{i.e.}, $\Phi(\mb{u}_{\mc{V}_1\setminus \mc{A}_1^\star}) \geq \Phi(\mb{u}_{\mc{V}_2\setminus \mc{A}_2^\star})$.  That is because, $|\mc{V}_1^b| = \alpha_s + |\mc{V}_2^g|$ and $\Phi(\mb{u}^b_{i}) \geq \Phi(\mb{u}^g_{i'})$ for any $i\in\mc{V}_1^b$ and $i'\in\mc{V}_2^g$. Hence, even though all $\alpha_s^1$ sensing attacks happen in $\mc{V}_1^b$, $\mc{V}_1$ still have $\alpha - \alpha_s^1 +|\mc{V}_2^g|$ bait robots left, that dominate $\mc{V}_2 \setminus \mc{A}_2^\star$. 
   
    If $\mc{V}_2 \setminus \mc{A}_2^\star$ contains both bait and greedy robots, \texttt{RATT} has the same approximation bound as in Pattern 2, Distribution 1-b) by following eqs.~\eqref{eq:nd_p2d1v2_1} - \eqref{eq:nd_p2d1v2_9}.  Thus, in total, one has 
    \begin{align}
         \Phi(\mb{u}_{\mc{V}_2\setminus \mc{A}_2^\star}) \geq (1-c_\Phi)^2 \frac{1}{\alpha_{c,s}} \Phi(\mb{u}^\star_{\mc{V}_2\setminus \mc{A}_2^\star}). ~\label{eq:nd_p2d3v2}
    \end{align}
    
    \vspace{2pt}
    To sum up, in Pattern 2, Distribution 3, by eq.~\eqref{eq:nd_p2d3v1}, eq.~\eqref{eq:nd_p2d3v2}, one has
    \begin{align*}
            &\max_{k\in \{1,\cdots, K\}}\Phi(\mb{u}_{\mc{V}_k\setminus \mc{A}_k^\star}) \\
            &\geq \min[(1-c_\Phi)^3, \frac{(1-c_\Phi)^2}{\alpha_{c,s}}] \max_{k\in \{1,\cdots, K\}}\Phi(\mb{u}^\star_{\mc{V}_k\setminus \mc{A}_k^\star}).
    \end{align*}
    Therefore, along with eqs.~\eqref{eq:team_subgroup} and \eqref{eq:opt_team_subgroup}, one has
    \begin{align}\label{eq:nd_p2_d3}
        &\Phi(\mc{V}\setminus\emptyset, \mc{E}\setminus\mc{A}_{c}^\star) \geq 
   \min[(1-c_\Phi)^3, \frac{(1-c_\Phi)^2}{\alpha_{c,s}}] \Phi_{\mc{V}, \alpha_s, \mc{E},\alpha_c}^\star. 
    \end{align}
    
    \vspace{3pt}
    Overall, in Pattern 2, by eqs.~\eqref{eq:nd_p2_d1}, \eqref{eq:nd_p2_d2}, and \eqref{eq:nd_p2_d3}, one has
    \begin{equation}\label{eq:nd_p2}
         \Phi(\mc{V}\setminus\emptyset, \mc{E}\setminus\mc{A}_{c}^\star) \geq 
   \min[(1-c_\Phi)^3, \frac{(1-c_\Phi)^2}{\alpha_{c,s}}] \Phi_{\mc{V}, \alpha_s, \mc{E},\alpha_c}^\star.
    \end{equation}
\end{itemize}

Due to the limited space, we omit the discussions of other patterns and the corresponding possible distributions, since \texttt{RATT} either gives the same approximation bound or a better approximation bound in those scenarios. Hence, to summarize \texttt{RATT}'s approximation bounds in Pattern 1 and Pattern 2, we have
\begin{equation}\label{eq:nd1}
         \Phi(\mc{V}\setminus\emptyset, \mc{E}\setminus\mc{A}_{c}^\star) \geq 
   \min[(1-c_\Phi)^3, \frac{(1-c_\Phi)^2}{\alpha_{c,s}}] \Phi_{\mc{V}, \alpha_s, \mc{E},\alpha_c}^\star.
 \end{equation}

We then consider $N \leq \alpha$. In this instance, all $N$ robots are bait robots (by \texttt{RATT}). The remaining robots after the worst-case communication and sensing attacks are still bait robots. This scenario is the same as Pattern 1, Distribution 2-b), and thus for all subgroups $\{\mc{V}_k\}_{k=1}^{K}$, one has
    \begin{align}
        \Phi(\mb{u}_{\mc{V}_k \setminus \mc{A}_k^\star}) \geq (1-c_\Phi)^2 \Phi(\mb{u}^\star_{\mc{V}_k\setminus \mc{A}_k^\star}). ~\label{eq:nd2}
    \end{align}

Thus, all in all, by eq.~\eqref{eq:nd1} and eq.~\eqref{eq:nd2}, we have
\begin{equation}\label{eq:nd}
         \Phi(\mc{V}\setminus\emptyset, \mc{E}\setminus\mc{A}_{c}^\star) \geq 
   \min[(1-c_\Phi)^3, \frac{(1-c_\Phi)^2}{\alpha_{c,s}}] \Phi_{\mc{V}, \alpha_s, \mc{E},\alpha_c}^\star.
 \end{equation}

Till now, eq.~\eqref{eqn:appro_non_descrease} is proved. 
  
\vspace{5pt}
\subsubsection{Proof of \texorpdfstring{eq.~\eqref{eqn:appro_submodular}}{}} 
We first consider $N > \alpha$. In this instance, $|\mc{V}^g| = N- \alpha > 0$, and thus, there exist some greedy robots. Then, the proof of eq.~\eqref{eqn:appro_submodular} can directly follow the same steps for proving eq.~\eqref{eqn:appro_non_descrease}, \textit{i.e.}, investigating two patterns and the corresponding distributions. Due to the limited space, we focus on Pattern 2, Distribution 1 that gives the worst-case approximation bound. We omit other patterns and cases since \texttt{RATT} either gives the same approximation bound or a better approximation bound in those scenarios. To distinguish from the proof of eq.~\eqref{eqn:appro_non_descrease}, we name Pattern 2, Distribution 1 as Pattern $2'$, Distribution $1'$ for proving eq.~\eqref{eqn:appro_submodular}. 

\begin{itemize}
    \item \textbf{Pattern $2'$:} Two subgroups, denoted by $\mc{V}_1$ and $\mc{V}_2$, contain all $N-\alpha$ greedy robots and some bait robots. That is, $|\mc{V}_1^g| + |\mc{V}_2^g| = N-\alpha$ and $|\mc{V}_1^b| + |\mc{V}_2^b| \leq \alpha = \alpha_s + \alpha_{c,s}$. In addition, $|\mc{V}_1^b| = \alpha_s + |\mc{V}_2^g|$, $|\mc{V}_2| \leq \alpha_{c,s}$, and $|\mc{V}_2^b|>0$. Other subgroups contain the remaining $\alpha - (|\mc{V}_1^b| + |\mc{V}_2^b|)$ bait robots, denoted as $\{\mc{V}_{k}\}_{k=3}^K$. 
    
    Notably, in subgroups $\{\mc{V}_{k}\}_{k=3}^K$, all the robots are bait control robots. If there exist sensing attacks, the remaining robots are still bait robots. Hence, $\texttt{RATT}$ has the same approximation bound whether there exist sensing attacks or not. That is, 
    \begin{align}
        \Phi(\mb{u}_{\mc{V}_k \setminus \mc{A}_k^\star}) &= \Phi(\mb{u}^b_{\mc{V}_k\setminus \mc{A}_k^\star}) ~\label{eq:sub_p2vk_1}\\
        & \geq (1-k_\Phi) \sum_{i \in \mc{V}_k\setminus \mc{A}_k^\star} \Phi(\mb{u}^b_{i}), ~\label{eq:sub_p2vk_2}\\
        & = (1-k_\Phi) \sum_{i \in \mc{V}_k\setminus \mc{A}_k^\star} \Phi(\mb{u}_{i}^\star), ~\label{eq:sub_p2vk_3} \\
        & \geq (1-k_\Phi) \Phi(\mb{u}^\star_{\mc{V}_k\setminus \mc{A}_k^\star}). ~\label{eq:sub_p2vk_4}
    \end{align}
    Eqs.~\eqref{eq:sub_p2vk_1} - \eqref{eq:sub_p2vk_4} hold for the following reasons. Eq.~\eqref{eq:sub_p2vk_1} holds since $\mc{V}_k = \mc{V}_k^b$. 
    Ineq.~\eqref{eq:sub_p2vk_2} follows from eq.~\eqref{eq:sub_p2vk_1} due to \cite[Lemma 2]{tzoumas2018resilient}, which is derived based on the definition of the curvature (eq.~\eqref{eq:curvature}) and the submodularity of $\Phi$. Eq.~\eqref{eq:sub_p2vk_3} follows from ineq.~\eqref{eq:sub_p2vk_2} since each bait control input $\mb{u}_{i}^b$ gives the maximum individual tracking quality. Ineq.~\eqref{eq:sub_p2vk_4} follows from eq.~\eqref{eq:sub_p2vk_3} due to the submodularity of $\Phi$.
    
    Hence, we focus on the distribution of the sensing attacks in subgroups $\mc{V}_1$ and $\mc{V}_2$. To this end, we consider three possible distributions of the $\alpha_s$ worst-case sensing attacks under Pattern 2. An example is shown in Fig.~\ref{fig:p2_3d}.
    
    \vspace{3pt} 
    \textbf{-- Distribution $1'$:} The $\alpha_s$ sensing attacks are in subgroup $\mc{V}_1$ (see Fig.~\ref{fig:p2_3d}-(a)). With Distribution 1, we evaluate \texttt{RATT}'s approximation bound in each subgroup $\mc{V}_k$ as follows. 
        
    \vspace{2pt}
    \textit{i):} In subgroup $\mc{V}_1$, one has $|\mc{A}_1^\star| = \alpha_s$ and $\mc{V}_1 = \mc{V}_1^b \cup \mc{V}_1^g$ with $|\mc{V}_1^b| = \alpha_s + |\mc{V}_2^g|$. Given the total number of sensing attacks is $\alpha_s$, the number of sensing attacks $\alpha_{s}$ in $\mc{V}_1^b$ is less than (or equal to) $\alpha_s$. Then there must exist a subset of $|\mc{V}_1^b| - \alpha_s = |\mc{V}_2^g|$ robots that does not encounter sensing attacks in $\mc{V}_1^b$. We denote this subset as $\mc{V}_1^{b1}\subseteq \mc{V}_1^{b}$ with $|\mc{V}_1^{b1}| = \alpha-\alpha_s^1$. In addition, denote the other robots in $\mc{V}_1^{b1}$ as $\mc{V}_1^{b2}= \mc{V}_1^{b}\setminus \mc{V}_1^{b1}$. Notably, $|\mc{V}_1^{b2}| = \alpha_s$. Then, we evaluate \texttt{RATT}'s approximation bound, $\Phi(\mb{u}_{\mc{V}_1 \setminus \mc{A}_1^\star})/\Phi(\mb{u}^\star_{\mc{V}_1 \setminus \mc{A}_1^\star})$, in  $\mc{V}_1$ as follows.
    
    \begin{align}
        & \Phi(\mb{u}_{\mc{V}_1 \setminus \mc{A}_1^\star}) = \Phi(\mb{u}_{(\mc{V}_1^b \setminus \mc{A}_1^\star) \cup (\mc{V}_1^g \setminus \mc{A}_1^\star)}), ~\label{eq:sub_p2d1v1_1}\\
        & \geq (1-k_\Phi) \sum_{i \in (\mc{V}_1^b \setminus \mc{A}_1^\star) \cup (\mc{V}_1^g \setminus \mc{A}_1^\star)} \Phi(\mb{u}_{i}), ~\label{eq:sub_p2d1v1_2}\\
        & = (1-k_\Phi) [\sum_{i \in \mc{V}_1^{b2} \setminus \mc{A}_1^\star} \Phi(\mb{u}^b_{i}) \nonumber
        \\ & \hspace{2cm} + \sum_{i \in \mc{V}_1^{b1}} \Phi(\mb{u}^b_{i})   + \sum_{i \in \mc{V}_1^g \setminus \mc{A}_1^\star} \Phi(\mb{u}^g_{i})], ~\label{eq:sub_p2d1v1_3} \\
        & \geq (1-k_\Phi) [\sum_{i \in \mc{V}_1^{b2} \setminus \mc{A}_1^\star} \Phi(\mb{u}^b_{i}) \nonumber\\ & \hspace{2cm} + \sum_{i \in \mc{V}_2^{g}} \Phi(\mb{u}^g_{i})   + \sum_{i \in \mc{V}_1^g \setminus \mc{A}_1^\star} \Phi(\mb{u}^g_{i})], ~\label{eq:sub_p2d1v1_4}\\
        & = (1-k_\Phi) [\sum_{i \in \mc{V}_1^{b2} \setminus \mc{A}_1^\star} \Phi(\mb{u}^b_{i}) + \sum_{i \in \mc{V}^g \setminus \mc{A}_1^\star} \Phi(\mb{u}^g_{i})], ~\label{eq:sub_p2d1v1_5} \\
        & \geq (1-k_\Phi) [\sum_{i \in \mc{V}^g\setminus (\mc{V}^g \setminus \mc{A}_1^\star) } \Phi(\mb{u}^g_{i}) + \sum_{i \in \mc{V}^g \setminus \mc{A}_1^\star} \Phi(\mb{u}^g_{i})], ~\label{eq:sub_p2d1v1_6}
        \end{align}
        \begin{align}
        & = (1-k_\Phi) \sum_{i \in \mc{V}^g} \Phi(\mb{u}^g_{i}), ~\label{eq:sub_p2d1v1_7}\\
        & \geq (1-k_\Phi) \Phi(\mb{u}^g_{\mc{V}^g}), ~\label{eq:sub_p2d1v1_8} \\
        & \geq \frac{1-k_\Phi}{1+k_\Phi}  \Phi(\mb{u}^\star_{\mc{V}^g}),
        ~\forall \mc{W}\subseteq \mc{V}_1, |\mc{W}| = |\mc{V}^g|, ~\label{eq:sub_p2d1v1_9} \\
        & \geq \frac{1-k_\Phi}{1+k_\Phi} \Phi(\mb{u}^\star_{\mc{V}_1 \setminus \mc{A}_1^\star}). ~\label{eq:sub_p2d1v1_10}
    \end{align}

    Eqs.~\eqref{eq:sub_p2d1v1_1} - \eqref{eq:sub_p2d1v1_10} hold for the following reasons. Eq.~\eqref{eq:nd_p1d3v1_1} holds since $\mc{V}_1 = \mc{V}_1^b + \mc{V}_1^g$. Ineq.~\eqref{eq:sub_p2d1v1_2} follows from eq.~\eqref{eq:sub_p2d1v1_1} due to \cite[Lemma 2]{tzoumas2018resilient}.
    Eq.~\eqref{eq:sub_p2d1v1_3} follows from ineq.~\eqref{eq:sub_p2d1v1_2} due to the definitions of $\mc{V}_1^{b1}$ and $\mc{V}_1^{b2}$. Ineq.~\eqref{eq:sub_p2d1v1_4} follows from eq.~\eqref{eq:sub_p2d1v1_3} since $\Phi(\mb{u}^{b1}_{i}) \geq \Phi(\mb{u}^g_{i'})$ for any $i\in\mc{V}_1^b$ and $i'\in\mc{V}_2^g$. Eq.~\eqref{eq:sub_p2d1v1_5} follows from ineq.~\eqref{eq:sub_p2d1v1_4} due to $\mc{V}^g = \mc{V}_1^g \cup \mc{V}_2^g$. Ineq.~\eqref{eq:sub_p2d1v1_6} follows from eq.~\eqref{eq:sub_p2d1v1_5} because: first, since $\mc{V}_1^{b2}$ are bait robots and $\mc{V}^g$ are greedy robots,  $\Phi(\mb{u}^b_{i}) \geq \Phi(\mb{u}^g_{i'})$ for any $i\in\mc{V}_1^{b2}\setminus \mc{A}_1^\star$ and $i'\in\mc{V}^g \setminus \mc{A}_1^\star$; second, $|\mc{V}_1^{b2} \cap \mc{A}_1^\star| + |\mc{V}^g \cap \mc{A}_1^\star| = \alpha_s$, and thus $|\mc{V}_1^{b2}\setminus \mc{A}_1^\star| = \alpha_s - |\mc{V}_1^{b2} \cap \mc{A}_1^\star| = |\mc{V}^g \cap \mc{A}_1^\star| = |\mc{V}^g\setminus (\mc{V}^g \setminus \mc{A}_1^\star)|$. Eq.~\eqref{eq:sub_p2d1v1_7} follows from ineq.~\eqref{eq:sub_p2d1v1_6} naturally. Ineq.~\eqref{eq:sub_p2d1v1_8} follows from eq.~\eqref{eq:sub_p2d1v1_7} due to the submodularity of $\Phi$. Ineq.~\eqref{eq:sub_p2d1v1_9} follows from eq.~\eqref{eq:sub_p2d1v1_8} due to \cite[Theorem 2.3]{conforti1984submodular}, which implies that the greedy algorithm, introduced in \cite[Section 2]{fisher1978analysis}, achieves $1-k_\Phi$ approximation bound for optimizing non-decreasing and submodular functions.
    Ineq.~\eqref{eq:sub_p2d1v1_10} follows from ineq.~\eqref{eq:sub_p2d1v1_9} due to \cite[Lemma 9]{tzoumas2018resilient}. 
    
    \vspace{2pt}
    \textit{ii):}  In subgroup $\mc{V}_2$, one has $|\mc{V}_2| \leq \alpha_{c,s}$, $|\mc{V}_2^b| > 0$, and $\mc{A}_2^\star = \emptyset$. Denote $i_2^{\star} = \text{arg} \max_{i\in \mc{V}_2^b} \Phi(\mb{u}^b_i)$. Then one has:
    \begin{align}
        & \Phi(\mb{u}_{\mc{V}_2\setminus \mc{A}_2^\star}) = \Phi(\mb{u}_{\mc{V}_2}) ~\label{eq:sub_p2d1v2_1}\\
        & \geq (1-k_\Phi) \sum_{i \in \mc{V}_2} \Phi(\mb{u}_{i}), ~\label{eq:sub_p2d1v2_2}\\
        & = (1-k_\Phi) [\Phi(\mb{u}^b_{i_2^{\star}}) + \sum_{i \in \mc{V}_2\setminus i_2^{\star}} \Phi(\mb{u}_{i})], ~\label{eq:sub_p2d1v2_3}\\ 
        & \geq (1-k_\Phi) \frac{1}{|\mc{V}_2|} [\sum_{i \in \mc{V}_2^b} \Phi(\mb{u}^b_{i}) + \sum_{i \in \mc{V}_2^g} \Phi(\mb{u}^b_{i})], ~\label{eq:sub_p2d1v2_4}\\
        & = (1-k_\Phi) \frac{1}{|\mc{V}_2|} \sum_{i \in \mc{V}_2} \Phi(\mb{u}^b_{i}), ~\label{eq:sub_p2d1v2_5}\\
         & \geq (1-k_\Phi) \frac{1}{|\mc{V}_2|} \sum_{i \in \mc{V}_2} \Phi(\mb{u}^\star_{i}), ~\label{eq:sub_p2d1v2_6}\\
         & \geq  (1-k_\Phi) \frac{1}{|\mc{V}_2|} \Phi(\mb{u}^\star_{\mc{V}_2}), ~\label{eq:sub_p2d1v2_7} \\
         & \geq  (1-k_\Phi) \frac{1}{\alpha_{c,s}} \Phi(\mb{u}^\star_{\mc{V}_2}). ~\label{eq:sub_p2d1v2_8} \\
         & =  (1-k_\Phi) \frac{1}{\alpha_{c,s}} \Phi(\mb{u}^\star_{\mc{V}_2\setminus \mc{A}_2^\star}). ~\label{eq:sub_p2d1v2_9}
    \end{align}
    Eqs.~\eqref{eq:sub_p2d1v2_1} - \eqref{eq:sub_p2d1v2_9} hold for the following reasons. Eq.~\eqref{eq:sub_p2d1v2_1} holds since $\mc{A}_2^\star = \emptyset$. 
    Ineq.~\eqref{eq:sub_p2d1v2_2} follows from eq.~\eqref{eq:sub_p2d1v2_1} due to \cite[Lemma 2]{tzoumas2018resilient}. Eq.~\eqref{eq:sub_p2d1v2_3} follows from ineq.~\eqref{eq:sub_p2d1v2_2} since $\mc{V}_2 = i_2^{\star}  \cup \mc{V}_2\setminus i_2^{\star}$ and $i_2^{\star} \in \mc{V}_2^b$.
    Ineq.~\eqref{eq:sub_p2d1v2_4} follows from eq.~\eqref{eq:sub_p2d1v2_3} because: first, by the definition of $i_2^{\star}$, one has $\Phi(\mb{u}^b_{i_2^{\star}}) \geq \Phi(\mb{u}^b_{i})$ for all $i\in\mc{V}_2^b$; second, since the control inputs of $\mc{V}_2^g$ are not selected as bait control inputs,  $\Phi(\mb{u}^b_{i}) \geq \Phi(\mb{u}^b_{i'})$ for any $i\in\mc{V}_2^b$ and $i'\in\mc{V}_2^g$, and thus $\Phi(\mb{u}^b_{i_2^{\star}}) \geq \Phi(\mb{u}^b_{i})$ for any $i\in\mc{V}_2^g$ as well; therefore $|\mc{V}_2|\Phi(\mb{u}^b_{i_2^{\star}}) \geq \sum_{i \in \mc{V}_2^b} \Phi(\mb{u}^b_{i}) + \sum_{i \in \mc{V}_2^g} \Phi(\mb{u}^b_{i})$. Eq.~\eqref{eq:sub_p2d1v2_5} follows from ineq.~\eqref{eq:sub_p2d1v2_4} since $\mc{V}_1 = \mc{V}_1^b + \mc{V}_1^g$.  Ineq.~\eqref{eq:sub_p2d1v2_6} follows from eq.~\eqref{eq:sub_p2d1v2_5} since the bait control input gives the maximum individual tracking quality. Ineq.~\eqref{eq:sub_p2d1v2_7} follows from  ineq.~\eqref{eq:sub_p2d1v2_6} due to the submodularity of $\Phi$. Ineq.~\eqref{eq:sub_p2d1v2_8} follows from ineq.~\eqref{eq:sub_p2d1v2_7} due to $|\mc{V}_2| \leq \alpha_{c,s}$. Ineq.~\eqref{eq:sub_p2d1v2_9} follows from ineq.~\eqref{eq:sub_p2d1v2_8} due to $\mc{A}_2^\star = \emptyset$.
        
    \vspace{2pt}
    To sum up, by eqs.~\eqref{eq:sub_p2vk_1} - \eqref{eq:sub_p2vk_4}, and eqs.~\eqref{eq:sub_p2d1v1_1}- \eqref{eq:sub_p2d1v1_10} and eqs.~\eqref{eq:sub_p2d1v2_1}-\eqref{eq:sub_p2d1v2_9} in Pattern $2'$, Distribution $1'$, one has
    \begin{align*}
            &\max_{k\in \{1,\cdots, K\}}\Phi(\mb{u}_{\mc{V}_k\setminus \mc{A}_k^\star}) \\
            &\geq \min[1-k_\Phi, \frac{1-k_\Phi}{1+k_\Phi}, \frac{1-k_\Phi}{\alpha_{c,s}}] \max_{k\in \{1,\cdots, K\}}\Phi(\mb{u}^\star_{\mc{V}_k\setminus \mc{A}_k^\star}).
    \end{align*}
    Therefore, along with eqs.~\eqref{eq:team_subgroup} and \eqref{eq:opt_team_subgroup}, one has
    \begin{align}\label{eq:sub_p2_d1}
        \Phi(\mc{V}\setminus\emptyset, \mc{E}\setminus\mc{A}_{c}^\star) \geq 
   \min[\frac{1-k_\Phi}{1+k_\Phi}, \frac{1-k_\Phi}{\alpha_{c,s}}] \Phi_{\mc{V}, \alpha_s, \mc{E},\alpha_c}^\star. 
    \end{align}
    
\end{itemize}    

We then consider $N \leq \alpha$. In this instance, all $N$ robots are bait robots (by \texttt{RATT}). The remaining robots after the worst-case communication and sensing attacks are still bait robots. Then we directly follow eqs.~\eqref{eq:sub_p2vk_1} - \eqref{eq:sub_p2vk_4} to obtain \texttt{RATT}'s approximation bound. That is, for all subgroups $\{\mc{V}_k\}_{k=1}^{K}$, one has
    \begin{align}
        \Phi(\mb{u}_{\mc{V}_k \setminus \mc{A}_k^\star}) \geq (1-k_\Phi) \Phi(\mb{u}^\star_{\mc{V}_k\setminus \mc{A}_k^\star}). ~\label{eq:sub2}
    \end{align}

Thus, all in all, by eq.~\eqref{eq:sub_p2_d1} and eq.~\eqref{eq:sub2}, we have
\begin{equation}\label{eq:sub}
    \Phi(\mc{V}\setminus\emptyset, \mc{E}\setminus\mc{A}_{c}^\star) \geq 
   \min[\frac{1-k_\Phi}{1+k_\Phi}, \frac{1-k_\Phi}{\alpha_{c,s}}] \Phi_{\mc{V}, \alpha_s, \mc{E},\alpha_c}^\star.
 \end{equation}
 
Till now, eq.~\eqref{eqn:appro_submodular} is proved.

\subsection{Proof of Theorem~\ref{thm:ratt_run_time}}
\texttt{RATT}'s running time is equal to the total running time of \texttt{CAA} (Algorithm~\ref{alg:comm_appro}) and the robust maximization (\texttt{RATT}'s lines~\ref{line:ratt_compute_totalatk}-\ref{line:ratt_forend_allbaits}). First, \texttt{CAA} determines $n_{\max}$ by evaluating all possible values from $[1, 2, \cdots, N]$ (Algorithm~\ref{alg:comm_appro}, lines~\ref{line:caa_forstart}-\ref{line:caa_endfor}), and thus, it takes linear time, \textit{i.e.}, $O(\mc{V})$ time. Second, in the robust maximization, the selection of bait control inputs takes $O(\sum_{i=1}^N |\mc{U}_{i}|)= O(|\mc{U}_{\mc{V}}|)$ time by evaluating all robots' candidate control inputs (Algorithm~\ref{alg:rob_tar_track}, lines~\ref{line:ratt_forst_idvmax}-\ref{line:ratt_idv_robot_max}) and takes $O(|\mc{V}|\log(|\mc{V}|))$ time to pick out the top $\alpha$ control inputs (Algorithm~\ref{alg:rob_tar_track}, line~\ref{line:ratt_baits}). Thus, the bait selection takes $O(|\mc{U}_{\mc{V}}|)+O(|\mc{V}|\log(|\mc{V}|))$ time. In addition, the selection of greedy control inputs (Algorithm~\ref{alg:rob_tar_track}, lines~\ref{line:ratt_gre_emptyset}-\ref{line:ratt_gre_update}) uses the standard greedy algorithm~\cite{fisher1978analysis}, and thus takes $O(|\mc{U}_{\mc{V}}|^2)$ time. Hence, the robust maximization takes $O(|\mc{U}_{\mc{V}}|)+ O(|\mc{V}|\log(|\mc{V}|)) + O(|\mc{U}_{\mc{V}}|^2)$ time. Since each robot has at least one control input, one has $|\mc{U}_{\mc{V}}| \geq |\mc{V}|$, and thus $O(|\mc{U}_{\mc{V}}|)+ O(|\mc{V}|\log(|\mc{V}|)) + O(|\mc{U}_{\mc{V}}|^2) = O(|\mc{U}_{\mc{V}}|^2)$. Therefore, in total, \texttt{RATT} takes $O(\mc{V}) + O(|\mc{U}_{\mc{V}}|^2) = O(|\mc{U}_{\mc{V}}|^2)$ time.

\bibliographystyle{IEEEtran}
\bibliography{refs}

\begin{IEEEbiography}[{\includegraphics[width=1in,height=1.25in,clip,keepaspectratio]{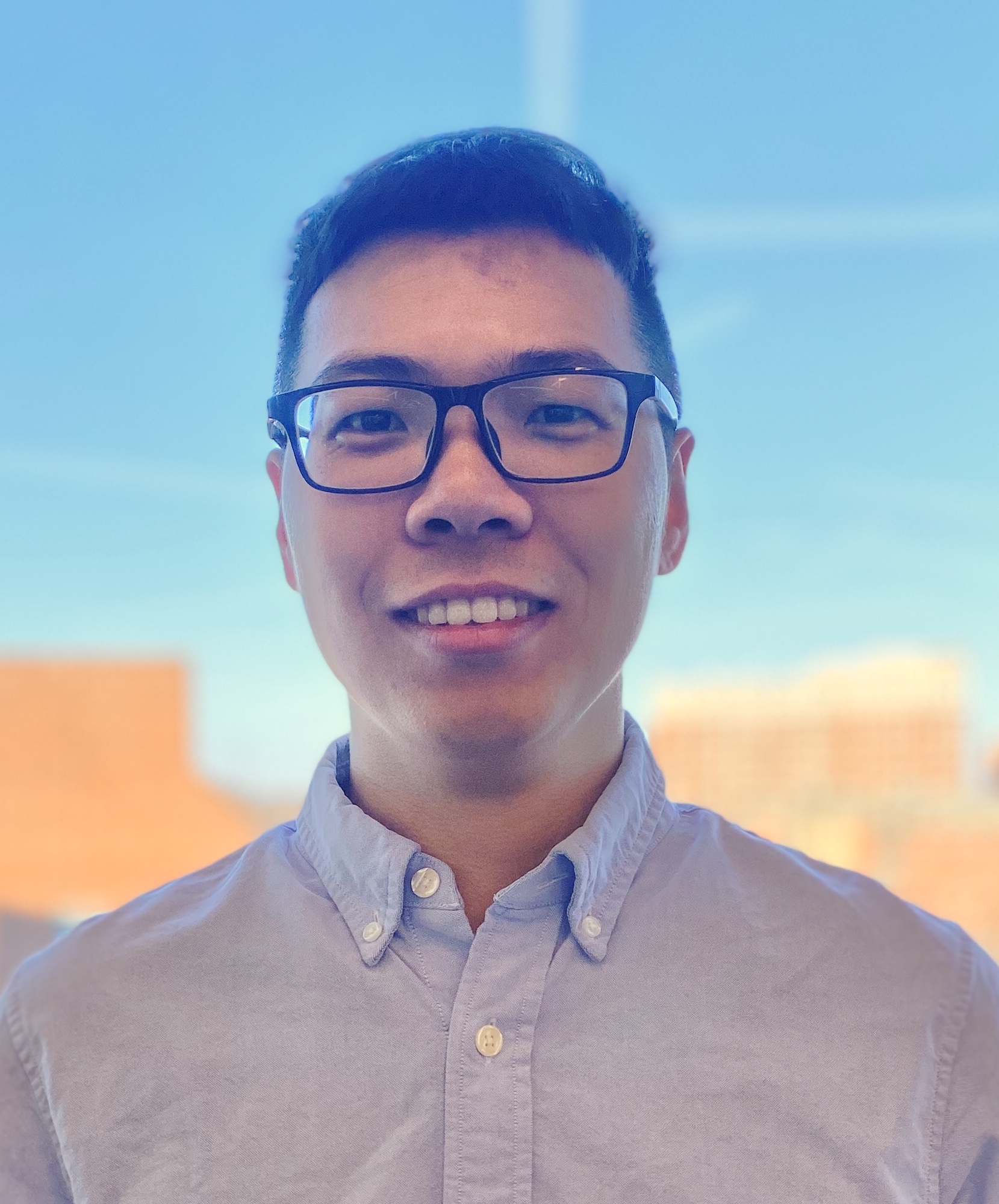}}]{Lifeng Zhou} is currently a Postdoctoral Researcher in the GRASP Lab at the University of Pennsylvania. He received his Ph.D. degree in Electrical \& Computer Engineering at Virginia Tech in 2020. He obtained his master’s degree in Automation from Shanghai Jiao Tong University, China in 2016, and his Bachelor's degree in Automation from Huazhong University of Science and Technology, China in 2013. 

His research interests include multi-robot coordination, approximation algorithms, combinatorial optimization, model predictive control, graph neural networks, and resilient, risk-aware decision making.
\end{IEEEbiography}

\begin{IEEEbiography}[{\includegraphics[width=1in,height=1.25in,clip,keepaspectratio]{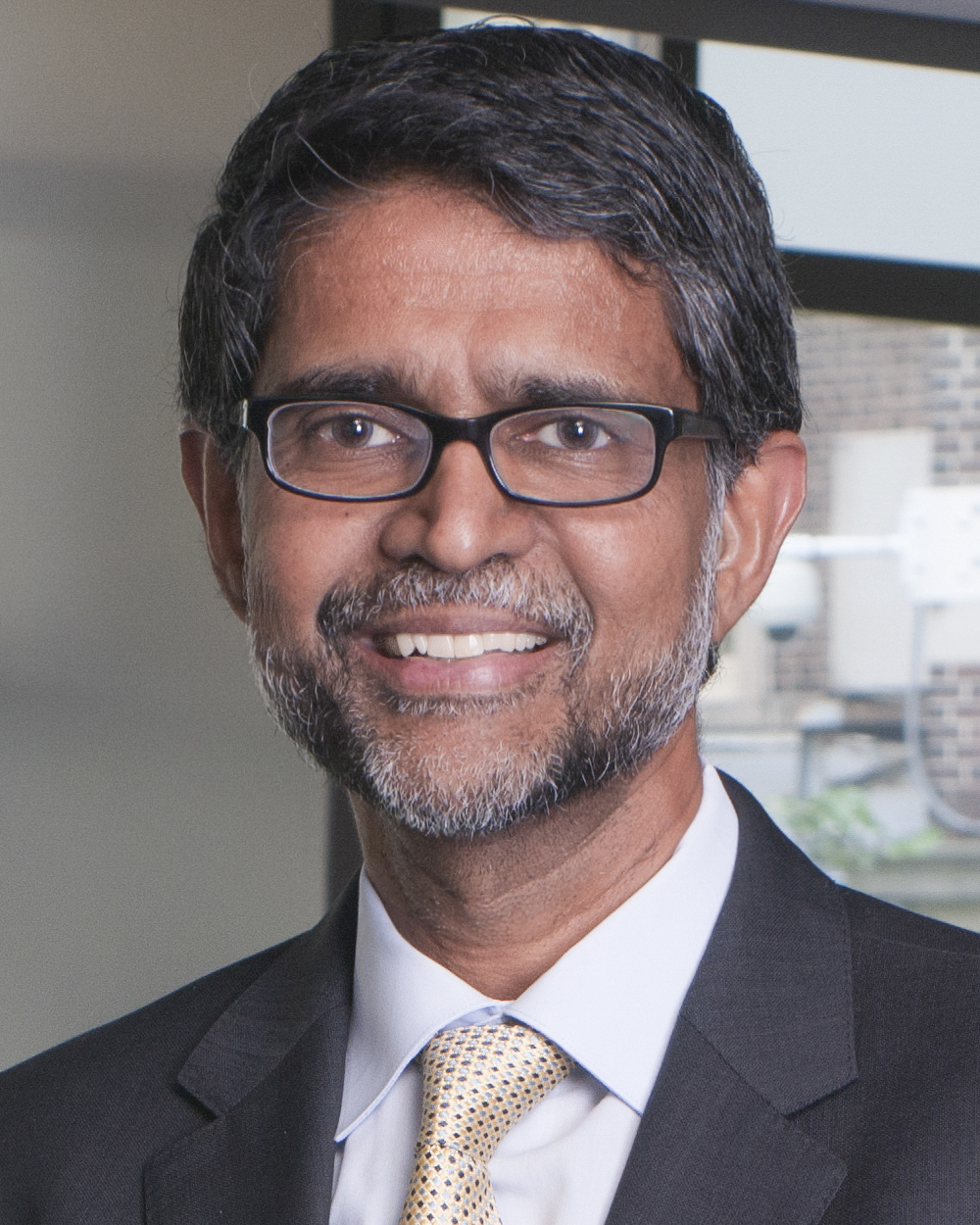}}]{Vijay Kumar}
  (F’05) received the Ph.D. degree in mechanical engineering from The Ohio State University, Columbus, OH, USA, in 1987.

He is the Nemirovsky Family Dean of Penn Engineering (with appointments) with the Department of Mechanical Engineering and Applied Mechanics, the Department of Computer and Information Science, and the Department of Electrical and Systems Engineering, University of Pennsylvania, Philadelphia, PA, USA. He served as the Assistant Director of Robotics and Cyber Physical Systems with the White House Office of Science and Technology Policy from 2012 to 2014. He is the Founder of Exyn Technologies, a company that develops solutions for autonomous flight. Dr. Kumar became a Fellow of the American Society of Mechanical Engineers in 2003 and a member of the National Academy of Engineering in 2013. He is the recipient of the 2012 World Technology Network Award, the 2013 Popular Science Breakthrough Award, the 2014 Engelberger Robotics Award, the 2017 IEEE Robotics and Automation Society George Saridis Leadership Award in Robotics and Automation, and the 2017 ASME Robert E. Abbott Award. He has received best paper awards at the 2002 International Symposium on Distributed Autonomous Robotic Systems (DARS), the 2004 IEEE International Conference on Robotics and Automation (ICRA), ICRA 2011, the 2011 Robotics: Science and Systems Conference (RSS), RSS 2013, and the 2015 EAI International Conference on Bio-Inspired Information and Communications Technologies. He has advised doctoral students, who have received best student paper awards at ICRA 2008, RSS 2009, and DARS 2010.
\end{IEEEbiography}

\end{document}